\documentclass{kais}
\usepackage{graphicx}
\usepackage[linesnumbered,algoruled,boxed,lined]{algorithm2e}
\usepackage{amsmath, amssymb}
\usepackage{mdframed}
\usepackage{caption}
\usepackage{subcaption}
\usepackage[sorting=none]{biblatex}
\usepackage{wrapfig}
\usepackage{biblatex}
\usepackage{todonotes}
\newtheorem{definition}{Definition}[section]

\newcommand{\mathbbm}[1]{\text{\usefont{U}{bbm}{m}{n}#1}}

\newmdtheoremenv[linewidth=1pt, skipabove=6pt, skipbelow=6pt]{theorem_md}{Theorem}
\newmdtheoremenv[linewidth=1pt, skipabove=6pt, skipbelow=6pt]{observation}{Observation}
\newmdtheoremenv[linewidth=1pt, skipabove=6pt, skipbelow=6pt]{proposition_md}{Proposition}

\makeatletter
\newcommand*{\myfnsymbolsingle}[1]{%
  \ensuremath{%
    \ifcase#1
    \or 
      *%
    \or 
      1
    \or 
      2
    \or 
      3
    \or 
      4
    \or 
      5
    \else 
      \@ctrerr  
    \fi
  }%
}   
\makeatother

\usepackage{alphalph}
\newalphalph{\myfnsymbolmult}[mult]{\myfnsymbolsingle}{}

\received{xxx}
\revised{xxx}
\accepted{xxx}

\pubyear{2000}
\pagerange{\pageref{firstpage}--\pageref{lastpage}}
\volume{xxx}

\begin{document}
\label{firstpage}

\title{Micro- and Macro-Level Churn Analysis of Large-Scale Mobile Games}

\author[X. Liu et al]{
\parbox{\linewidth}{Xi Liu$^{1}$\thanks{This work was done during the authors' internships at Samsung Research America}, Muhe Xie$^2$, Xidao Wen$^{3*}$, Rui Chen$^2$, Yong Ge$^4$, Nick Duffield$^1$ and Na Wang$^2$}\\
$^1$Texas A\&M University, College Station, TX, USA\\
$^2$Samsung Research America, Mountain View, CA, USA\\
$^3$University of Pittsburgh, Pittsburgh, PA, USA\\
$^4$The University of Arizona, Tucson, AZ, USA
}
 
\maketitle

\begin{abstract}
As mobile devices become more and more popular, mobile gaming has emerged as a promising market with billion-dollar revenues. A variety of mobile game platforms and services have been developed around the world. A critical challenge for these platforms and services is to understand the churn behavior in mobile games, which usually involves churn at micro level (between an app and a specific user) and macro level (between an app and all its users). Accurate micro-level churn prediction and macro-level churn ranking will benefit many stakeholders such as game developers, advertisers, and platform operators. In this paper, we present the first large-scale churn analysis for mobile games that supports both micro-level churn prediction and macro-level churn ranking. For micro-level churn prediction, in view of the common limitations of the state-of-the-art methods built upon traditional machine learning models, we devise a novel semi-supervised and inductive embedding model that jointly learns the prediction function and the embedding function for user-app relationships. We model these two functions by deep neural networks with a unique edge embedding technique that is able to capture both contextual information and relationship dynamics. We also design a novel attributed random walk technique that takes into consideration both topological adjacency and attribute similarities. To address macro-level churn ranking, we propose to construct a relationship graph with estimated micro-level churn probabilities as edge weights and adapt link analysis algorithms on the graph. We devise a simple algorithm \emph{SimSum} and adapt two more advanced algorithms \emph{PageRank} and \emph{HITS}. The performance of our solutions for the two-level churn analysis problems is evaluated on real-world data collected from the Samsung Game Launcher platform. The data includes tens of thousands of mobile games and hundreds of millions of user-app interactions. The experimental results with this data demonstrate the superiority of our proposed models against existing state-of-the-art methods.
\end{abstract}


\section{Introduction}
\label{section:introduction}

With the wide adoption of mobile devices (e.g., smartphones and tablets), the past decade has seen a rapid increase of the mobile gaming industry into a billion-dollar market around the globe. \$70.3 billion revenue was generated by mobile games in 2018 according to Newzoo's Global Games Market Report~\cite{NewZooReport}. And it is expected that mobile games will generate \$106.3 billion revenue in 2021, accounting for more than half of the overall game market. Driven by this increasingly vital market, software and hardware providers of mobile devices (e.g., Apple, Google and Samsung) have provided integrated mobile game platforms and services for end users, game developers and other stakeholders. Some examples of such platforms and services are Apple App Store, Google Play and Samsung Game Launcher~\cite{SamsungGameLauncher}.

One particularly crucial task, within these mobile game platforms and services, is understanding the churn behavior in mobile games, which usually involves two levels of analysis: the micro level (between an app and a specific user) and the macro level (between an app and all its users). In many traditional business applications~\cite{zaki2016fallacy}, accurately predicting micro-level churn has been a long-standing and important task, the objective of which is to predict the likelihood that a user will stop using a service or product. In our problem setting, the aim is to predict the likelihood that a user will stop using a particular game app in the future. The task is vital for the following reasons. First, the churn rate of a mobile game is an important business metric to measure its success. With accurately predicted churn probabilities of player-game pairs, a game platform is able to prioritize its resources for better operation and management. Second, predicting individual churn probabilities will enable a game platform to design better marketing strategies to improve user retention. Examples include sending push notifications and providing free items in games to users who are likely to churn. Since, as well known, the acquisition cost for new users is much higher than the retention cost for existing users, successful micro-level churn prediction could largely reduce costs for game developers and platform operators to increase the number of active users, which plays a critical role in the success of a mobile game. Third, the micro-level churn prediction provides direct input to determine the right timing of app recommendations for a game platform. The results of this research will enable testing of the hypothesis that a user is more likely to act on an app recommendation when he/she is about to stop playing other games.

The objective of macro-level churn ranking is to provide a list of games ranked by their total number of users who will churn in the near future. It is an equally important problem, and its solution can be applicable to various mobile app services such as trend prediction~\cite{malmi2014quality}, rating and review spam detection, ranking fraud detection~\cite{zhu2015popularity}, and especially trend-sensitive recommendation~\cite{cao2017version}. There has been a general consensus that the mobile app market is highly dynamic; new apps continuously enter the market and distract users' attention from existing ones. A recommender can use the results of macro-level churn ranking as a prior knowledge, e.g., an attribute in the representation of games. A recommender can also slightly adjust the position of recommended games before delivering to users in preference to those that have smaller total numbers of near-future churn, e.g., a post-filter. Macro-level churn ranking in general can be very challenging because the order of the games on the list (1) can change very fast with the dynamics of the market, and (2) is based on future events, which can only be estimated. The macro-level churn ranking problem is intimately related to the micro-level churn prediction problem. Consider the extreme case that all micro-level churn predictions are 100\% correct, then the number of users to churn in the future can be directly computed. 

There have been several previous studies \cite{hadiji2014predicting, runge2014churn, xie2015predicting, perianez2016churn,  tamassia2016predicting,  drachen2016rapid, xie2016predicting, bertens2017games, viljanen2017measuring, viljanen2017playtime,kim2017churn} on micro-level mobile game churn prediction by using traditional machine learning models (e.g., logistic regression, random forests, Cox regression). However, we observe several major limitations of these studies. First, they were developed for predicting churn of a single or a few mobile games; none is capable of handling churn prediction of large-scale mobile apps and users. For real-world applications, a solution needs to handle tens of thousands of mobile games and hundreds of millions of user-app interactions on a daily basis. Second, user-app interaction data often comes with rich contextual information (e.g., WiFi connection status, screen brightness, and audio volume), which has never been considered in the existing studies. Third, the existing methods rely on handcrafted features that usually cannot scale well in practice. 

To overcome all these limitations, we propose a novel inductive semi-supervised embedding model that \emph{jointly} learns the prediction function and the embedding function for user-game interaction. The user-app interaction data includes detailed information of opens, closes, installs and uninstalls of game apps for each individual user. This data is collected from the Samsung Game Launcher platform\footnote{https://www.samsung.com/au/support/mobile-devices/how-to-use-game-tools/}, which is pre-loaded in most smartphones manufactured by Samsung. We model the interplay between users and games by an attributed bipartite graph and then learn these two functions by deep neural networks with a unique embedding technique that is able to capture both contextual information and dynamic user-game interaction. Our method is fully automatic and can be easily integrated into existing mobile platforms.

With the predicted micro-level churn probabilities, we devise a simple method \emph{SimSum} to estimate the macro-level churn ranking. Under the presupposition that the churn probabilities are properly estimated, we show that \emph{SimSum} is based on an unbiased estimation. Furthermore, as the churn behavior may be related among users and games, we propose to construct a user-game graph with edges weighted by the estimated churn probabilities and adapt link analysis algorithms to infer the macro-level churn ranking of games. Among the family of link analysis algorithms, \emph{PageRank}~\cite{page1999pagerank} and \emph{HITS} (Hyperlink-Induced Topic Search)~\cite{kleinberg1999authoritative} are chosen because they have been widely applied in abundant real-world applications and shown to have stable performance.

\vspace{3mm}
\noindent \textbf{Contributions.} Our research contributions are summarized as follows:
\begin{enumerate}
\item To the best of our knowledge, this paper is the first to develop a solution for both micro-level churn prediction and macro-level churn ranking of large-scale mobile games using hundreds of millions of user-app interaction records. This solution has been tested in Samsung Game Launcher, one of the largest commercial mobile gaming platforms. Although the paper mainly applies the proposed solution to mobile game churn analysis, the solution is also applicable to churn analysis in other contexts.

\item We propose a novel semi-supervised and inductive model based on embedding. Our model can capture the dynamics between users and mobile games based on the introduced temporal loss in the formulated objective function. The model is able to embed new users or games not used in training. This is critical for mobile game churn prediction because new games and users continually enter the market. 

\item We develop an attributed random walk technique that enables us to sample the contexts of edges in an attributed bipartite graph and that takes into account both topological adjacency and attribute similarities.

\item We propose a simple method \emph{SimSum} and adapt two link analysis algorithms \emph{PageRank} and \emph{HITS} to solve the macro-level churn ranking problem.

\item We conduct a comprehensive experimental evaluation with large-scale real-world data collected from Samsung Game Launcher. The experimental results demonstrate that our model outperforms all state-of-the-art methods with respect to different evaluation metrics for micro-level churn prediction and macro-level churn ranking.
\end{enumerate}

The rest of the paper is organized as follows. Section~\ref{sec:problem} formulates the two-level mobile game churn analysis problem. Section~\ref{sec:methods} discusses our solution in detail. Section~\ref{sec:experiment} presents our experimental results on large-scale real-world data. Section~\ref{sec:related_work} reviews the related literature. Finally, Section~\ref{sec:conclusion} concludes our work.

\section{Problem Formulation}
\label{sec:problem}

\begin{figure}
\centering
\includegraphics[width=.5\textwidth]{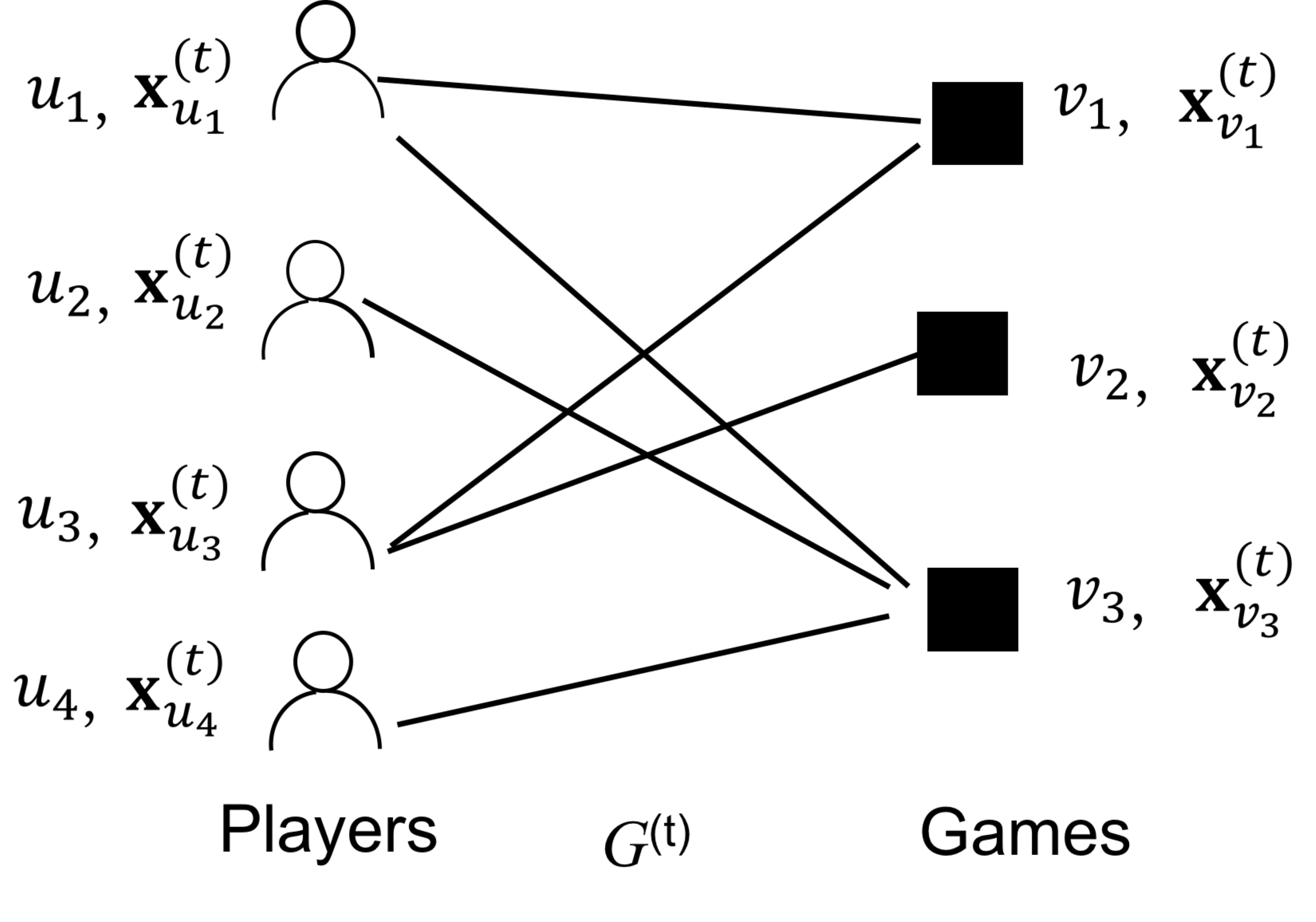}
\caption{An example of an attributed bipartite graph for mobile game churn prediction}
\label{fig:dislink_detection}
\end{figure}

In the context of mobile games, \emph{churn} is defined as a player stopping using a game within a given period (i.e., there is no app usage in the period). The duration $T$ of the period may vary from application to application depending on different business goals. $T=14$ days and $T=30$ days are some typical settings used in industry \cite{hadiji2014predicting,runge2014churn,drachen2016rapid}. In this paper, we consider the generic micro-level game churn prediction and macro-level churn ranking problems without assuming any particular value of $T$. We note that uninstall is different from churn. Regarding only uninstall as churn would be problematic since there may be a large time gap between cessation of playing and uninstall, if any. 

\begin{table}
\centering
\caption{Notations and definitions}
\begin{tabular}{|c | c|}
\hline
Notations & Descriptions or Definitions \\
\hline
$\mathcal{G}^{(t)}$ & Attributed graph at time $t$ \\ \hline
$\mathcal{H}^{(t)}$ & Historical attributed graphs $\{\mathcal{G}_{i}\}_{i = t_0}^{i=t}$ \\ \hline
$\mathcal{U}^{(t)}$ & Set of all player nodes in $\mathcal{G}^{(t)}$ \\ \hline
$\mathcal{V}^{(t)}$ & Set of all game nodes in $\mathcal{G}^{(t)}$ \\ \hline
$\mathcal{E}^{(t)}$ & Set of all edges in $\mathcal{G}^{(t)}$ \\ \hline
$e_{uv}^{(t)}$ & Indicator of the existence of edge $(u,v)$ in $\mathcal{G}^{(t)}$ \\ \hline
$\mathbf{x}_{u}^{(t)}$ & Feature vector of user $u\in \mathcal{U}^{(t)}$ \\ \hline
$\mathbf{x}_{v}^{(t)}$ & Feature vector of game $v \in \mathcal{V}^{(t)}$ \\ \hline
$\mathbf{z}_{uv}^{(t)}$ & Aggregated feature vector of edge $(u,v) \in \mathcal{E}^{(t)}$ \\ \hline
$n_u$, $n_v$ & Numbers of attributes in $\mathbf{x}_{u}^{(t)}$ and $\mathbf{x}_{v}^{(t)}$, resp. \\ \hline
$d$ & Number of attributes in $\mathbf{z}_{uv}^{(t)}$ \\ \hline
$m$ & Embedding dimension \\ \hline
$g$ & Edge embedding function $g:\mathcal{R}^{d} \rightarrow \mathcal{R}^{m}$ \\ \hline
$f$ & Churn prediction function $f: \mathcal{R}^{m} \rightarrow [0, 1]$ \\ \hline
$l_{p}$ & Number of embedding hidden layers in Part I \\ \hline
$l_{n}$ & Number of prediction hidden layers in Part III \\ \hline
$|\cdot|$ & Cardinality of a set \\ \hline
$A \setminus B$ & Set of elements in $A$ but not in $B$
\\ \hline
$\mathbbm{1}(c)$ & Indicator function for condition $c$
\\ \hline
$\mathcal{S}^{(t)}(v)$ & Ranking score of game $v$ at time $t$
\\ \hline
\end{tabular}
\label{table:notations}
\end{table}

The relationship between players and games can be represented by an attributed bipartite graph as illustrated in Fig.\ref{fig:dislink_detection}, whose two parts correspond to players and games. In the sequel, we use the terms \emph{player} and \emph{user}, and \emph{game} and \emph{app} interchangeably. Let $\mathcal{G}^{(t)}$ be the attributed bipartite graph at time $t$, the vertex set $\mathcal{U}^{(t)}$ denote the set of users and the vertex set $ \mathcal{V}^{(t)}$ denote the set of games. A player is represented by a node $u\in \mathcal{U}^{(t)}$ and a game is represented by a node $v \in \mathcal{V}^{(t)}$. Each user $u$ is associated with a feature vector $\mathbf{x}_{u}^{(t)}\in \mathcal{R}^{n_{u}}$, where $n_u$ is the size of $\mathbf{x}_{u}^{(t)}$; each game $v$ is associated with a feature vector $\mathbf{x}_{v}^{(t)}\in \mathcal{R}^{n_{v}}$, where $n_{v}$ is the size of $\mathbf{x}_{v}^{(t)}$. There is an edge between nodes $u$ and $v$ in $\mathcal{G}^{(t)}$ if player $u$ has played $v$ in the time window $[t+1, t+T]$. The set of neighbor nodes that connect to game $v$ at time $t$ is denoted by $\mathcal{N}_{v}^{(t)}$. The set of edges at time $t$ is denoted by $\mathcal{E}^{(t)}$.

Now we are ready to define the two-level mobile game churn analysis problem below.
\begin{definition} [\emph{Mobile game churn analysis}]
Consider a collection of attributed bipartite graphs observed from time $t_0$ to time $t$ ($t > t_0$), which is denoted by $\mathcal{H}^{(t)} = \{\mathcal{G}^{(i)}\}_{i = t_0}^{i=t}$. Let $\mathcal{N}_{v}^{(i)}$ be the number of neighbors of game $v$ at time $i$ and $e_{uv}^{(i)}$ be the indicator of the existence of edge $(u, v)$ in $\mathcal{G}^{(i)}$. The dual objectives are:
\begin{enumerate}
    \item \emph{Micro-level churn prediction} that estimates the probability $Pr(e_{uv}^{(t+1)} = 0 | e_{uv}^{(t)} = 1, \mathcal{H}^{(t)})$ for an edge $(u, v)$ in $\mathcal{G}^{(t)}$, which is the probability that $(u, v)$ disappears in $\mathcal{G}^{(t+1)}$;
    \item \emph{Macro-level churn ranking} that ranks the games in a descending order of the total number of users to churn at time $t + 1$. 
\end{enumerate}
\end{definition}

The notations used in this paper and their descriptions are listed in Table~\ref{table:notations}.

\section{Methods}
\label{sec:methods}

\subsection{Overview of Our Solution}

Most existing works~\cite{hadiji2014predicting, runge2014churn, xie2015predicting, perianez2016churn,  tamassia2016predicting,  drachen2016rapid, xie2016predicting, bertens2017games, viljanen2017measuring, viljanen2017playtime,kim2017churn} rely on traditional machine learning models to solve the micro-level churn prediction problem, which suffer from several major limitations as mentioned in Section~\ref{section:introduction}. In view of the recent progress in graph embedding, a promising way is to adopt graph embedding frameworks for churn prediction. However, we face several key technical challenges that have not been addressed in prior studies: (1) Most existing methods are transductive, and thus cannot produce embeddings for new player-game pairs. (2) Existing methods are either purely supervised or unsupervised, and thus do not take full advantage of relevance between embedding and a task. (3) Existing methods are node-centric, and thus are not directly applicable to edge-centric tasks. (4) Existing methods mainly handle a static graph and do not incorporate graph dynamics in embedding. In the mobile game industry, players and games, however, change very quickly. There are many new players, new games, and new player-game relationships almost every day. 

\begin{figure}
\centering
\includegraphics[width=1\textwidth, height=.425\textwidth]{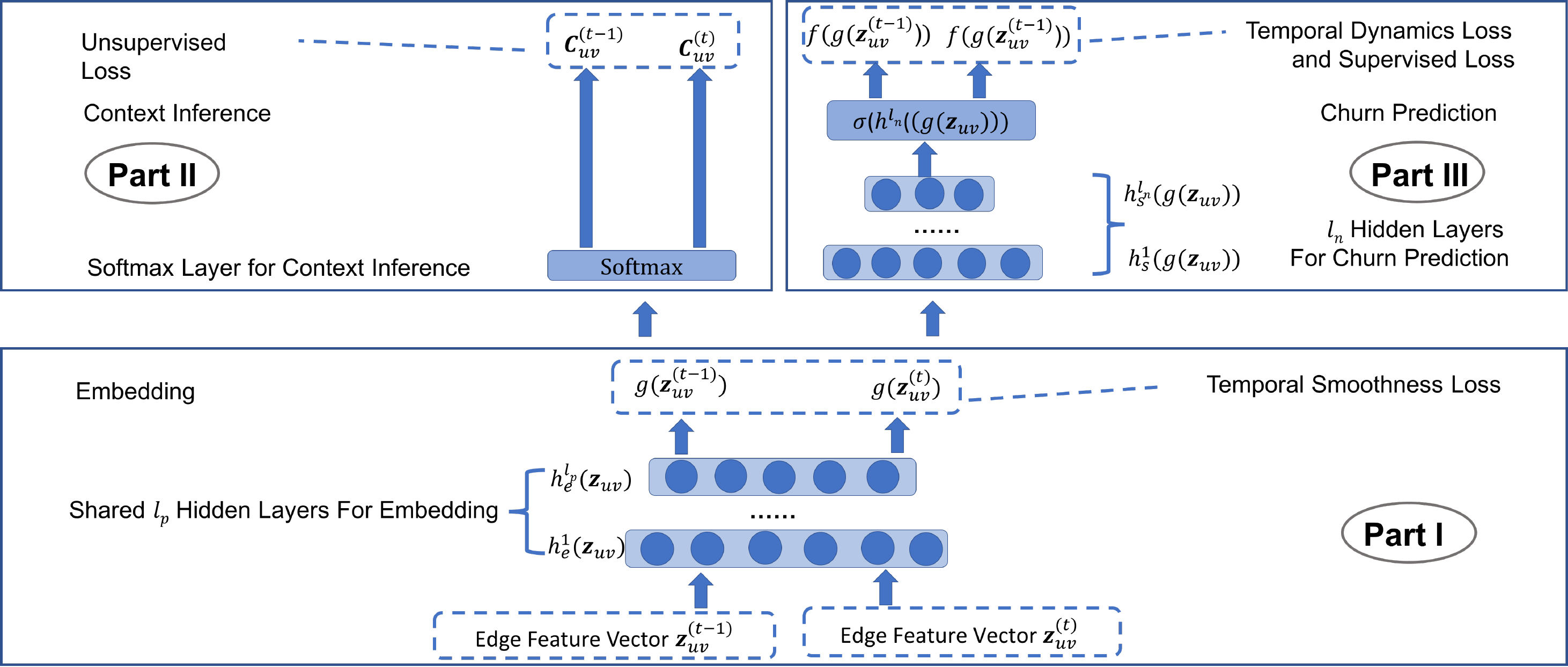} 
\caption{Deep neural network architecture of the inductive semi-supervised embedding model in training}
\label{fig:neural_net_archi}
\end{figure}

In addressing these challenges, we propose a novel inductive semi-supervised embedding model in dynamic graphs that jointly learns the prediction function $f$ and the embedding function $g$. The prediction function $f$ and the embedding function $g$ are learned by deep neural networks (DNN). The architecture of the proposed DNN is presented in Fig. \ref{fig:neural_net_archi}, which consists of three parts. Part I is responsible for producing embedding feature vectors $g(\mathbf{z}_{uv}^{(t)})$ from raw edge feature vectors $\mathbf{z}_{uv}^{(t)} \in \mathcal{R}^{d}$, where $d$ is the size of raw edge feature vectors. To learn the probability of churn, we need to construct a feature vector for each $(u,v)$ with $e_{uv}^{(t)} = 1$. However, it is impractical to calculate features for all possible edges that may appear in the prediction period because the number of possible edges is huge, which is $\mathcal{O}(|\mathcal{U}^{(t)}|\cdot |\mathcal{V}^{(t)}|)$. Instead, we construct the feature vector $\mathbf{z}_{uv}^{(t)}$ of $(u,v)$ from attribute-wise cosine similarity aggregation of $\mathbf{x}_{u}(t)$ and $\mathbf{x}_{v}(t)$. As such, all $\mathbf{z}_{uv}^{(t)}$ can be readily computed from the features of $|\mathcal{U}^{(t)}| +|\mathcal{V}^{(t)}|$ nodes, where $|\cdot|$ denotes the cardinality of a set. 

Part II is for inferring contexts from embedding feature vectors. Here the context of an edge refers to the edges that are similar to and co-occur with the edge under some graph sampling strategy, for example, random walk. Part I and Part II form the \emph{unsupervised} component of our model. They are jointly trained by minimizing the error of incorrect context inference and inconsistency with temporal smoothness (see Section~\ref{subsection:temporal_loss} for an explanation of temporal smoothness). Part I and Part II are trained in an inductive and edge-centric way. In contrast to transductive node embedding that learns a distinct embedding vector for each node, our model is to learn an embedding function that is generalized to any \emph{unseen} edges as long as their feature vectors are available. We propose a novel attributed random walk to sample similar edges as contexts (see Section~\ref{subsection:attributed_random_walk} for details). 

Part III fulfills the supervised churn prediction task from embedding feature vectors. Part III forms the \emph{supervised} component of the proposed model, which is trained by minimizing the error of incorrect churn predictions. The supervised component and unsupervised component are simultaneously trained as combined into a single objective function. Traditional unsupervised embedding techniques are not designed in a task-specific way and hence are not able to incorporate task-specific information to improve performance. In contrast, in our model Part III and Part II share the common hidden layers in Part I, and therefore they are implicitly coupled with each other. This helps the embedding align with the supervised prediction task. 

Part I and Part III both consider graph dynamics in training. Part I handles graph dynamics by requiring the embeddings of the same edge at two consecutive timestamps to stay close. Part III handles graph dynamics by requiring the churn probabilities of the same edge at two consecutive timestamps to follow a decaying pattern.

The overall objective function of our model consists of four parts:
\begin{align}\label{equ:objective}
    \mathcal{L} := \mathcal{L}_{S} + \alpha \mathcal{L}_{U} + \beta  \mathcal{L}_{T} + \gamma \mathcal{L}_{R}.
\end{align}
$\mathcal{L}_{S}$ denotes the supervised loss due to incorrect predictions and will be discussed in Section~\ref{subsection:supervised_loss}. $\mathcal{L}_{U}$ denotes the unsupervised loss, which comes from failures of context inference and will be addressed in Section~\ref{subsection:unsupervised_loss}. $\mathcal{L}_{T}$ is the temporal loss that consists of two parts: temporal smoothness and temporal dynamics, and will be explained in Section~\ref{subsection:temporal_loss}. $\mathcal{L}_{R}$ presented in Section~\ref{subsection:reg_loss} is the regularization term, and ($\alpha$, $\beta$, $\gamma$) are trade-off weights. 


After the prediction function $f$ and embedding function $g$ are learned through minimizing the loss function in Equation (\ref{equ:objective}), the churn probability $Pr(e_{uv}^{(t+1)} = 0 | e_{uv}^{(t)} = 1, \mathcal{H}^{(t)})$ of any individual user-game pair $(u,v)$ can be estimated by $f(g(\mathbf{z}^{(t)}_{uv}))$. Under the assumption that the micro-level churn probability is properly estimated, we are able to show that for a specific game $v$, the sum of its churn probabilities of all its users is an \emph{unbiased} estimator of the ground truth $|\mathcal{N}_{v}^{(t)}\setminus \mathcal{N}_{v}^{(t+1)}|$. This inspires us to derive a simple solution for macro-level churn ranking: first estimate the total number of churns $|\mathcal{N}_{v}^{(t)}\setminus \mathcal{N}_{v}^{(t+1)}|$ for any game $v \in \mathcal{V}^{(t)}$ by summing all its users' churn probabilities and then rank all games based on the estimations. This method is referred to as \emph{SimSum}. Furthermore, we observe that there may be correlations in the churn behavior between different games and different users. For instance, for the same user, the probability of his/her churning a game is likely to be influenced by the probability of churning other games. On the other hand, similar users may exhibit similar patterns of churning the same game. Therefore, we propose to construct a user-game graph with edges weighted by the estimated churn probabilities and adapt link analysis algorithms to estimate macro-level churn ranking. Specifically, \emph{PageRank}~\cite{page1999pagerank} and \emph{HITS}~\cite{kleinberg1999authoritative} are chosen due to their wide applications and stable performance. Details will be discussed in Section~\ref{sec:macro-level}.

\subsection{Static Loss Functions}\label{subsection:loss_functions}
\subsubsection{Supervised Loss Function $\mathcal{L}_{S}$}\label{subsection:supervised_loss}
The supervised loss function $\mathcal{L}_{S}$ is designed for Part III. Let $h_{s}^{k}(g(\mathbf{z}_{uv}^{(t)})) = \phi(W_{s}^{k}h_{s}^{k-1}(g(\mathbf{z}_{uv}^{(t)})) + b^{k})$ be the $k$-th hidden layer for churn prediction (referred to as \emph{prediction hidden layer} in the sequel), where $W_{s}^{k}$ and $b_{s}^{k}$ are the weights and biases in the $k$-th prediction hidden layer, and $\phi(\cdot)$ is a non-linear activation function. We model the churn prediction function $f$ by $l_{n}$ such layers in Part III. Then the prediction output layer can be represented by:
\begin{align}\label{equ:supervised_loss}
    f(g(\mathbf{z}_{uv}^{(t)})) &:=\widehat{P}r(e_{uv}^{(t+1)} = 0 | e_{uv}^{(t)} = 1, \mathcal{H}^{(t)}) \\ \nonumber &:=\sigma(h_{s}^{l_{n}}(g(\mathbf{z}_{uv}^{(t)}))) \\ \nonumber
    &:= \frac{\exp{\big(h_{s}^{l_{n}}(g(\mathbf{z}_{uv}^{(t)}))^{T}w_{s}}\big)}{1+\exp{\big(h_{s}^{l_{n}}(g(\mathbf{z}_{uv}^{(t)}))^{T}w_{s}}\big)}
\end{align}
where $\widehat{P}r(\cdot)$ is the estimate of $Pr(\cdot)$, $\sigma(\cdot)$ is the sigmoid function, and $w_{s}$ is the sigmoid weights vector that combines the output from the last hidden layer to predict churn. Now we can define the supervised loss $\mathcal{L}_{S}$ as follows:
\begin{align}
    \mathcal{L}_{S} =& \frac{1}{L}\sum_{i = 0}^{t-1} \sum_{(u,v) \in \mathcal{E}^{(i)}} \delta_{uv}^{(i+1)}\Big(1-e_{uv}^{(i+1)} - \frac{\exp{\big(h_{s}^{l_{n}}(g(\mathbf{z}_{uv}^{(i)}))^{T}w_{s}}\big)}{1+\exp{\big(h_{s}^{l_{n}}(g(\mathbf{z}_{uv}^{(i)}))^{T}w_{s}}\big)}\Big)^{2},
\end{align}
 where $L$ is the number of training examples and $\delta_{uv}^{(i+1)}$ is a censoring indicator that will be discussed in Section~\ref{subsection:temporal_loss}.

\subsubsection{Unsupervised Loss Function $\mathcal{L}_{U}$}\label{subsection:unsupervised_loss}
The unsupervised loss function $\mathcal{L}_{U}$ is devised for Part II, which guides to embed the handcrafted features $\mathbf{z}_{uv}^{(t)}\in \mathcal{R}^{d}$ into a latent space $g(\mathbf{z}_{uv}^{(t)})\in \mathcal{R}^{m}$, where $m$ is the size of the latent space. Denote the $k$-th hidden layer for embedding (referred to as \emph{embedding hidden layer} in the sequel) by $h_{e}^{k}(\mathbf{z}_{uv}^{(t)}) = \phi(W_{e}^{k}h_{e}^{k-1}(\mathbf{z}_{uv}^{(t)}) + b_{e}^{k})$, where $W_{e}^{k}$ and $b_{e}^{k}$ are the weights and biases in the $k$-th embedding hidden layer. We use the $l_{p}$ layers in Part I to represent the embedding function $g$, and the embedding output layer can be represented by:
\begin{align}\label{equ:unsupervised_loss}
    &g(\mathbf{z}_{uv}^{(t)}) := h_{e}^{l_{p}}(\mathbf{z}_{uv}^{(t)}).
\end{align}
  
We can define the unsupervised loss function as follows:
\begin{align}
    \mathcal{L}_{U} = -\sum_{i=t_0}^{t}\sum_{(u,v) \in \mathcal{E}^{(i)}} \sum_{(u',v')\in C_{uv}^{(i)}}\delta_{uv}^{(i)} \log\Big(Pr\big((u',v')|g(\mathbf{z}_{uv}^{(i)}\big)\Big),
\end{align}
where $C_{uv}^{(i)}$ denotes the context (i.e., contextual edges) of $(u,v)$ in $\mathcal{G}^{(i)}$. The contextual edges $C_{uv}^{(i)}$ are obtained by attributed random walk on the bipartite graph, which will be discussed in Section~\ref{subsection:attributed_random_walk}.

The likelihood of having a contextual edge $(u', v')$ of $(u,v)$ conditional on the embedding of $(u,v)$ is:
\begin{align}
    Pr\big((u',v')|g(\mathbf{z}_{uv}^{(i)})\big) = \frac{\exp\Big( g(\mathbf{z}_{uv}^{(i)})^{T}w_{u'v'}\Big)}{{\sum_{(u^{*},v^{*})}\exp\Big(g(\mathbf{z}_{uv}^{(i)})^{T}w_{u^{*}v^{*}}\Big)}},
\end{align}
where $w_{u^{*}v^{*}}$ and $w_{u'v'}$ are the vectors of weights for edges $(u^{*},v^{*})$ and $(u', v')$ in the softmax layer, respectively. The denominator is computed by negative sampling \cite{liang2017seano}. Although the embedding function is learned by training a context inference task, it is still considered as ``unsupervised'' because the contexts are calculated by sampling on the attributed graph, which is independent of any supervised learning task~\cite{yang2016revisiting}.

The objective function in Equation (\ref{equ:objective}) contains both the supervised loss function and the unsupervised loss function. Thus the embedding hidden layers are jointly trained with prediction hidden layers. Compared to traditional embedding methods, the semi-supervised approach makes the embedding more suitable for the prediction task. 

Note that the target of the embedding process is to learn a mapping function from the feature space to the embedding space, instead of directly learning the embeddings. Thus, the input for the embedding hidden layers only contains attributes. A new edge can be embedded for churn prediction as long as we can observe its attributes. This indicates that the proposed approach is inductive. It is able to produce the embedding for an edge that is not used in training.

\subsubsection{Regularization Loss $\mathcal{L}_{R}$}\label{subsection:reg_loss}
Regularization loss is introduced mainly to avoid overfitting. The weights for regularization consists of $\{\{W^{i}_{e}, b_{e}^{i}\}_{i=1}^{l_{p}}, \{W^{i}_{s}, b_{s}^{i}\}_{i=1}^{l_{n}}, w_{s}\}$. Therefore the regularization part can be expressed as:
\begin{eqnarray}
    \mathcal{L}_{R} &:=& \lambda_{0}\sum_{i = 1} ^{l_{p}}\| W^{i}_{e} \|^{2}_{2} +\lambda_{1}\sum_{i = 1} ^{l_{p}}\| b^{i}_{e}\|^{2}_{2} +  \lambda_{2}\sum_{i = 1} ^{l_{n}}\| W^{i}_{s} \|^{2}_{2}  \nonumber \\
     &&+\lambda_{3}\sum_{i = 1} ^{l_{n}}\| b_{s}^{k}\|^{2}_{2}  + \lambda_{4}\| w_{s}\|^{2}_{2},
\end{eqnarray}
where $\{\lambda_{i}\}_{i=0}^{4}$ are trade-off weights on different regularization terms.

\subsection{Temporal Loss Function}
\label{subsection:temporal_loss}
Temporal loss refers to the loss related to graph dynamics. Different from existing works~\cite{liang2017seano,yang2016revisiting,hamilton2017inductive}, the proposed embedding model takes into account graph dynamics. Understanding the temporal dynamics of attributed bipartite graphs is crucial to precisely model the churn behavior. We make several key observations of the temporal dynamics, which help to achieve good prediction performance in practical settings. 

\begin{observation}
\label{obs1}
For a given user-game play relationship, the longer the relationship exists, the more likely the user is to churn the game.
\end{observation}

\begin{figure}
\centering
\includegraphics[width=.65\textwidth, height=.42\textwidth]{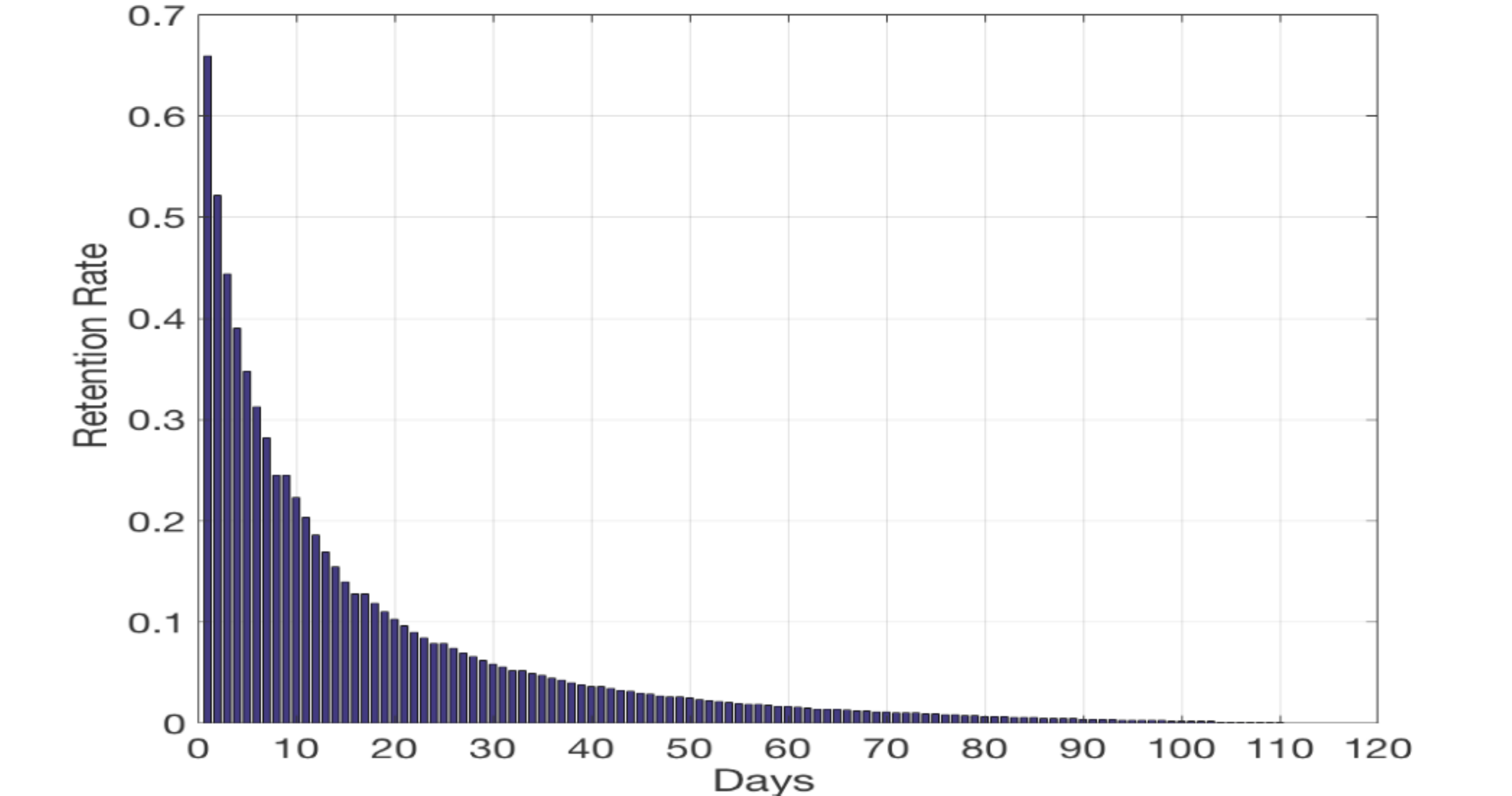}
\caption{Average mobile game retention rate as a function of time}
\label{fig:life_value}
\end{figure}

This is because the content of a mobile game is usually somewhat fixed. Players can easily lose interests after going through all contents and passing all levels in the game, let alone many players churn before passing all levels. With more days of play, their initial interests in the game are gradually effacing. Indeed, $71\%$ of all mobile app users churn within 90 days~\cite{AppChurn}. The churn rate of mobile games is even higher. We plot the average retention rate of mobile games as a function of time, based on Game Launcher data in Fig.~\ref{fig:life_value}. It shows that 95\% of user-game play relationships end after 40 days, which well justifies our observation. We formally state Observation~\ref{obs1} below.
\begin{align}\label{equ:temporal_dynamics}
    f(g(\mathbf{z}_{uv}^{(i)})) \lessapprox f(g(\mathbf{z}_{uv}^{(i+1)})) \ \forall 0\leq i \leq t-1, (u,v) \in D^{(i)},
\end{align}
where $D^{(i)} = \{(u,v): (u,v) \in \mathcal{E}^{(i)}\cap \mathcal{E}^{(i+1)} \}$ and $\lessapprox$ represents ``almost always smaller than''. We refer to this observation as \emph{temporal dynamics} in the following discussion. 

The second observation we make is stated in Observation~\ref{obs2}.
\begin{observation}
\label{obs2}
For a given user-game play relationship denoted by an edge in the attributed bipartite graph, its context usually evolves slowly at two consecutive timestamps.
\end{observation}

This observation is also derived from the real-world data. Due to space limitation, we omit the figure here. It follows that the topology and attribute values of the attributed bipartite graph mostly evolve smoothly at two consecutive timestamps, resulting in similar contexts for a given edge at consecutive timestamps. Therefore, its embeddings at these two timestamps should also be close, that is,
\begin{align}\label{equ:temporal_smoothness}
    g(\mathbf{z}_{uv}^{(i+1)}) \approxeq g(\mathbf{z}_{uv}^{(i)}) \ \forall 0\leq i \leq t-1, (u,v) \in D^{(i)},
\end{align}
where $\approxeq$ represents ``almost always equal to''. We refer to this observation as \emph{temporal smoothness} in the later discussion.

\begin{figure}
\centering
\includegraphics[width=.7\textwidth]{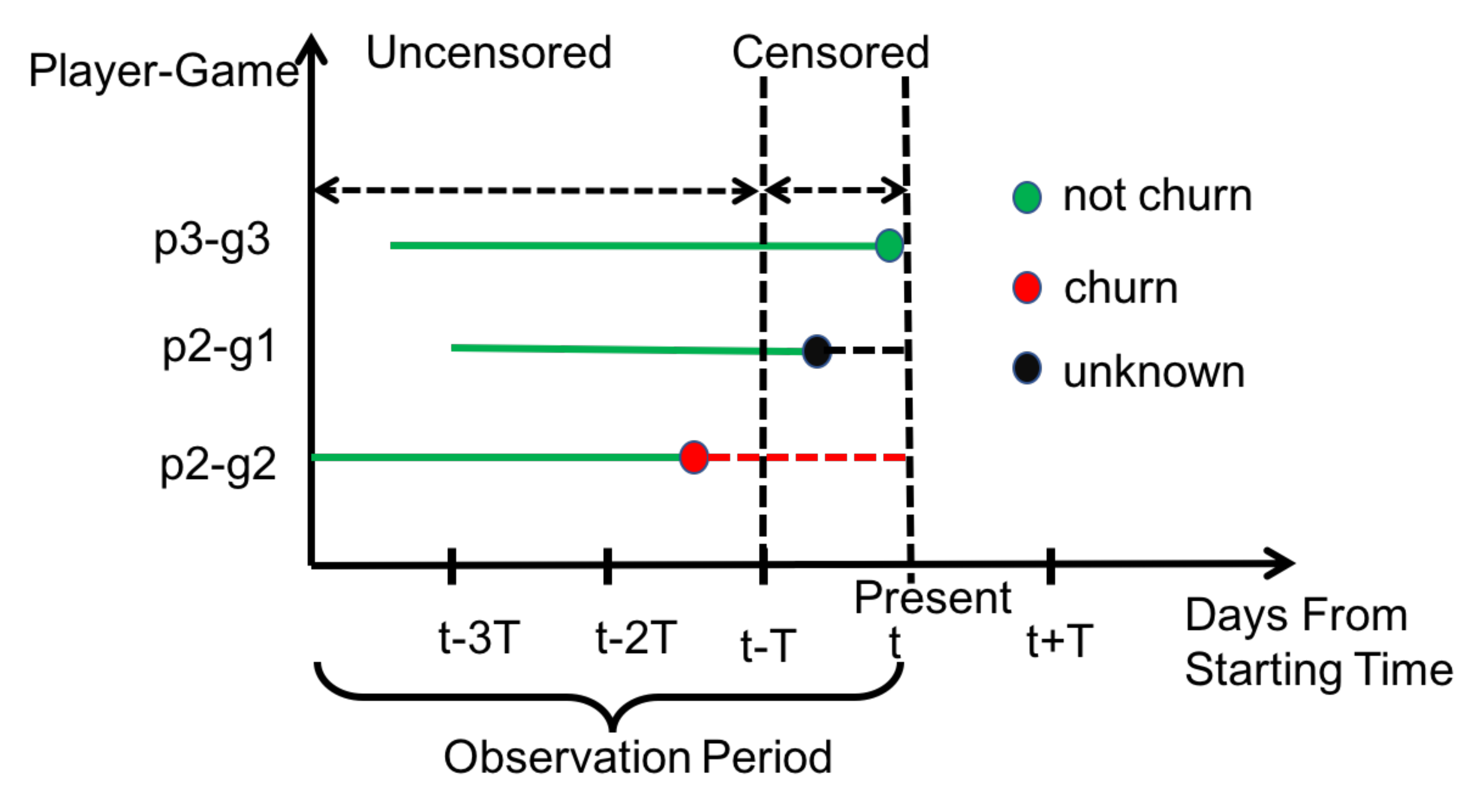}
\caption{An illustration of censored data by three randomly sampled player-game pairs}
\label{fig:censored}
\end{figure}

The final observation is:
\begin{observation}
By definition, churn in nature introduces right censoring to the training dataset.
\end{observation}
The problem of right censoring has been widely studied~\cite{wu1988estimation} and is illustrated in Fig.~\ref{fig:censored}. 
The observation period refers to some time duration in history. Suppose we are at time $t$ and the observation period is from time $t_0$ to time $t$. Data for training and testing all comes from the observation period. Since the label of a player-game pair at a specific timestamp is determined by their interaction in the next $T$ time duration, the labels of some player-game pairs in the last $T$ duration could be \emph{unknown}. For instance, the last observation of the pair of player 2 and game 1 (denoted by p2-g1) was in the last $T$ time duration, and therefore the labels after that time are unknown. This is known as right censoring. In contrast, the pair p3-g3 has play records every day in the last $T$ days, and it is not censored during the observation period.

Considering the existence of censored instances, we introduce a binary indicator $\delta_{uv}^{(i)}$ to indicate whether an edge $(u, v)$ is censored at timestamp $i$. $\delta_{uv}^{(i)}=0$ if $(u, v)$ is censored; $\delta_{uv}^{(i)} = 1$ otherwise. Then Inequality～(\ref{equ:temporal_dynamics}) needs to be updated by
\begin{align}\label{equ:censor}
    &f(g(\mathbf{z}_{uv}^{(i)})) \lessapprox f(g(\mathbf{z}_{uv}^{(i+1)})) \ \forall \ 0\leq i \leq t_{uv}-1, (u,v) \in D^{(i)} \\ 
    &f(g(\mathbf{z}_{uv}^{(t_{uv})}) \lessapprox f(g(\mathbf{z}_{uv}^{(i)})) \ \forall \ t_{uv}\leq i \leq t, (u,v) \in \mathcal{E}^{(t_{uv})}, \nonumber
\end{align}
where $t_{uv}$ denotes the timestamp when the edge $(u,v)$ was observed, i.e., $t_{uv} = \max\{i: \delta_{uv}^{(i)} = 1\}$. The update reflects the fact that after timestamp $t_{uv}$ the label of $(u,v)$ becomes unknown. Since the existence of edge $(u, v)$ after $t_{uv}$ is unknown, it is more reasonable to just require that Inequality~(\ref{equ:censor}) holds \emph{pairwise} between time point $t_{uv}$ and all time points after $t_{uv}$. Therefore the temporal loss can be expressed as:
\begin{align}\label{equ:temporal_loss}
    \mathcal{L}_{T} :=& \sum_{i = t_0}^{t-1} \sum_{(u,v) \in D^{(i)}}  \Big\{\|g(\mathbf{z}_{uv}^{(i+1)}) - g(\mathbf{z}_{uv}^{(i)})\|_{2} + \\ \nonumber
    &\big[\mathbbm{1}(\delta_{uv}^{(i+1)}=1)f(g(\mathbf{z}_{uv}^{(i)})) +  \mathbbm{1}(\delta_{uv}^{(i+1)}=0)f(g(\mathbf{z}_{uv}^{(t_{uv})})) - f(g(\mathbf{z}_{uv}^{(i+1)}))\big]_{+}\Big\},
\end{align}
where $\mathbbm{1}(A)$ denotes the indicator function for event $A$ and $[x]_{+} = \max\{x, 0\}$. The first term corresponds to the temporal smoothness and the second term corresponds to the temporal dynamics. Taking Equations (\ref{equ:supervised_loss}) and (\ref{equ:unsupervised_loss}) into Equation (\ref{equ:temporal_loss}), we have the temporal loss $\mathcal{L}_{T}$ as follows.
\begin{align}
    \mathcal{L}_{T} :=& \sum_{i = 0}^{t-1} \sum_{(u,v) \in D^{(i)}}  \Big\{\|h_{e}^{l_{p}}(\mathbf{z}_{uv}^{(i+1)}) - h_{e}^{l_{p}}(\mathbf{z}_{uv}^{(i)})\|_{2} + \\ \nonumber
    &\big[\mathbbm{1}(\delta_{uv}^{(i+1)}=1)\frac{\exp{\big(h_{s}^{l_{n}}(g(\mathbf{z}_{uv}^{(i)}))^{T}w_{s}}\big)}{1+\exp{\big(h_{s}^{l_{n}}(g(\mathbf{z}_{uv}^{(i)}))^{T}w_{s}}\big)} +  \\\nonumber
    &\mathbbm{1}(\delta_{uv}(i+1)=0)\frac{\exp{\big(h_{s}^{l_{n}}(g(\mathbf{z}_{uv}^{(t_{uv})}))^{T}w_{s}}\big)}{1+\exp{\big(h_{s}^{l_{n}}(g(\mathbf{z}_{uv}^{(t_{uv})}))^{T}w_{s}}\big)} - \\ \nonumber &\frac{\exp{\big(h_{s}^{l_{n}}(g(\mathbf{z}_{uv}^{(i+1)}))^{T}w_{s}}\big)}{1+\exp{\big(h_{s}^{l_{n}}(g(\mathbf{z}_{uv}^{(i+1)}))^{T}w_{s}}\big)}\big]_{+}\Big\}.
\end{align}


\subsection{Context Generation by Attributed Random Walk}
\label{subsection:attributed_random_walk}
All above discussion assumes the availability of contexts of edges. There have been a series of node-centric research on graph embedding proposing to apply random walk to sample contexts~\cite{grover2016node2vec}. These methods are normally topology-based. In our problem setting, our goal is to embed an \emph{edge}, not a node. In this case, a simple topology-based random walk may return two adjacent edges having the same player or the same game while totally ignoring the similarity of the other end. This is undesirable. In contrast, attributed random walk measures such similarity by attributes and allows to transit to similar nodes even if they are not connected.





\begin{figure} 
\centering
\includegraphics[width=.5\textwidth, height=.45\textwidth]{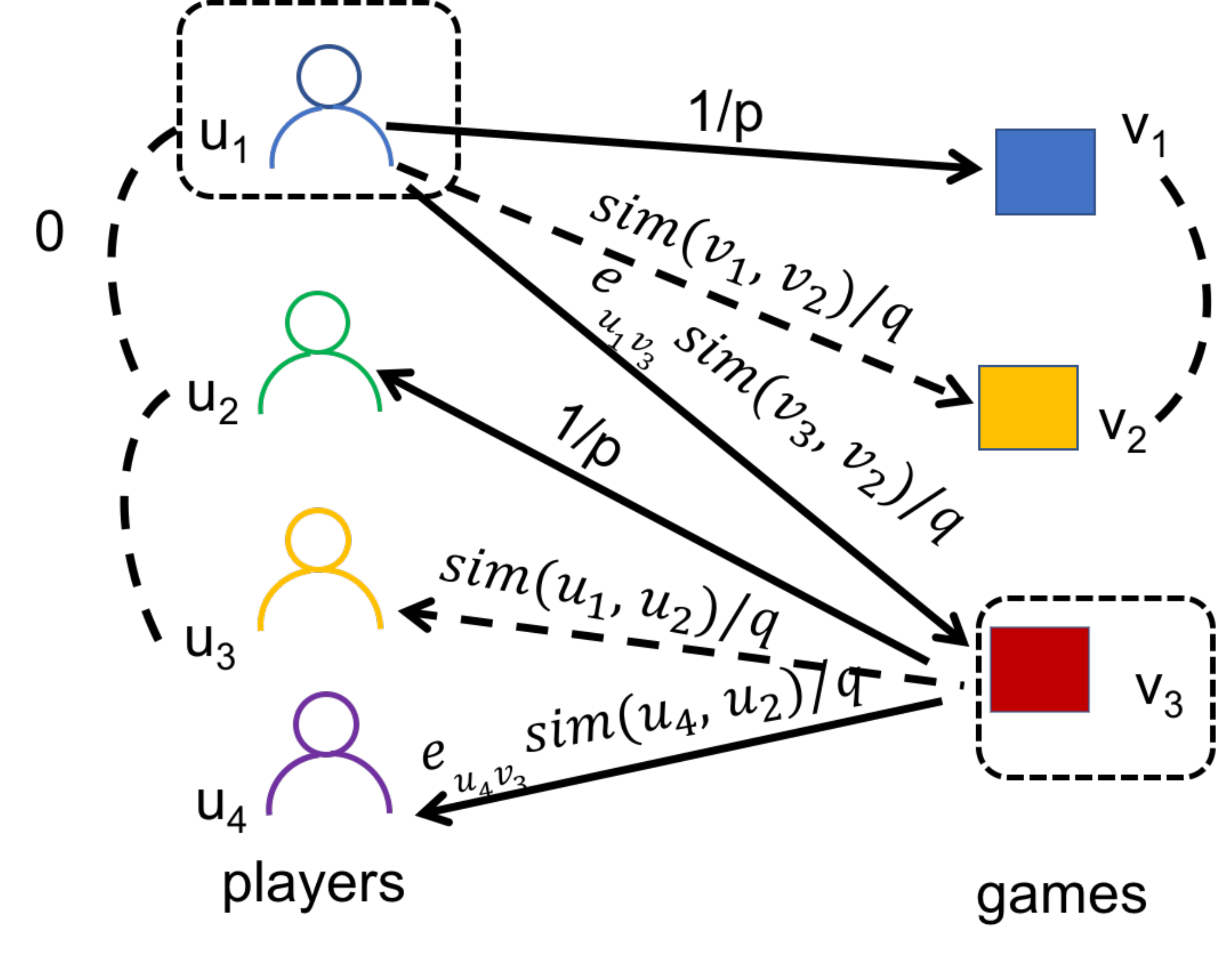}
\caption{An illustrative example of attributed random walk in an attributed bipartite graph}
\label{fig:attributed_random_walk}
\end{figure}

To this end, we propose a novel attributed random walk technique that takes into account both topological adjacency and attribute similarities to make the transition decision of the walk. Fig.~\ref{fig:attributed_random_walk} gives an illustrative example of attributed random walk in an attributed bipartite graph. For clarity, we omit the time index in the following discussion. The \emph{solid line} indicates that there exists an edge in the attributed bipartite graph. We denote the type of node $o$ by $type(o)$. A node's type can be of value either player or game. The \emph{dashed directed} lines do not exist in the original attributed graph but may be considered as transitions by our attributed random walk due to attribute similarities.

It is time-consuming to calculate pairwise similarities between a node and all other nodes with the same type. For this reason, we add augmented edges for a proportion of the same-type nodes. For example, the augmented edges of node $o_{1}$ are $\{(o_{1},o): sim(o_{1}, o) > 1 - \epsilon, type(o_{1}) = type(o)\}$, where $0<\epsilon<1$ is a filtering parameter and:
\begin{align}
    sim(o_{1}, o) := \dfrac{\mathbf{x}_{o_{1}}\cdot \mathbf{x}_{o}}{\|\mathbf{x}_{o_{1}}\| \ \|\mathbf{x}_{o}\|}.
\end{align}
The \emph{dashed undirected} lines in Fig.~\ref{fig:attributed_random_walk} represent the added augmented edges between the nodes and their similar same-typed nodes. 

Consider a random walker that just traversed edge $(v_{1}, u_{1})$ in Fig.~\ref{fig:attributed_random_walk} and now resides at node $u_{1}$. Now the walker needs to decide which node to transit to. Since attribute similarities matter in this walk, the walker cannot just evaluate those nodes that are neighbors of $u_{1}$ as suggested in~\cite{grover2016node2vec}. The walker needs to evaluate all nodes of the same type within the two-hop neighborhood of $v_{1}$. We denote the one-hop same-typed adjacent nodes of $v_{1}$ by $N_{1}(v_{1}) = \{v : d(v_{1},v) = 1, type(v) = type(v_{1})\}$, where $d(v_{1}, v)$ is the length of shortest path between $v_{1}$ and $v$. For example, $v_{2} \in N_{1}(v_{1})$ (due to the augmented edge between $v_{2}$ and $v_{1}$). Similarly, we denote the set of two-hop same-typed neighbor nodes of $v_1$ by $N_{2}(v_{1}) = \{v : d(v_{1},v) = 2, type(v) = type(v_{1})\}$. Then the transition probability in attributed random walk can be calculated as follows:
\begin{equation}
  p(o|u_{1})=\left\{
  \begin{array}{@{}ll@{}}
    0, & \text{if}\ type(o) \neq type(v_{1}) \\
    \frac{1}{p}, & \text{if} \ o = v_{1} \\
    \frac{sim(v_{1}, o)}{q}, & \text{if} \ o\in N_{1}(v_{1})\\
    \frac{sim(v_{1}, o) e_{u_{1}o}}{q}, & \text{if}\ o\in N_{2}(v_{1}) \\
  \end{array}\right.
\end{equation}
where $p$ and $q$ are normalization constants used to control the walk strategy. Recall that $e_{u_{1}o} = 1$ if there is an edge between $u_{1}$ and $o$. The attributed walk in an attributed bipartite graph walks through different types of nodes repeatedly. It enlarges the probability of making any two consecutive edges in a path similar, and thus the probability of making the whole set of edges in the attributed random walk path similar. As an example, suppose we are at $u_{1}$ from $v_{1}$ as shown in Fig.~\ref{fig:attributed_random_walk}. Since $v_{1}$ and $v_{2}$ are highly similar, the walker may transit to $v_{2}$ and produce a path with edges $(v_{1}, u_{1})$ and $(u_{1}, v_{2})$ although $u_{1}$ is not connected to $v_{2}$ in the bipartite graph.

Our proposed method shares some similarities with the idea of~\cite{ahmed2017framework}. However, in their work attribute similarities have no influence on which nodes a walker goes through. In our proposed method, when choosing the next node to visit, there is a nonzero probability to choose those that are not connected to the present node but share similar attributes to the previous node. For those that are connected to the present node, the probability to choose from them is weighted by the similarity between the descendant and the ancestor nodes. In this way, an edge can be the context of another even when they are not adjacent but similar in both ends.

\subsection{Macro-Level Churn Ranking}
\label{sec:macro-level}
The objective of macro-level churn ranking is to provide a ranked list of games based on their total numbers of users to churn in the near future. The ground truth of the number for game $v \in \mathcal{V}^{(t)}$ is $|\mathcal{N}_{v}^{(t)}\setminus \mathcal{N}_{v}^{(t+1)}|$. Let $\mathcal{S}^{(t)}(v)$ denote the score of ranking for game $v$, where a larger $\mathcal{S}^{(t)}(v)$ means a higher position. One solution is to first estimate the total number of users to churn $|\mathcal{N}_{v}^{(t)}\setminus \mathcal{N}_{v}^{(t+1)}|$ as the score $\mathcal{S}^{(t)}(v)$ for a game $v \in \mathcal{V}^{(t)}$ and then rank all games based on the estimations. After the prediction function $f$ and the embedding function $g$ are learned, the churn probability $\widehat{P}r(e_{uv}^{(t+1)} = 0 | e_{uv}^{(t)} = 1, \mathcal{H}^{(t)})$ of any individual user-game pair $(u,v)$ can be estimated by $f(g(\mathbf{z}^{(t)}_{uv}))$.  We sum up the churn probabilities across all users of game $v$ and consider the sum as the estimation of the total number of users to churn:
\begin{align}
    \mathcal{S}^{(t)}_{ss}(v) = \sum_{u\in \mathcal{U}^{(t)}} e_{uv}^{(t)}f(g(\mathbf{z}^{(t)}_{uv})).
\end{align}
where the subscript ``ss'' marks that the score is computed by the \emph{SimSum} method. 

Now we show that \emph{SimSum} provides an unbiased estimation of the ground truth $|\mathcal{N}_{v}^{(t)}\setminus \mathcal{N}_{v}^{(t+1)}|$ under the assumption that the micro-level churn probability is properly estimated. This is because we have:
\begin{align*}
    &\mathbbm{E}\big[|\mathcal{N}_{v}^{(t)}\setminus \mathcal{N}_{v}^{(t+1)}| - \mathcal{S}^{(t)}_{ss}(v)| \mathcal{H}^{(t)}\big] = \mathbbm{E}\big[|\mathcal{N}_{v}^{(t)}\setminus \mathcal{N}_{v}^{(t+1)}| \mathcal{H}^{(t)}\big] - \mathbbm{E}\big[\mathcal{S}^{(t)}_{ss}(v)| \mathcal{H}^{(t)}] \\ \nonumber 
    &=  \mathbbm{E}\big[\sum_{u\in \mathcal{U}^{(t)}}\mathbbm{1}(e_{uv}^{(t)}=1) \mathbbm{1}(e_{uv}^{(t+1)}=0)| \mathcal{H}^{(t)}\big] - \sum_{u\in \mathcal{U}^{(t)}} e_{uv}^{(t)}f(g(\mathbf{z}^{(t)}_{uv})),
\end{align*}
and $\mathcal{H}^{(t)} = \{\mathcal{G}^{(i)}\}_{i = t_0}^{i=t}$, then
\begin{align}\label{eq:unbiased}
    &\mathbbm{E}\big[|\mathcal{N}_{v}^{(t)}\setminus \mathcal{N}_{v}^{(t+1)}| - \mathcal{S}^{(t)}_{ss}(v)| \mathcal{H}^{(t)}\big]  \\ \nonumber 
    & =\sum_{u\in \mathcal{U}^{(t)}}e_{uv}^{(t)}\mathbbm{E}\big[\mathbbm{1}(e_{uv}^{(t+1)}=0)| \mathcal{H}^{(t)},e_{uv}^{(t+1)}=0\big] - \sum_{u\in \mathcal{U}^{(t)}} e_{uv}^{(t)}f(g(\mathbf{z}^{(t)}_{uv}))  \\ \nonumber
    &= \sum_{u\in \mathcal{U}^{(t)}} e_{uv}^{(t)}\big(Pr((e_{uv}^{(t+1)} = 0 | e_{uv}^{(t)} = 1, \mathcal{H}^{(t)}))-\widehat{P}r(e_{uv}^{(t+1)} = 0 | e_{uv}^{(t)} = 1, \mathcal{H}^{(t)})\big).
\end{align}
Since the right hand side of Equation~\ref{eq:unbiased} is $0$, we complete the proof.


Furthermore, we observe that there may exist correlation of the churn behavior between different games and users. 
As such, we employ link analysis algorithms, which are able to take into consideration the global information and mutual reinforcing effects among different games and users, for inferring the macro-level churn of games. In particular, we choose \emph{PageRank}~\cite{page1999pagerank} and \emph{HITS}~\cite{kleinberg1999authoritative} algorithms due to their wide application and stable performance. To apply both algorithms, we first construct a user-game relation graph $\mathcal{G}^{(t)}_{r} = \{\mathcal{V}^{(t)}, \mathcal{E}^{(t)}, \mathcal{W}^{(t)}_{r}\}$, where $\mathcal{V}^{(t)}$ and $\mathcal{E}^{(t)}$ are copied from the attributed graph $\mathcal{G}^{(t)}$ and $\mathcal{W}_{r}^{(t)}$ is an adjacency matrix with the element at $u$-th row and $v$-th column satisfying
\begin{align}
    \mathcal{W}_{r}^{(t)}(u,v) = f(g(\mathbf{z}^{(t)}_{uv})).
\end{align}
This relation graph will be the input of both algorithms. The ranking scores computed by \emph{PageRank} and \emph{HITS} are referred to as $\mathcal{S}_{pg}^{(t)}$ and  $\mathcal{S}_{hits}^{(t)}$, respectively. 

\begin{algorithm}
  \KwIn{Graph $\mathcal{G}^{(t)}_{r}$, maximal number of iterations $M$, damping factor $d_{\alpha}$}
  \KwOut{$\mathcal{S}_{pg}^{(t)}(v)$ for each $v\in \mathcal{V}^{(t)}$ ranking scores of games}
  \BlankLine
  $k = 0$, initialize $\mathcal{S}_{pg}^{(t)}(i)= 1/|\mathcal{V}^{(t)}\cup \mathcal{U}^{(t)}|$ for $i\in \mathcal{V}^{(t)}\cup \mathcal{U}^{(t)}$;\\
  \While{ $k \leq M$}{
  Copy last iteration scores to $\mathcal{T}^{(t)}$: $\mathcal{T}^{(t)}(i)= \mathcal{S}_{pg}^{(t)}(i)$ for $i\in \mathcal{V}^{(t)}\cup \mathcal{U}^{(t)}$; \\
    \For {$i \in \mathcal{V}^{(t)}\cup \mathcal{U}^{(t)}$}{
        $\mathcal{S}_{pg}^{(t)}(i) = \dfrac{1-d_{\alpha}}{|\mathcal{V}^{(t)}\cup \mathcal{U}^{(t)}|} + d_{\alpha}\times\sum_{j \in \mathcal{N}_{i}^{(t)}} \mathcal{T}^{(t)}(j) \times \dfrac{\mathcal{W}_{r}^{(t)}(i,j)}{\sum_{\ell \in \mathcal{N}_{i}^{(t)}}\mathcal{W}_{r}^{(t)}(i,\ell)}$
    }
  }
  \Return{ $\mathcal{S}^{(t)}_{pg}(v)$ for $v \in \mathcal{V}^{(t)}$
  }
\caption{\emph{PageRank} Based Macro-Level Churn Ranking}\label{alg:pagrank}
\end{algorithm}

The procedures to compute $\mathcal{S}_{pg}^{(t)}$ is given in Algorithm~\ref{alg:pagrank}. The algorithm assigns a larger ranking score to a game if the game has more about-to-churn players with larger scores. Moreover, the score will be propagated along with a damping factor from one game to another game. The algorithm handles different players differently for a specific game. In the algorithm, players with distinct propensities of churn contribute differently to the computation of the game's ranking score. Considering that $\mathcal{G}_{r}^{(t)}$ is a bipartite graph, in each iteration, the algorithm computes a game's ranking score by summing up its players' scores. The sum is weighted by the churn probabilities, where the players with larger churn probabilities are more likely to contribute greater proportions to the game's ranking score. Meanwhile, this preponderance is normalized by the total number of games the player is playing. The intuition is that the algorithm reduces the proportion if the player is more probable to churn other games at this time point as a player is rarely observed to churn all games at a single time point in experiments. In this way, the correlation between game churn for a specific player is addressed.

\begin{algorithm}
  \KwIn{Graph $\mathcal{G}^{(t)}_{r}$, maximal number of iterations $M$}
  \KwOut{$\mathcal{S}_{hits}^{(t)}(v)$ for each $v\in \mathcal{V}^{(t)}$ ranking scores of games}
  \BlankLine
  $k = 0$, initialize $\mathcal{A}^{(t)}_{hits}(i) = \mathcal{H}^{(t)}_{hits}(i) = 1$ for $i\in \mathcal{V}^{(t)}\cup \mathcal{U}^{(t)}$;\\
  \While{ $k \leq M$}{
  Copy last iteration scores to $\mathcal{T}^{(t)}$: $\mathcal{T}^{(t)}(i)= \mathcal{A}_{hits}^{(t)}(i)$ for $i\in \mathcal{V}^{(t)}\cup \mathcal{U}^{(t)}$; \\
    \For {$i \in \mathcal{V}^{(t)}\cup \mathcal{U}^{(t)}$}{
        $\mathcal{A}_{hits}^{(t)}(i) = \sum_{j \in \mathcal{N}_{i}^{(t)}} \mathcal{W}_{r}^{(t)}(i,j) \mathcal{H}_{hits}^{(t)}(j)$ \\
        $\mathcal{H}_{hits}^{(t)}(i) = \sum_{j \in \mathcal{N}_{i}^{(t)}} \mathcal{W}_{r}^{(t)}(i,j) \mathcal{T}^{(t)}(j)$
    }
    {
    Normalize $\{\mathcal{A}_{hits}^{(t)}(i)\}_{i \in \mathcal{V}^{(t)}\cup \mathcal{U}^{(t)}}$ and $\{\mathcal{H}_{hits}^{(t)}(i)\}_{i \in \mathcal{V}^{(t)}\cup \mathcal{U}^{(t)}}$
    }
  }
  \Return{ $\mathcal{A}^{(t)}_{hits}(v)$ for $v \in \mathcal{V}^{(t)}$
  }
\caption{\emph{HITS} Based Macro-Level Churn Ranking}\label{alg:hits}
\end{algorithm}

The procedures to compute $\mathcal{S}_{hits}^{(t)}$ is presented in Algorithm~\ref{alg:hits}. The algorithm uses the final authority score $\mathcal{A}^{(t)}_{hits}(v)$ as the ranking score $\mathcal{S}_{hits}^{(t)}(v)$ for game $v$. Since the graph $\mathcal{G}_{r}^{(t)}$ is a bipartite graph composed of players and games, a larger authority score is assigned to a game if the game has more players with large hub scores. Similarly, a player has a large hub score when the player is playing many games with large authority scores. In other words, the hub score for a player in some sense describes the chance that a churn will happen with the player. And the authority score of a game to some extent captures the number of its players with high churn propensity. Meanwhile, for a specific game, the algorithm handles different players differently, and for a specific player, different games contribute different proportions. This is reflected by Lines $5-6$ of Algorithm~\ref{alg:hits}, where the computation of the authority score and the hub score are weighted by the estimated probability of churn. 

\section{Experimental Evaluation}
\label{sec:experiment}

In this section, we conduct a comprehensive experimental evaluation over the large-scale real data collected from the Samsung Game Launcher platform. We compare our semi-supervised model (referred to as {\tt SS} in the sequel) with the following state-of-the-art models for mobile game churn prediction:
\begin{itemize}
    \item {\tt LR}: the logistic regression based solution used in~\cite{kim2017churn,xie2016predicting, xie2015predicting}
    \item {\tt RS}: the supervised variant of our model, in which the loss function contains only the supervised component and the regularization term
    \item {\tt DT}: a decision tree based solution
    \item {\tt RF}: the random forests based solution used in~\cite{kim2017churn,xie2015predicting}
    \item {\tt SVM}: the SVM based solution used in~\cite{xie2016predicting,xie2015predicting}.
\end{itemize}

In the experiments, we consider the churn duration $T=14$, but again the proposed solution is not restricted to any particular choice of $T$.

\subsection{Dataset and Feature/Label Construction}
\label{subsec:dataset}

Two anonymous datasets were collected \emph{independently} from the Samsung Game Launcher platform within a 4-month period (from August 1st, 2017 to November 30th, 2017) with users' consent, one from the users in \emph{USA} and the other from the users in \emph{Korea}. We summarize the key statistics of these two datasets in Table~\ref{table:data}.

\begin{table}
\centering
\caption{Dataset statistics}
\begin{tabular}{|c|c|c|c|}
\hline
\textbf{Dataset}   & \textbf{\# of users}   & \textbf{\# of games} & \textbf{\# of play records} \\ \hline
\emph{USA}       & 15,000      & 19,705      & 76,468,301 \\ \hline
\emph{Korea}    & 25,000      & 18,470      & 106,544,313   \\ \hline
\end{tabular}
\label{table:data}
\end{table}

The collected data contains three major types of information: (1) play history, (2) game profiles, and (3) user information. Each play record in the play history contains the anonymous user id, the game package name, and the timestamp of play. It is also accompanied with rich contextual information, such as WiFi connection status, screen brightness, audio volume, etc. Game profiles are collected from different game stores, which include features like genre, developer, number of downloads, rating values, number of ratings, etc. User information contains the device model, region, OS version, etc. However, data within games (e.g., levels) is not available due to privacy concerns.

Features and labels need to be generated with care in order to avoid data leakage. In general, labels and features are taken from disjoint periods to avoid data leakage. The semi-supervised model is trained based on features and labels within the observation period. Labels are whether a player churns a game on that day. Features are constructed from historical data before the day to be predicted. The training set and test set are split by label days in chronological order~\cite{chamberlain2017customer}, for example, taking the labeled data in the first 2/3 of the observation period as the training set and the remaining 1/3 as the test set. This is also to ensure that there is a time difference between the test set and the training set. The semi-supervised model along with the \emph{SimSum}, \emph{PageRank} and \emph{HITS} methods are evaluated on the test datasets.

\subsection{Experimental Settings} 
We tune the hyperparameters, including learning rate, batch size, regularization terms, number of layers and number of neurons per layer, based on the model performance on the test datasets. To determine the parameters for micro-level churn prediction, we follow what we did in our previous work~\cite{liu2018semi}. A grid search on these parameters is performed and the combination yielding the best performance is chosen. The regularization parameters $\{\lambda_{i}\}_{i=0}^{4}$ are all set to be 1.  $\alpha$, $\beta$, and $\gamma$ in Equation (\ref{equ:objective}) are chosen to be 0.02, 0.01, 1e-5, respectively. $\epsilon$, $p$, and $q$ discussed in Section~\ref{subsection:attributed_random_walk} are set to be 1, 1, and 0.05, respectively. The parameters used for training the deep neural network are summarized below. The maximal number of iterations $M$ for the \emph{PageRank} and \emph{HITS} methods is set to be $100$. The damping factor for \emph{PageRank} is set to be $0.85$. We summarize the parameter settings of the two datasets below.

\vspace{3mm}
\noindent\textbf{Parameter settings of the \emph{USA} dataset}:
\begin{itemize}
    \item Player feature dimension: 10,042
    \item Game feature dimension: 10,042
    \item Player-game feature dimension: 30
    \item Learning rate: the initial value is 0.017 and decay by $\eta = \eta_0/ (1 + k/2)$, where $k$ is the number of epochs 
    \item Number of neurons: input layers 30, embedding layers 50, output layers 380K
    \item Number of epochs: 6-8
    \item Batch size: 1,024
    \item Context number per user-game pair: 4
    \item Optimizer: Adam method~\cite{kingma2014adam}
    \item Activation function: rectified linear unit (ReLU)
\end{itemize}

\noindent\textbf{Parameter settings of the \emph{Korea} dataset}:
\begin{itemize}
    \item Player feature dimension: 10,042
    \item Game feature dimension: 10,042
    \item Player-game feature dimension: 30
    \item Learning rate: the initial value is 0.019 and decay by $\eta = \eta_0/ (1 + k/2)$, where $k$ is the number of epochs 
    \item Number of neurons: input layers 30, embedding layers 50, output layers 632K
    \item Number of epochs: 8-12
    \item Batch size: 4,096
    \item Context number per user-game pair: 4
    \item Optimizer: Adam method~\cite{kingma2014adam}
    \item Activation function: rectified linear unit (ReLU)
\end{itemize}


\begin{table}
\centering
\caption{Performance of different micro-level churn prediction models on \emph{USA}}
\begin{tabular}{|p{1.8cm}|p{1.5cm}|p{1.5cm}|p{1.5cm}|}
\hline
Model & AUC & Recall & Precision
\\ \hline
{\tt SS} & \textbf{\underline{0.82}} &  \textbf{\underline{0.78}} & 0.32
\\ \hline
{\tt RS} & 0.77 & 0.75 & 0.27
\\ \hline
{\tt LR} & 0.66 & 0.38 & 0.26
\\  \hline
{\tt DT} & 0.59 & 0.28 & 0.32
\\ \hline
{\tt RF} & 0.75 & 0.31 & \textbf{\underline{0.41}} 
\\ \hline
{\tt SVM} & 0.61 & 0.78 & 0.18
\\ \hline
\end{tabular}
\label{table:performanceusa}
\end{table}

\begin{table}
\centering
\caption{Performance of different micro-level churn models on \emph{Korea}}
\begin{tabular}{|p{1.8cm}|p{1.5cm}|p{1.5cm}|p{1.5cm}|}
\hline
Model & AUC & Recall & Precision
\\ \hline
{\tt SS} & \textbf{\underline{0.82}} &  \textbf{\underline{0.70}} & 0.34
\\ \hline
{\tt RS} & 0.76 & 0.70 & 0.25
\\ \hline
{\tt LR} & 0.67 & 0.59 & 0.21
\\  \hline
{\tt DT} & 0.58 & 0.26  & 0.30
\\ \hline
{\tt RF} & 0.73 & 0.27 & \textbf{\underline{0.42}} 
\\ \hline
{\tt SVM} & 0.63  &  0.67 & 0.18
\\ \hline
\end{tabular}
\label{table:performancekorea}
\end{table}

\subsection{Experimental Results}
\subsubsection{Micro-Level Churn Prediction}
We use three widely-used evaluation metrics to compare the performance of different micro-level churn prediction models. The most important metric with respect to the business goals is \emph{the area under the ROC curve} (AUC). Following previous studies~\cite{kim2017churn,xie2016predicting, xie2015predicting}, we also consider \emph{precision} and \emph{recall}. Accuracy is not used because our data is imbalanced with around $85\%$ negative instances in the \emph{USA} dataset and $86\%$ negative instances in the \emph{Korea} dataset. 

\begin{figure}
    \centering
    \begin{subfigure}{0.47\textwidth}
    \includegraphics[width=\textwidth]{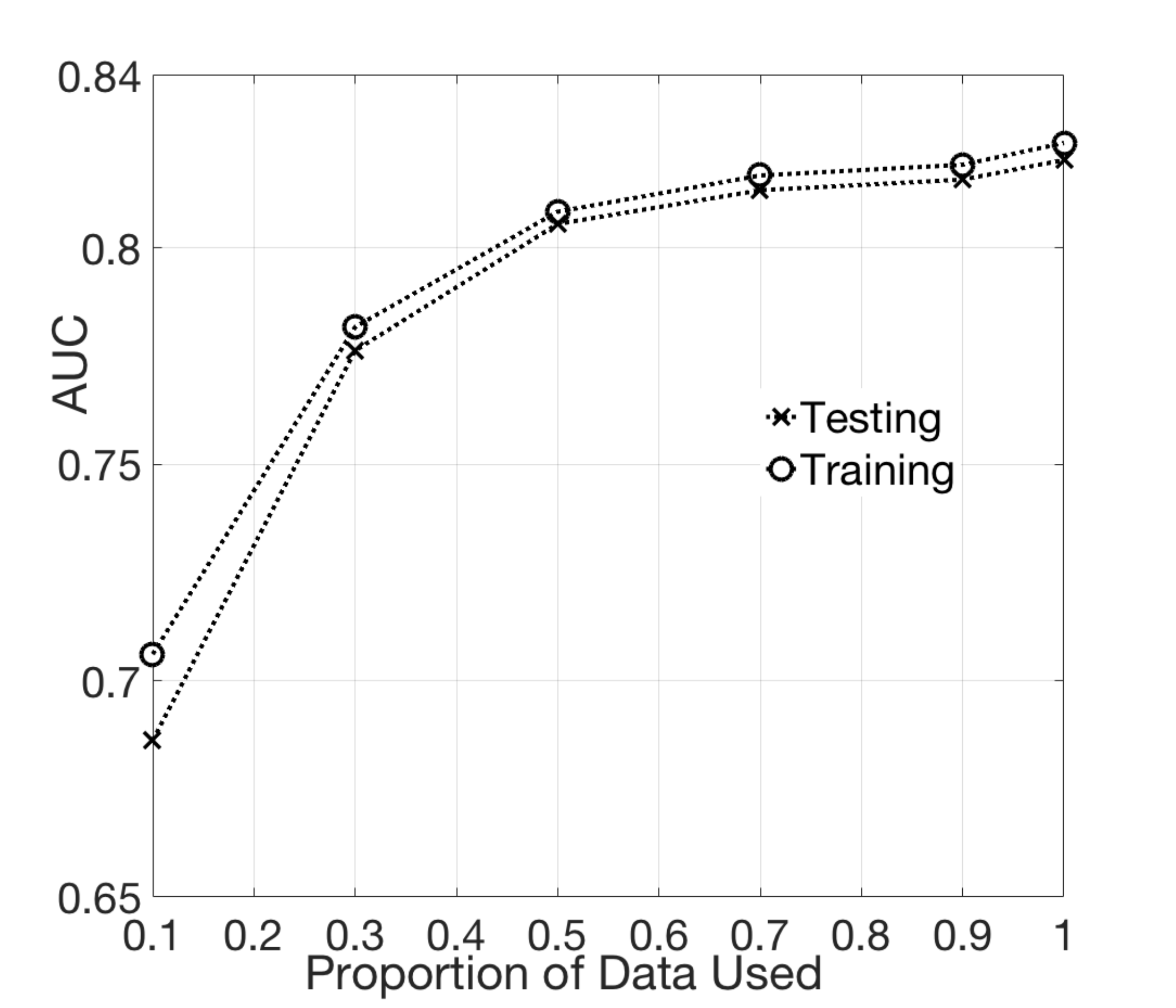} 
    \caption{AUC on \emph{USA}}
    \end{subfigure}
    \begin{subfigure}{0.47\textwidth}
    \includegraphics[width=\textwidth]{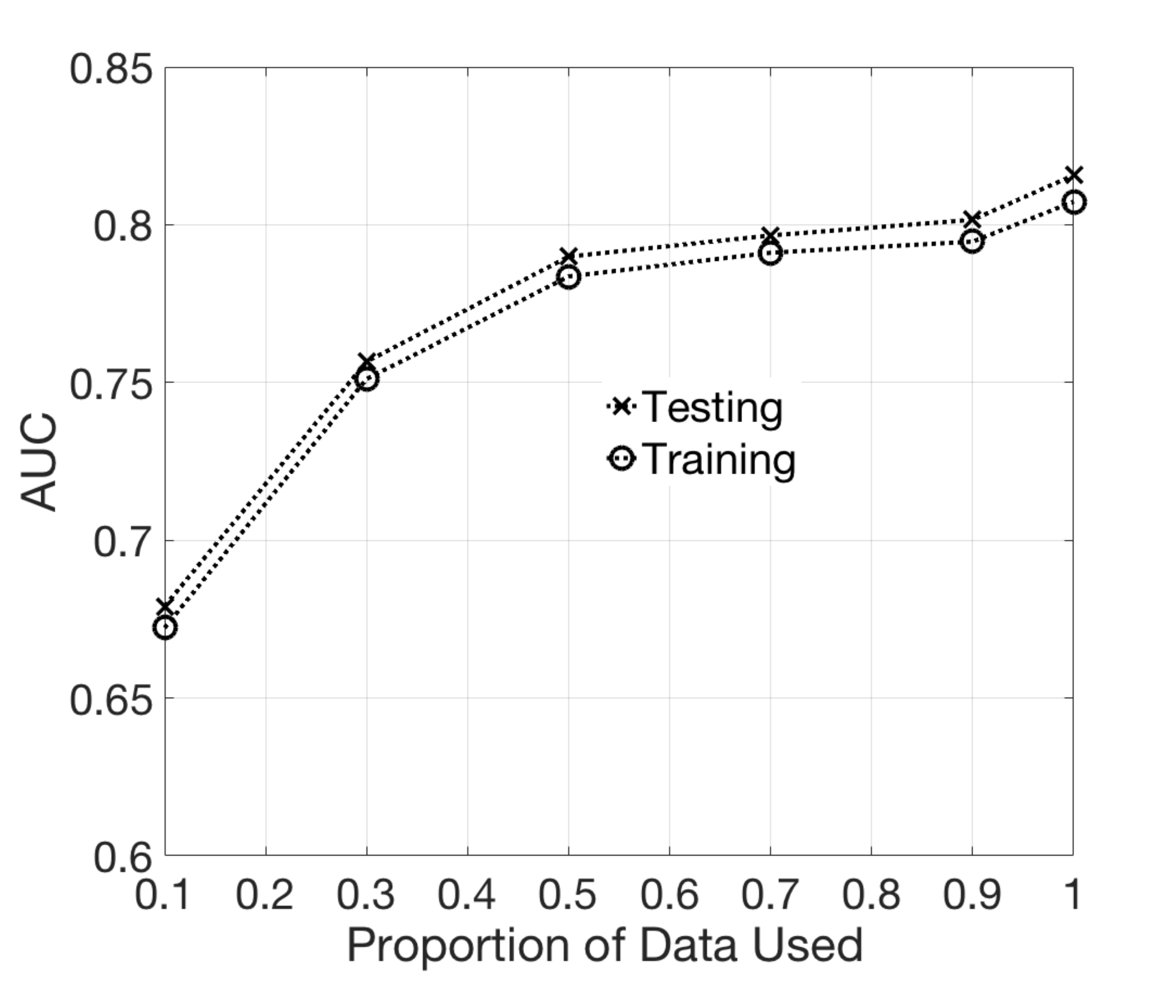}
    \caption{AUC on \emph{Korea}}
    \end{subfigure}
\caption{Comparison of AUC for {\tt SS} in training and testing}
\label{fig:training_testing}
\end{figure}

We report the main experimental results in Table~\ref{table:performanceusa} and Table~\ref{table:performancekorea}. It can be observed that in general our model achieves the best AUC and recall on both datasets. Our model outperforms all single models (i.e., {\tt LR}, {\tt DT} and {\tt SVM}) in terms of all the three metrics. In particular, it is worth mentioning that {\tt SVM} achieves a high recall at the cost of a very low precision. This is because it makes a large number of false positives. This fact makes it less useful for business decision making. Compared with the ensemble method {\tt RF}, which has been considered so far the best method in the field, our model still achieves $34\%$ AUC improvement and $250\%$ recall improvement. {\tt RF} achieves the highest precision on the datasets. However, we point out that this number is actually misleading because it can easily overfit and only recognize a small proportion of churn labels. This is partially evidenced by its poor recall, which makes it difficult to meet business requirements of churn prediction (e.g., targeted promotion campaigns). We also carefully compare the AUC of {\tt SS} in training and testing in Fig.~\ref{fig:training_testing}. Since the curves are very close, it can be learned that our model is neither overfitting nor underfitting.

\begin{figure}
    \centering
    \begin{subfigure}{0.47\textwidth}
    \includegraphics[width=\textwidth]{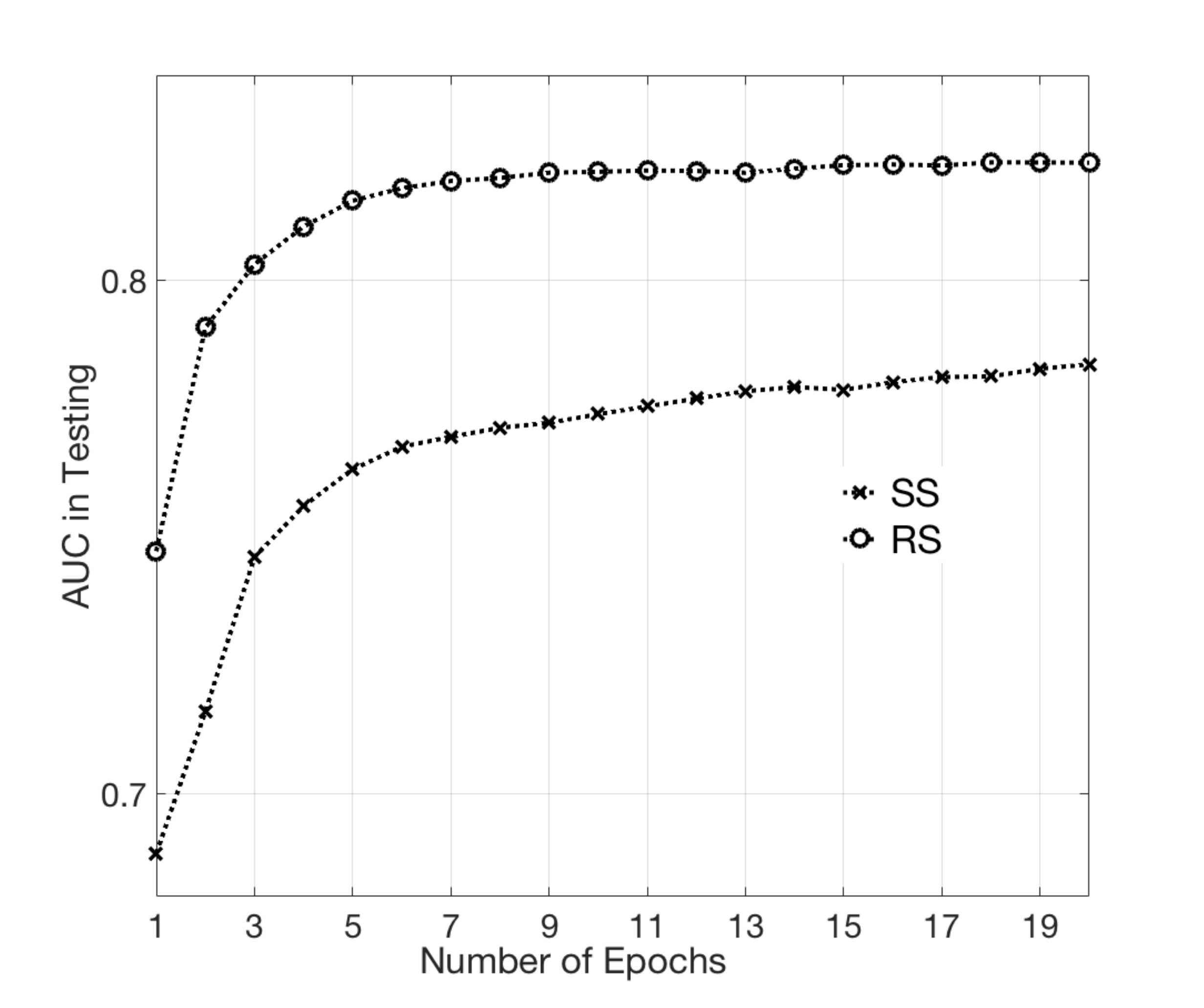} 
    \caption{AUC on \emph{USA}}
    \end{subfigure}
    \begin{subfigure}{0.47\textwidth}
    \includegraphics[width=\textwidth]{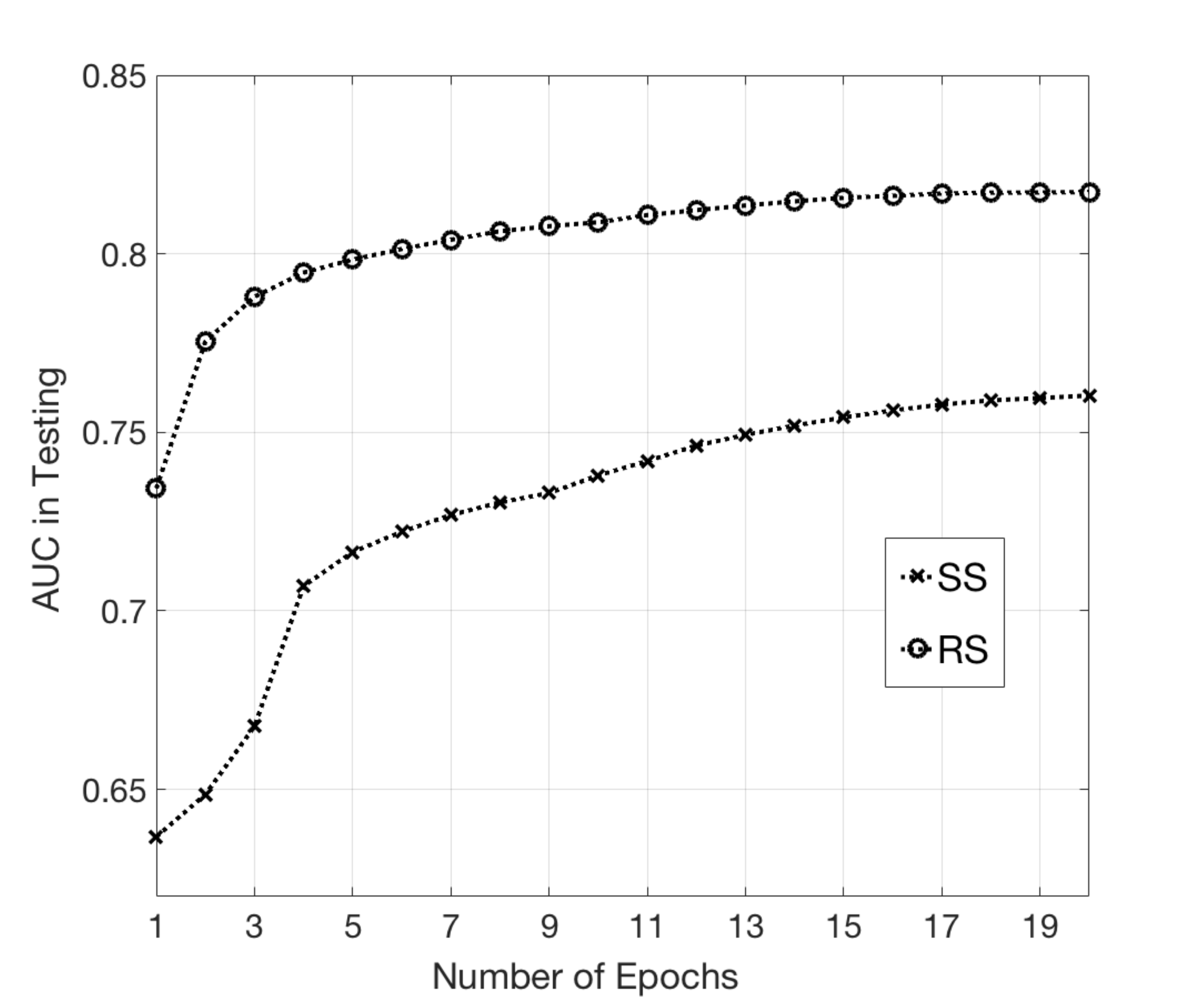}
    \caption{AUC on \emph{Korea}}
    \end{subfigure}
\caption{AUC comparison between {\tt SS} and {\tt RS} under different numbers of epochs}
\label{fig:auc_ss_rs}
\end{figure}

The performance difference between {\tt SS} and {\tt RS} justifies the benefits of incorporating unsupervised loss and temporal loss in the objective function. We provide a further comparison between {\tt SS} and {\tt RS} with respect to the number of epochs in Fig.~\ref{fig:auc_ss_rs}. Both models take 5-7 epochs to reach relatively stable performance. We observe that {\tt SS} outperforms {\tt RS} in general under different numbers of epochs and for both Korean users and USA users. 

\begin{figure}
    \centering
    \begin{subfigure}{0.47\textwidth}
    \includegraphics[width=\textwidth]{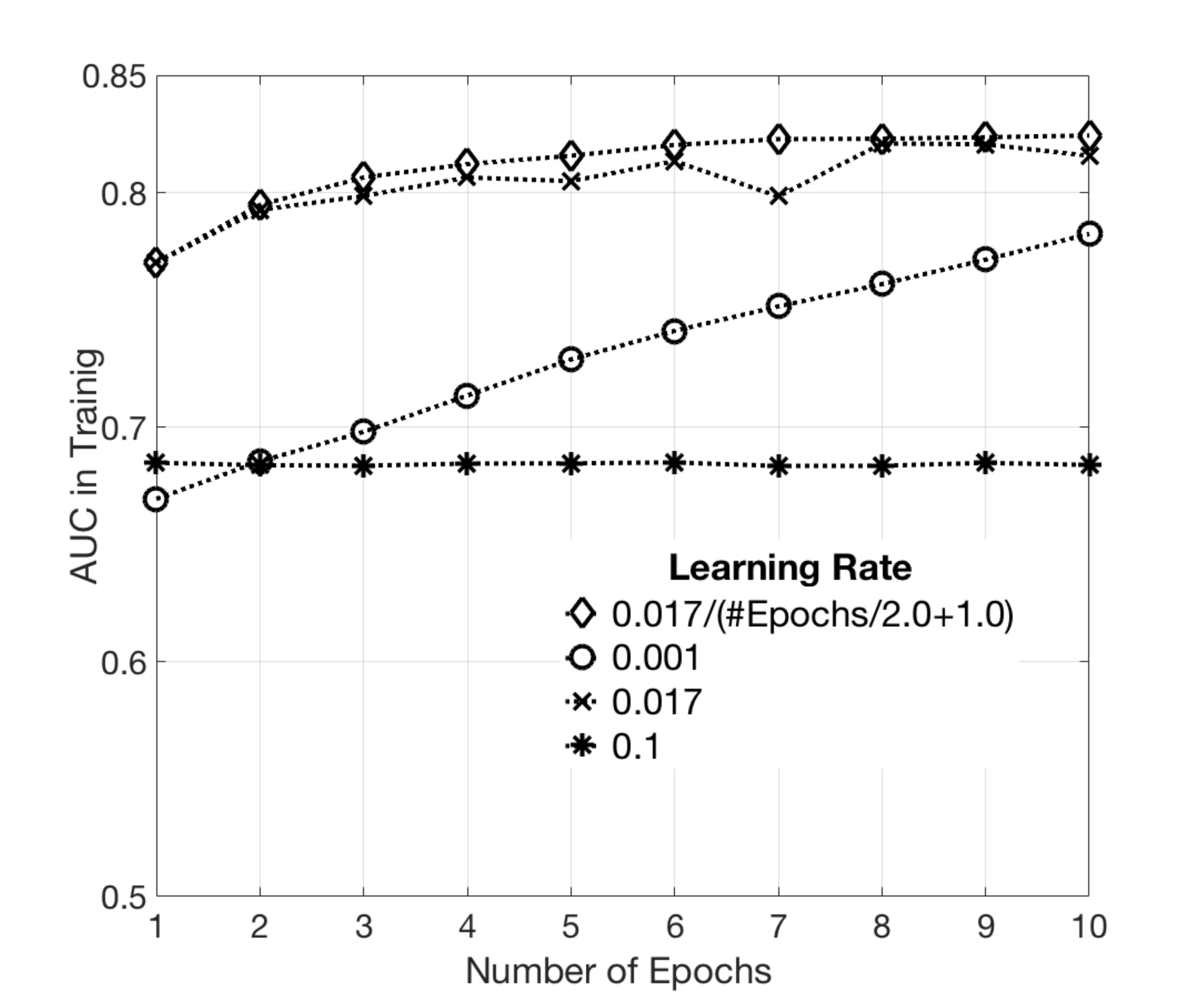} 
    \caption{AUC on \emph{USA}}
    \end{subfigure}
    \begin{subfigure}{0.47\textwidth}
    \includegraphics[width=\textwidth]{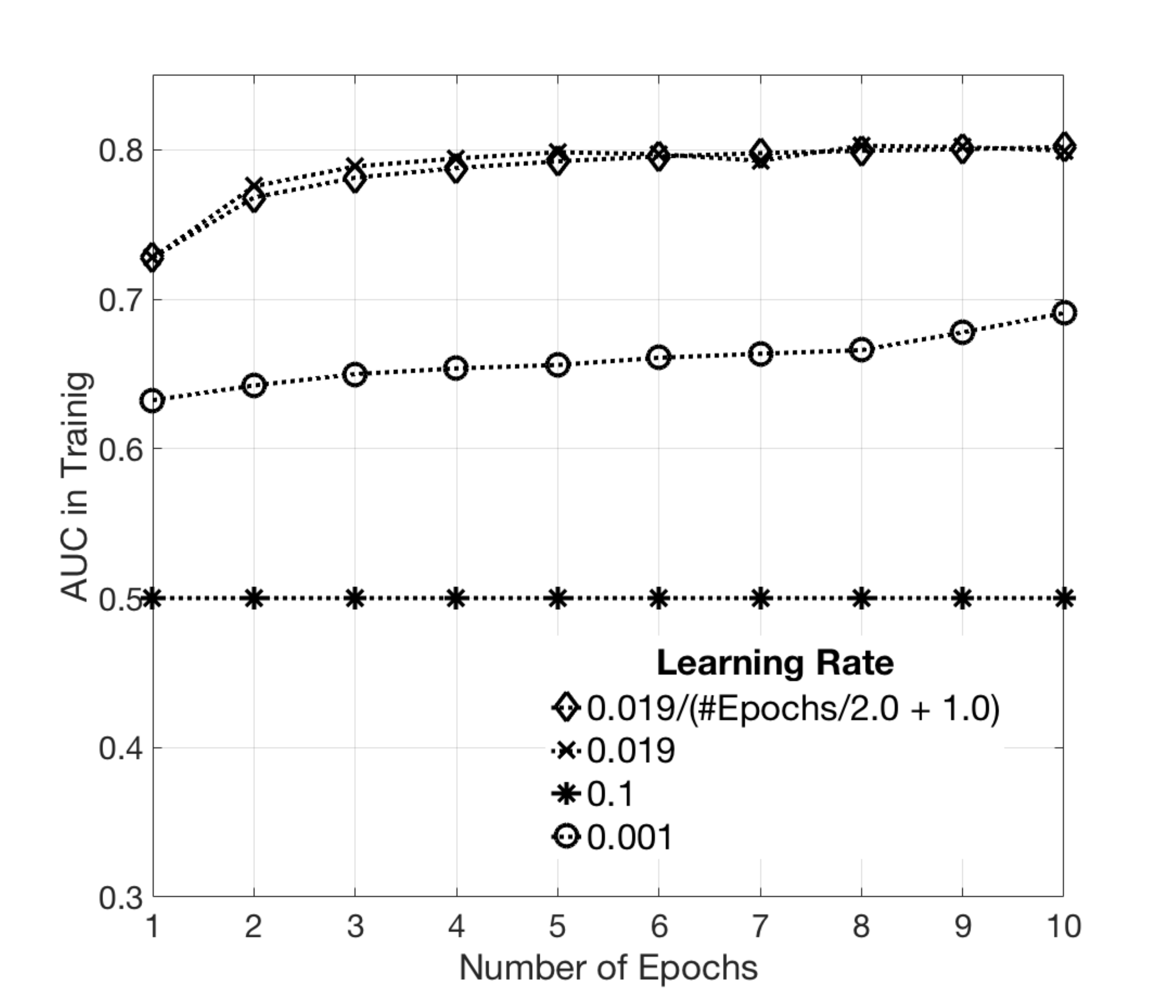}
    \caption{AUC on \emph{Korea}}
    \end{subfigure}
\caption{Comparison of AUC for {\tt SS} in training under different learning rates}
\label{fig:learning_rate}
\end{figure}

\begin{figure}
    \centering
    \begin{subfigure}{0.47\textwidth}
    \includegraphics[width=\textwidth]{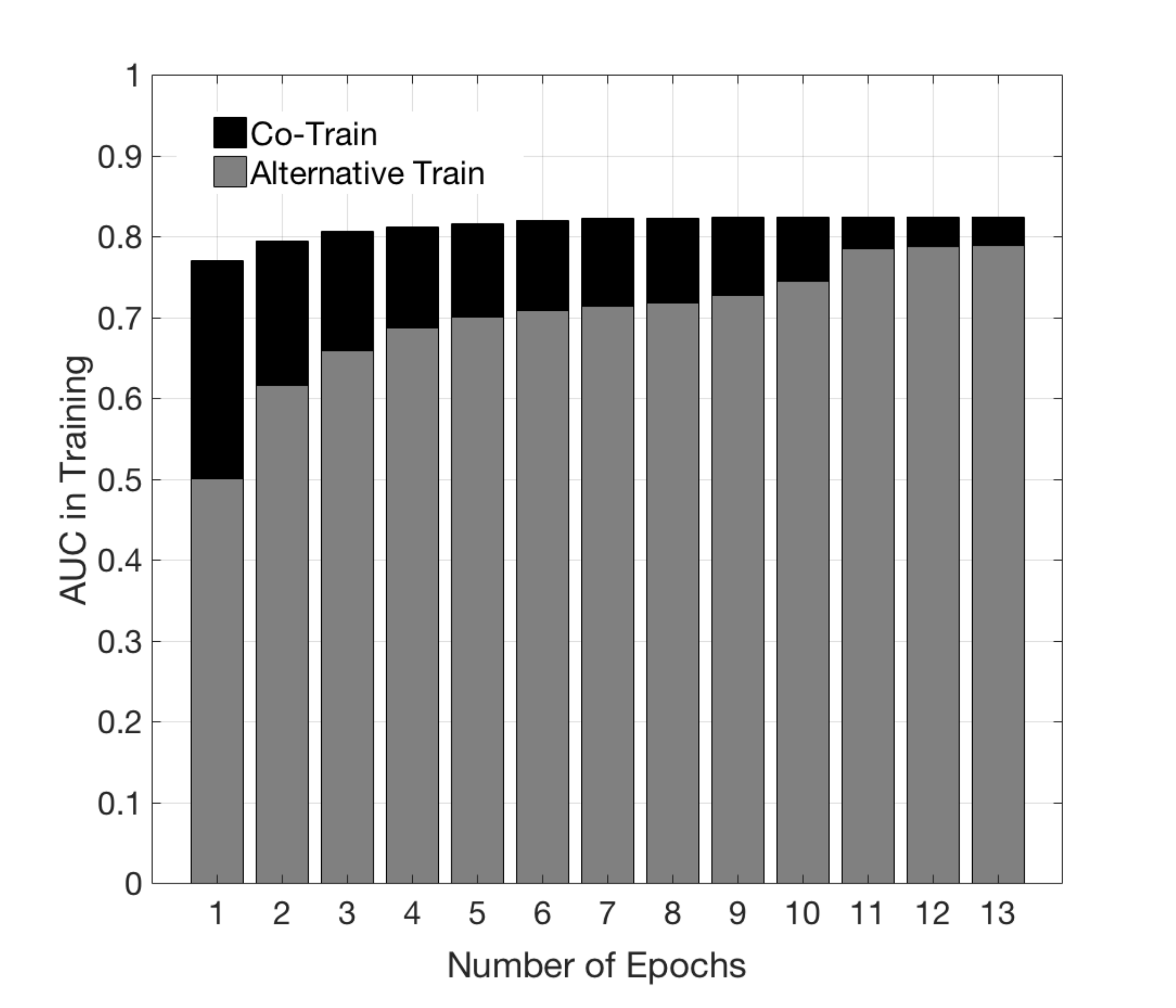} 
    \caption{AUC on \emph{USA}}
    \end{subfigure}
    \begin{subfigure}{0.47\textwidth}
    \includegraphics[width=\textwidth]{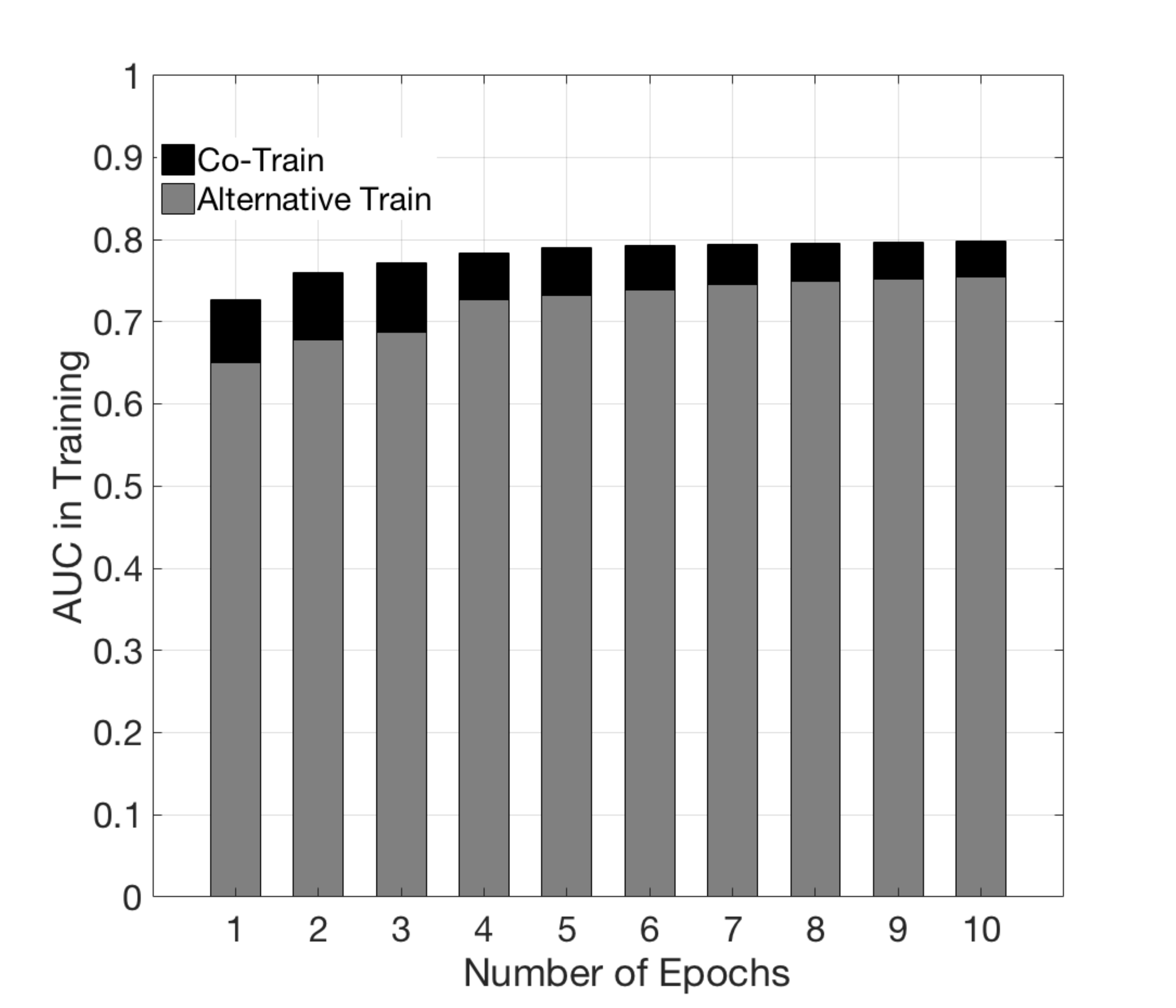}
    \caption{AUC on \emph{Korea}}
    \end{subfigure}
\caption{Comparison of AUC for {\tt SS} in training under different training methods}
\label{fig:training_method}
\end{figure}

Since the architecture of our DNN is novel and unique (e.g., contain both supervised outputs and unsupervised outputs), we expose more details on how we choose the parameters and train the model. A comparison of AUC in training under different learning rates is given in Fig.~\ref{fig:learning_rate}. It can be observed that the choice of the learning rate greatly influences the model performance after the initial epoch. We experimentally find that $0.1$ is too large for the learning rate, which makes the step in gradient descent too large to find a good minimum and that $0.001$ is too small, making it converge very slowly to the optimal point. Therefore we experimentally test learning rates between $0.1$ and $0.001$ and find that $0.017/\text{(\#Epochs/2+1)}$ works best for training on \emph{USA} while $0.019/\text{(\#Epochs/2+1)}$ works best for training on \emph{Korea}.

For the supervised and unsupervised components, we try two different training methods: \emph{co-train} and \emph{alternative train}. Co-train means that we simultaneously train the supervised loss function and the unsupervised loss function; alternative train means that we alternately train the unsupervised component with the unsupervised loss function and the supervised component with the supervised loss function. Alternative train is a widely-used training method for similar structures~\cite{yang2016revisiting, liang2017seano}. It is interesting to observe that, however, co-train outperforms alternative train in terms of AUC under different numbers of epochs as shown in Fig.~\ref{fig:training_method}. Therefore, we choose co-train as the final training method in our experiments.

\subsubsection{Macro-Level Churn Ranking}
We use five widely-used evaluation metrics to compare the performance of different macro-level churn ranking methods based on different micro-level churn prediction methods, including \emph{Kendall's Tau correlation coefficient}, \emph{weighted Kendall's Tau correlation coefficient}, \emph{Spearman correlation coefficient}, \emph{Average Precision at K} and \emph{Mean Average Precision (MAP)}. 

\begin{figure}
    \centering
    \begin{subfigure}{0.47\textwidth}
    \includegraphics[width=\textwidth]{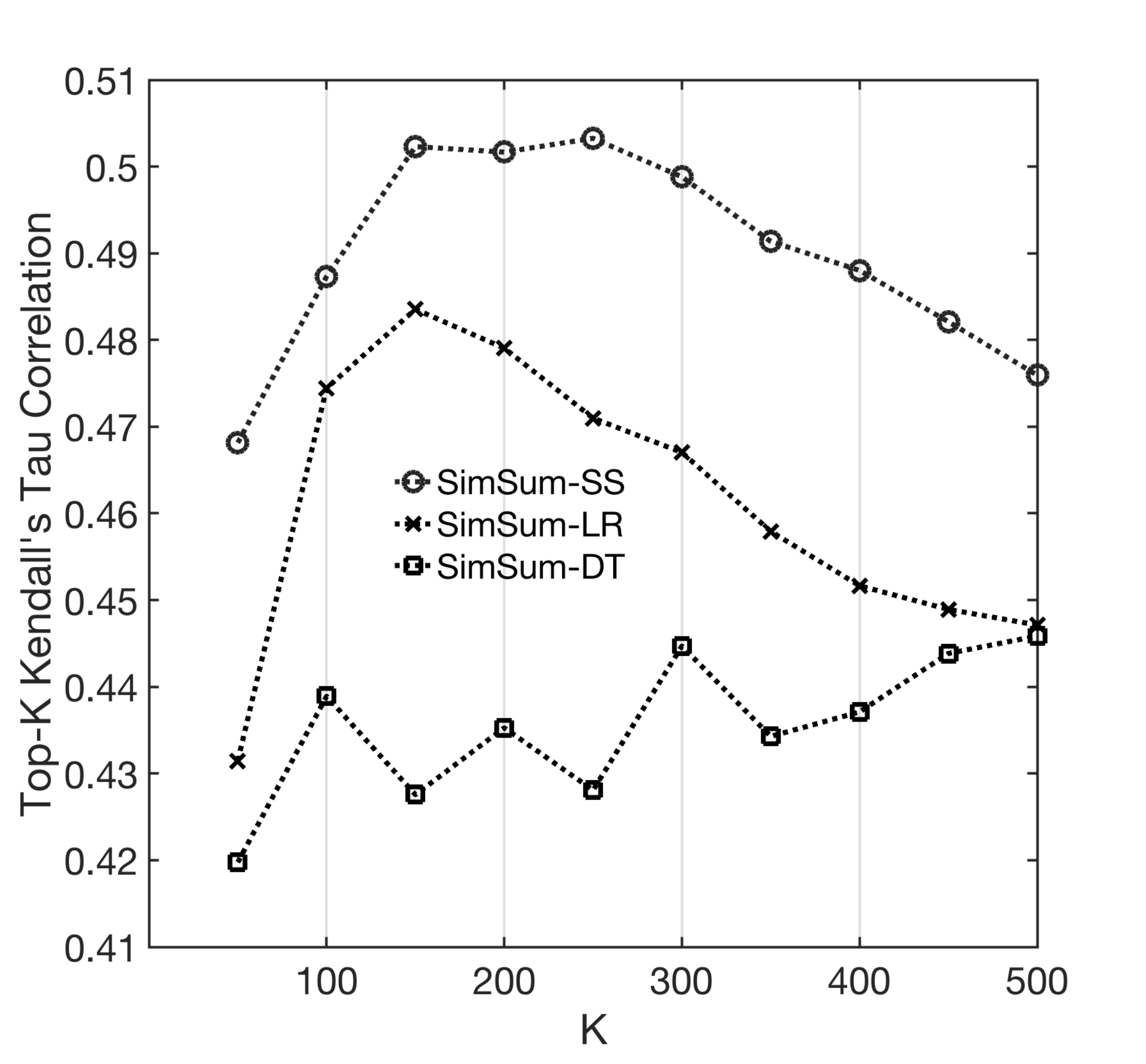} 
    \caption{\emph{SimSum} Kendall's Tau on \emph{USA}}
    \end{subfigure}
    \begin{subfigure}{0.47\textwidth}
    \includegraphics[width=\textwidth]{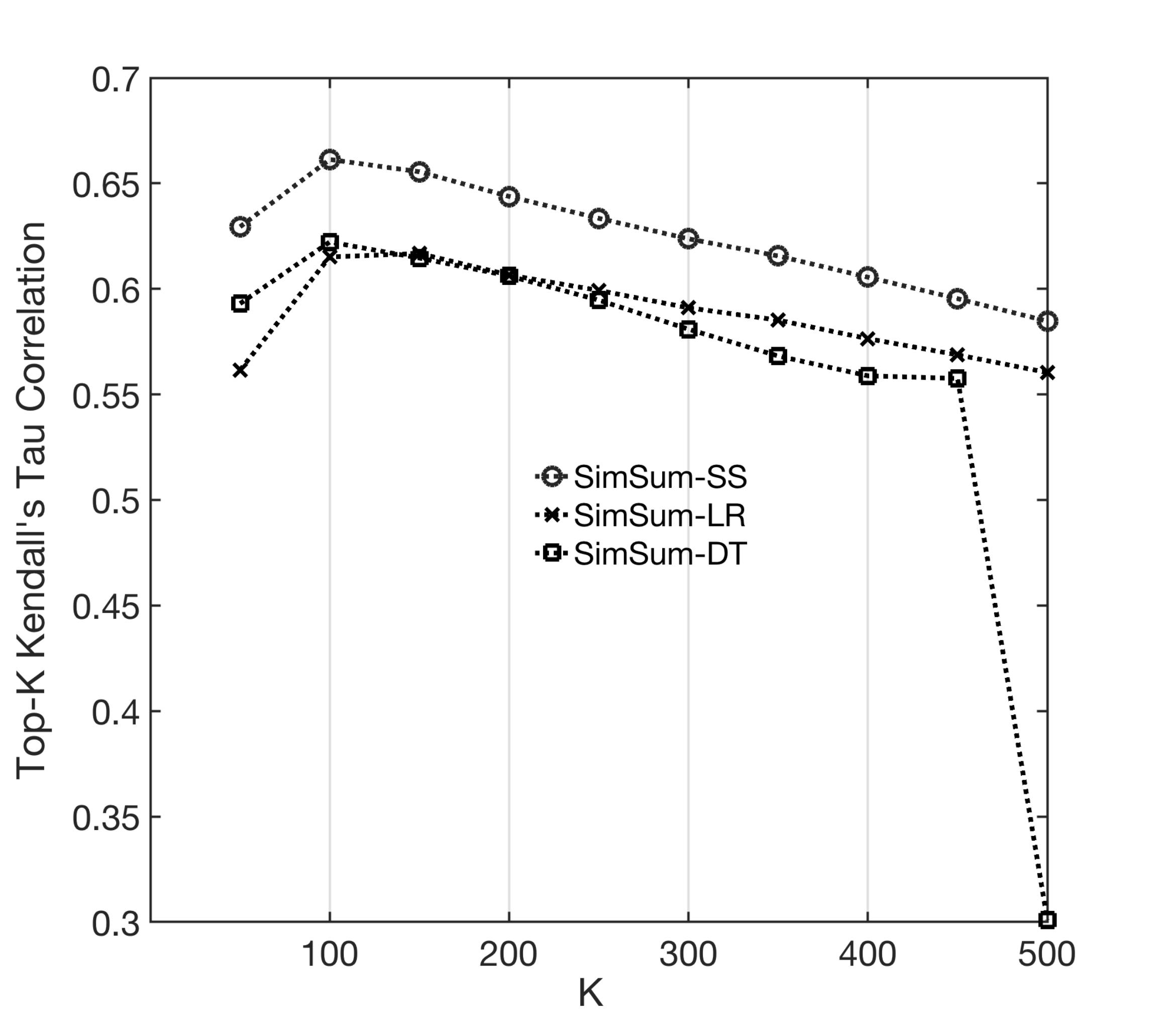}
    \caption{\emph{SimSum} Kendall's Tau on \emph{Korea}}
    \end{subfigure}
    
    \begin{subfigure}{0.47\textwidth}
    \includegraphics[width=\textwidth]{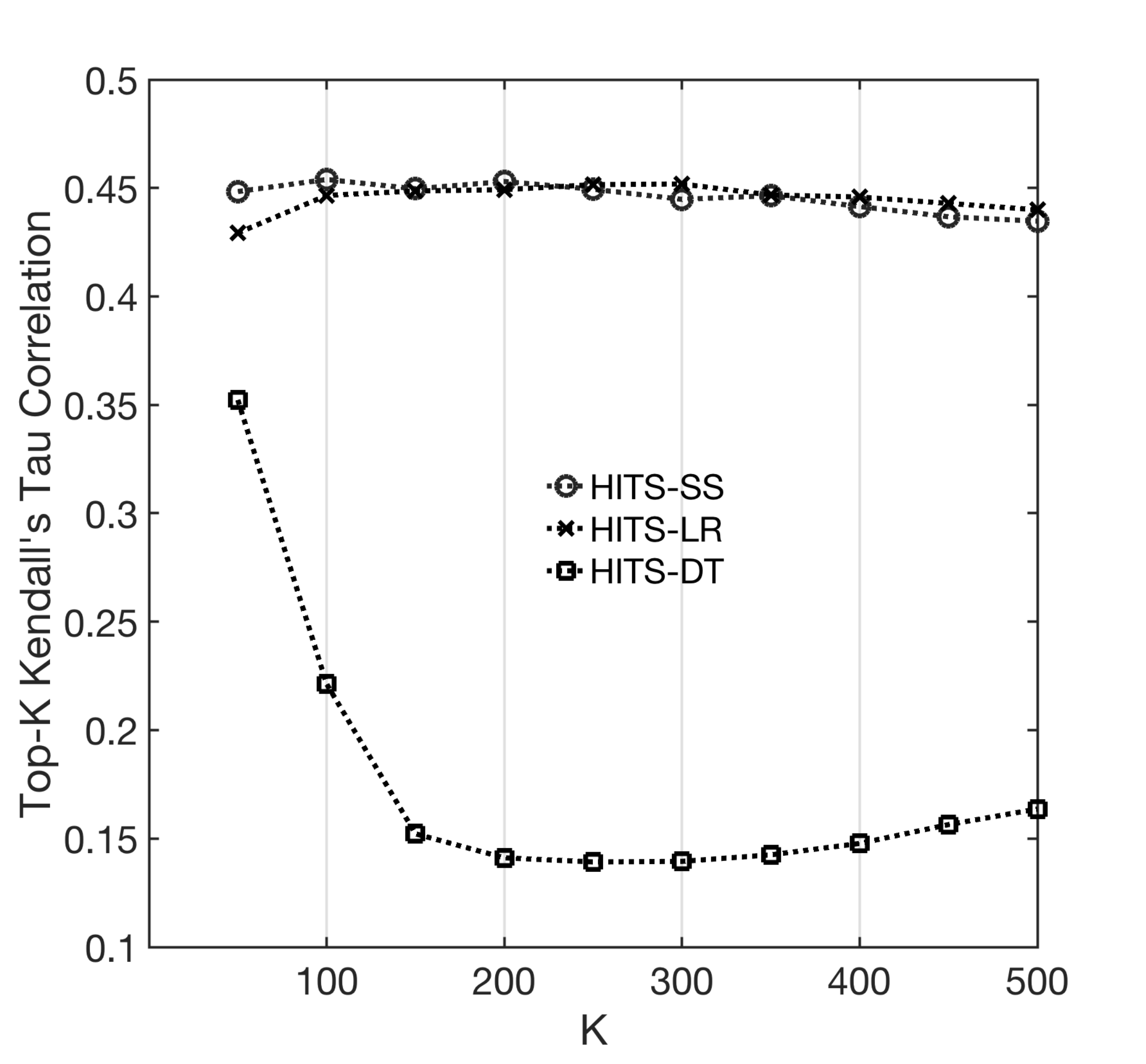}
    \caption{\emph{HITS} Kendall's Tau on \emph{USA}}
    \end{subfigure}
    \begin{subfigure}{0.47\textwidth}
    \includegraphics[width=\textwidth]{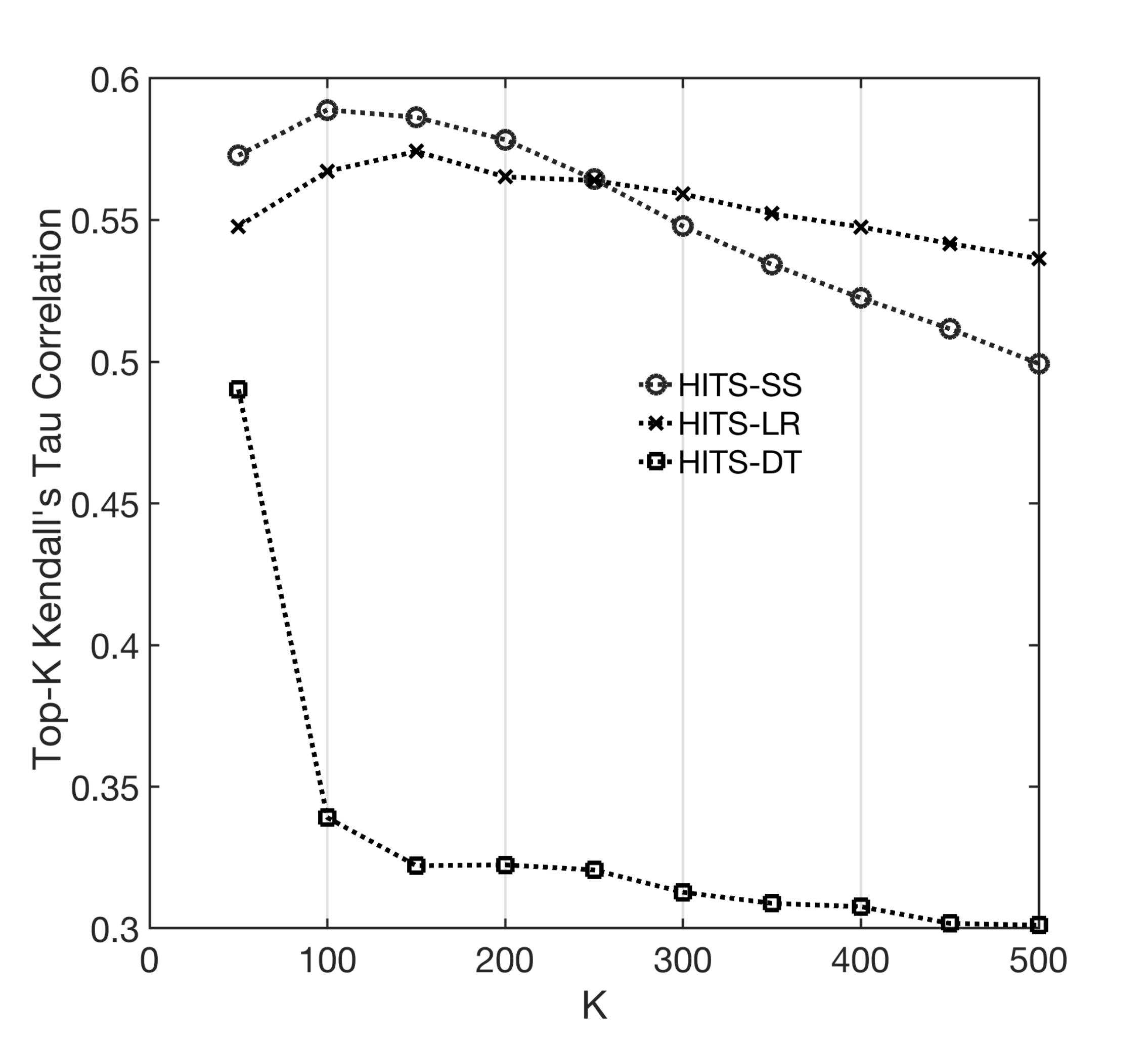}
    \caption{\emph{HITS} Kendall's Tau on \emph{Korea}}
    \end{subfigure}
    
    \begin{subfigure}{0.47\textwidth}
    \includegraphics[width=\textwidth]{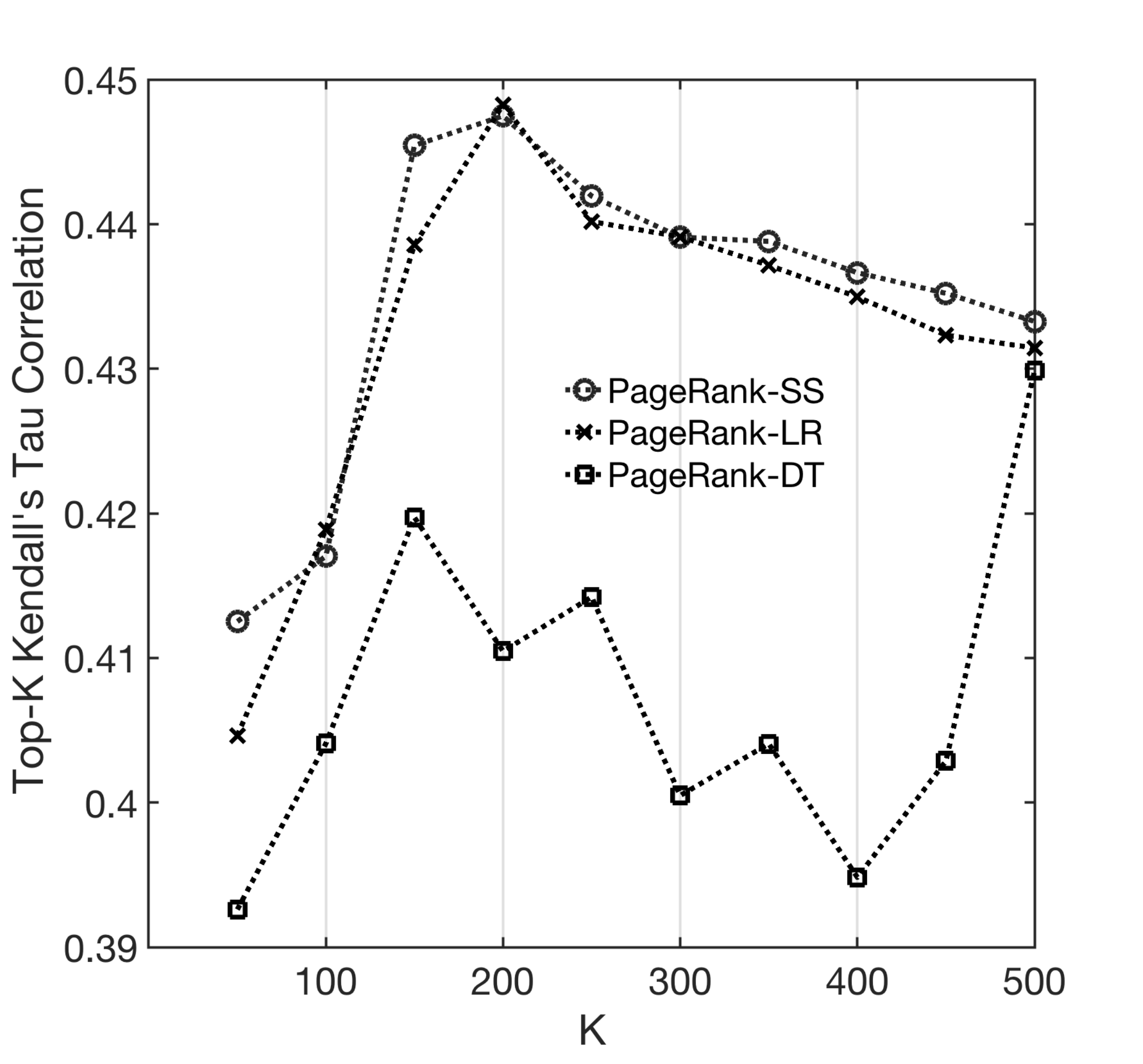}
    \caption{\emph{PageRank} Kendall's Tau on \emph{USA}}
    \end{subfigure}
    \begin{subfigure}{0.47\textwidth}
    \includegraphics[width=\textwidth]{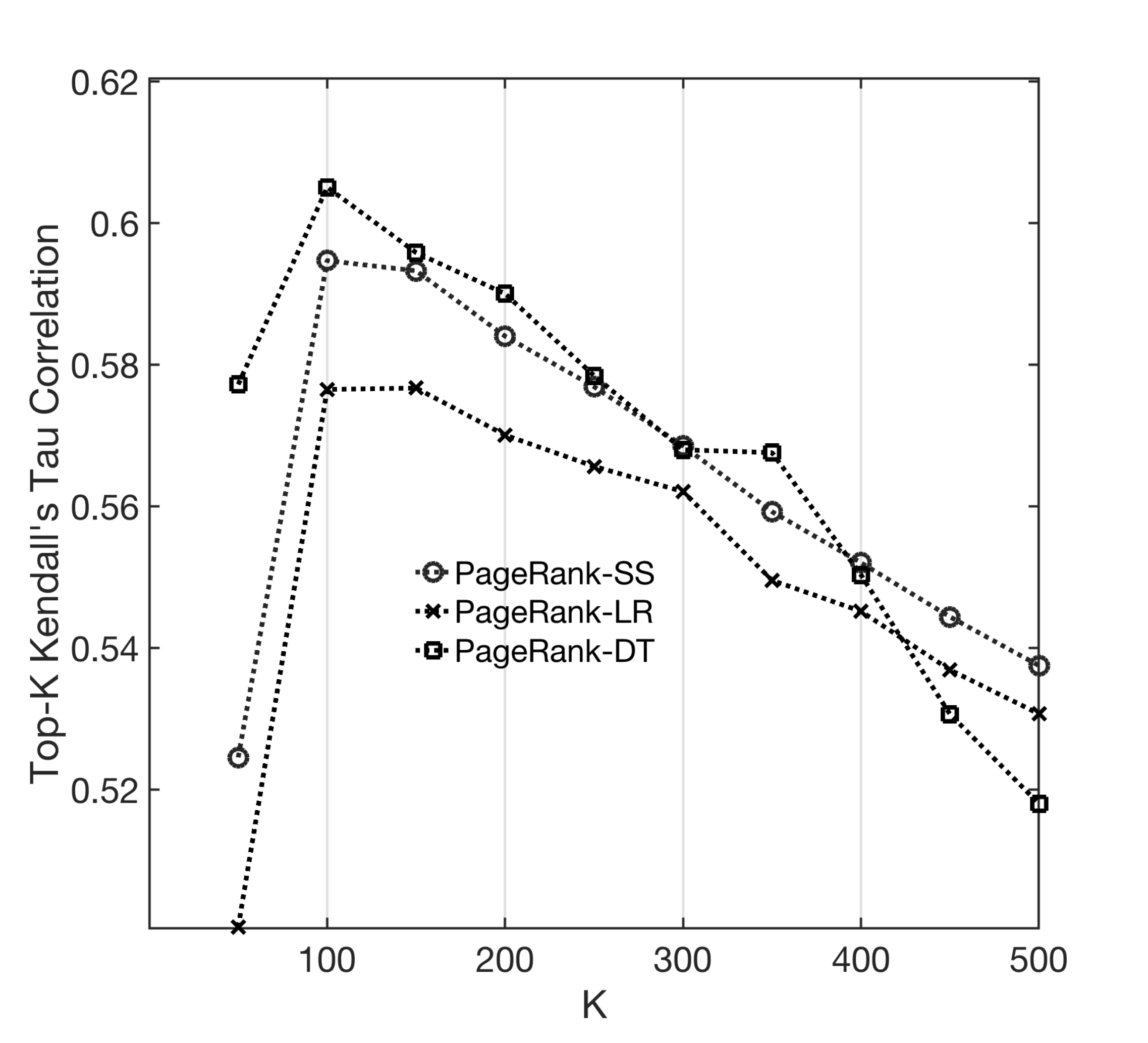}
    \caption{\emph{PageRank} Kendall's Tau on \emph{Korea}}
    \end{subfigure}
\caption{Comparison of Kendall's Tau correlation coefficients in testing under different macro-level churn ranking and micro-level churn prediction methods}
\label{fig:kt}
\end{figure}

In statistics, the Kendall's Tau coefficient, also known as Kendall rank correlation coefficient, the weighted Kendall's tau coefficient and the Spearman coefficient are all measures of rank correlation: the similarity of two orderings on the same data. For instance, Kendall's Tau coefficient~\cite{kendall1945treatment} is defined as:
\begin{align*}
    \text{Kendall's Tau} := \dfrac{P-Q}{n(n-1)/2},
\end{align*}
where $P$ is the number of concordant pairs in two orderings, $Q$ is the number of discordant pairs, and $n$ is the number of elements in the ranked list. In our problem setting, one ordering is obtained by ranking the games based on one of ranking scores, namely $\mathcal{S}_{ss}^{(t)}(v)$, $\mathcal{S}_{hits}^{(t)}(v)$, or $\mathcal{S}_{pg}^{(t)}(v)$, and the other ordering is obtained by ranking the games based on the ground truth $|\mathcal{N}_{v}^{(t)}\setminus \mathcal{N}_{v}^{(t+1)}|$. The weighted Kendall's Tau is a weighted version of Kendall's Tau, in which tie-breaking exchanges of high ranking are more influential than exchanges of low rankings, e.g., in~\cite{vigna2015weighted} an exchange to break the ties between elements with rank $r$ and $s$ has weight $1/(r+1) + 1/(s+1)$. Spearman coefficient, on the other hand, quantifies how well the relationship between two ranked lists can be described using a monotonic function. Unlike Pearson correlation that can only assess linear relationships, Spearman correlation considers both linear and nonlinear monotonic relationships. The range of all three evaluation metrics is between $-1$ and $1$. Having a value closer to $1$ indicates that the ranking of the games is more similar to that based on the ground truth.

\begin{figure}
    \centering
    \begin{subfigure}{0.47\textwidth}
    \includegraphics[width=\textwidth]{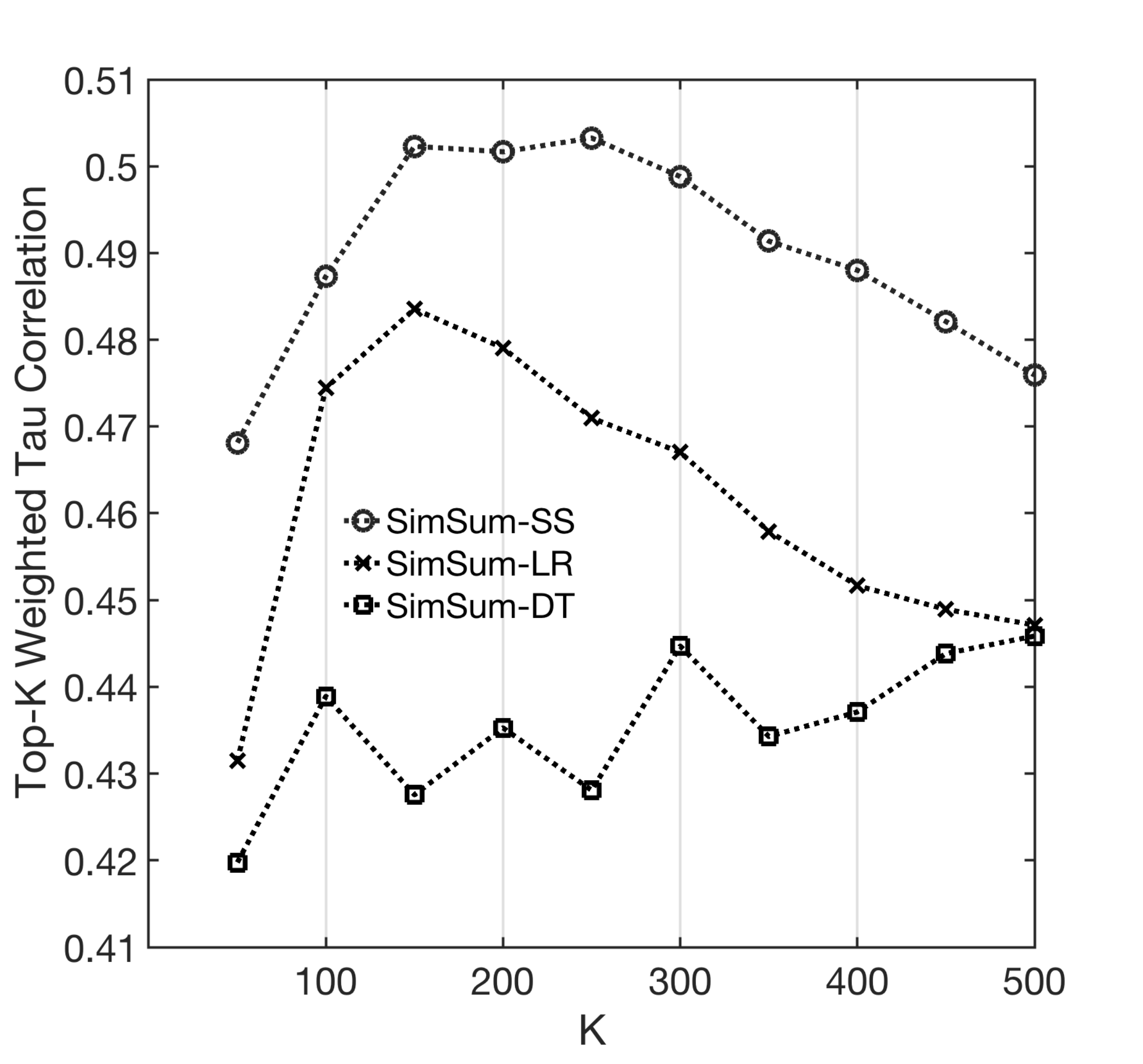} 
    \caption{\emph{SimSum} Weighted Kendall's Tau on \emph{USA}}
    \end{subfigure}
    \begin{subfigure}{0.47\textwidth}
    \includegraphics[width=\textwidth]{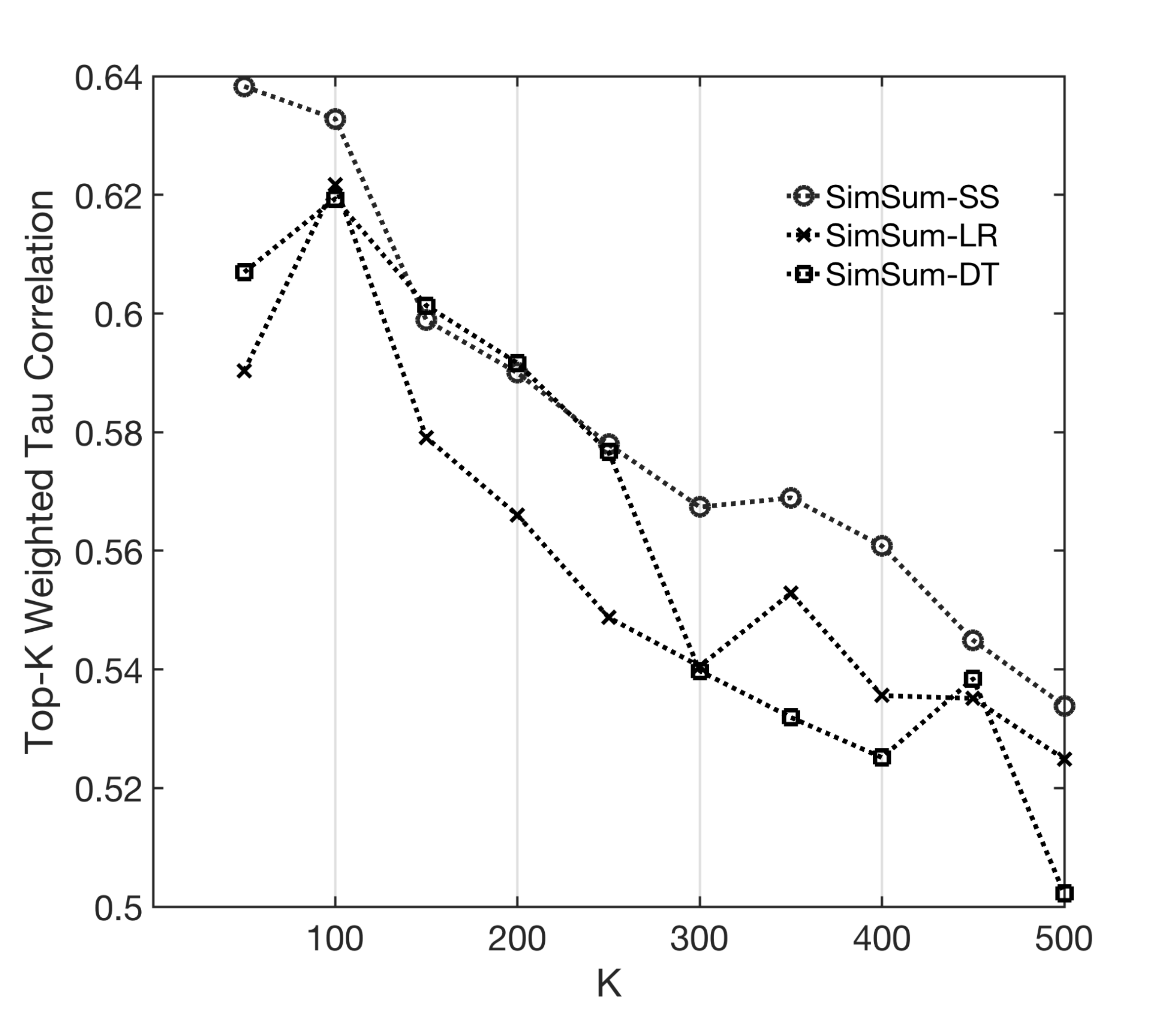}
    \caption{\emph{SimSum} Weighted Kendall's Tau  on \emph{Korea}}
    \end{subfigure}
    
    \begin{subfigure}{0.47\textwidth}
    \includegraphics[width=\textwidth]{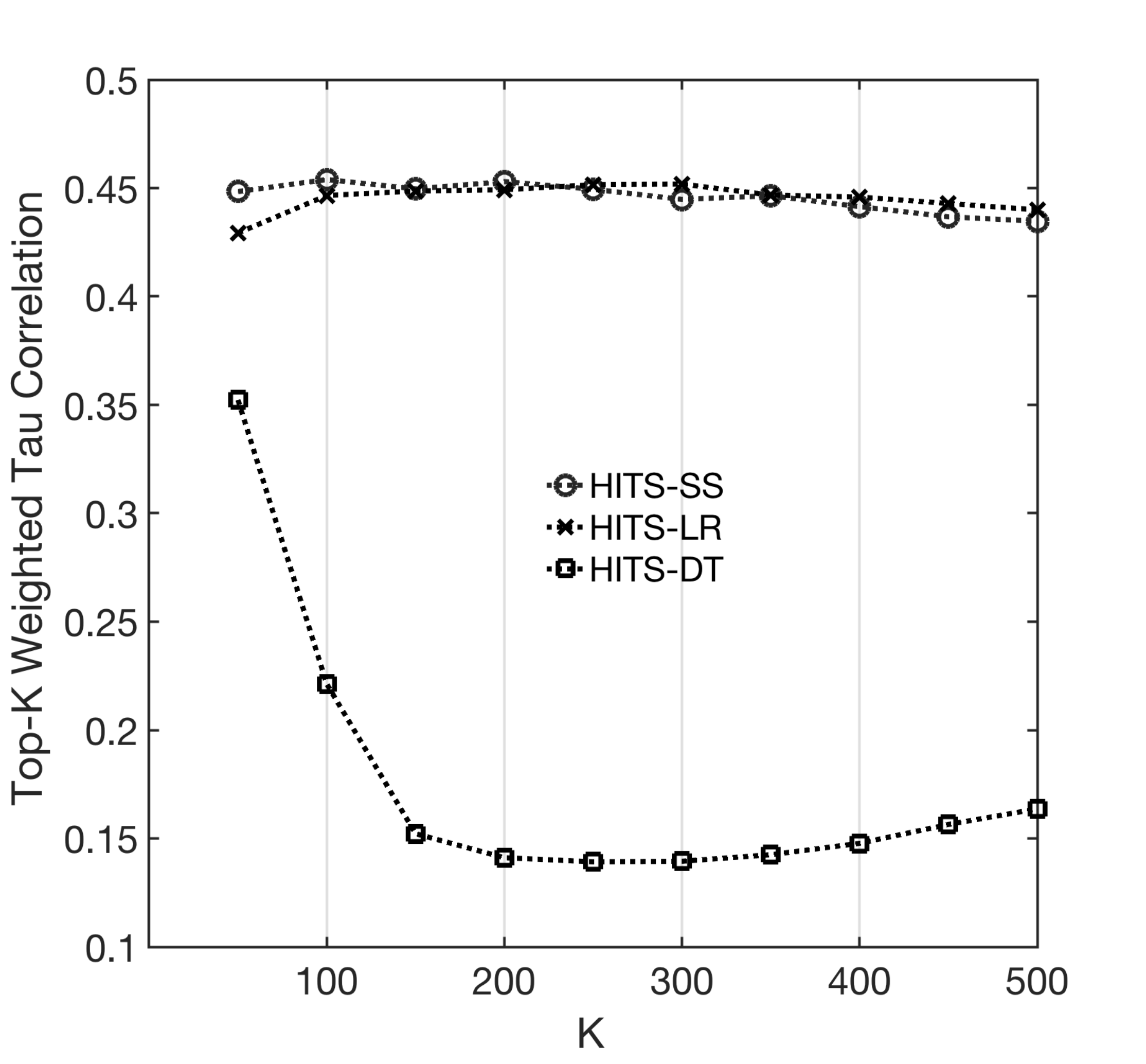}
    \caption{\emph{HITS} Weighted Kendall's Tau  on \emph{USA}}
    \end{subfigure}
    \begin{subfigure}{0.47\textwidth}
    \includegraphics[width=\textwidth]{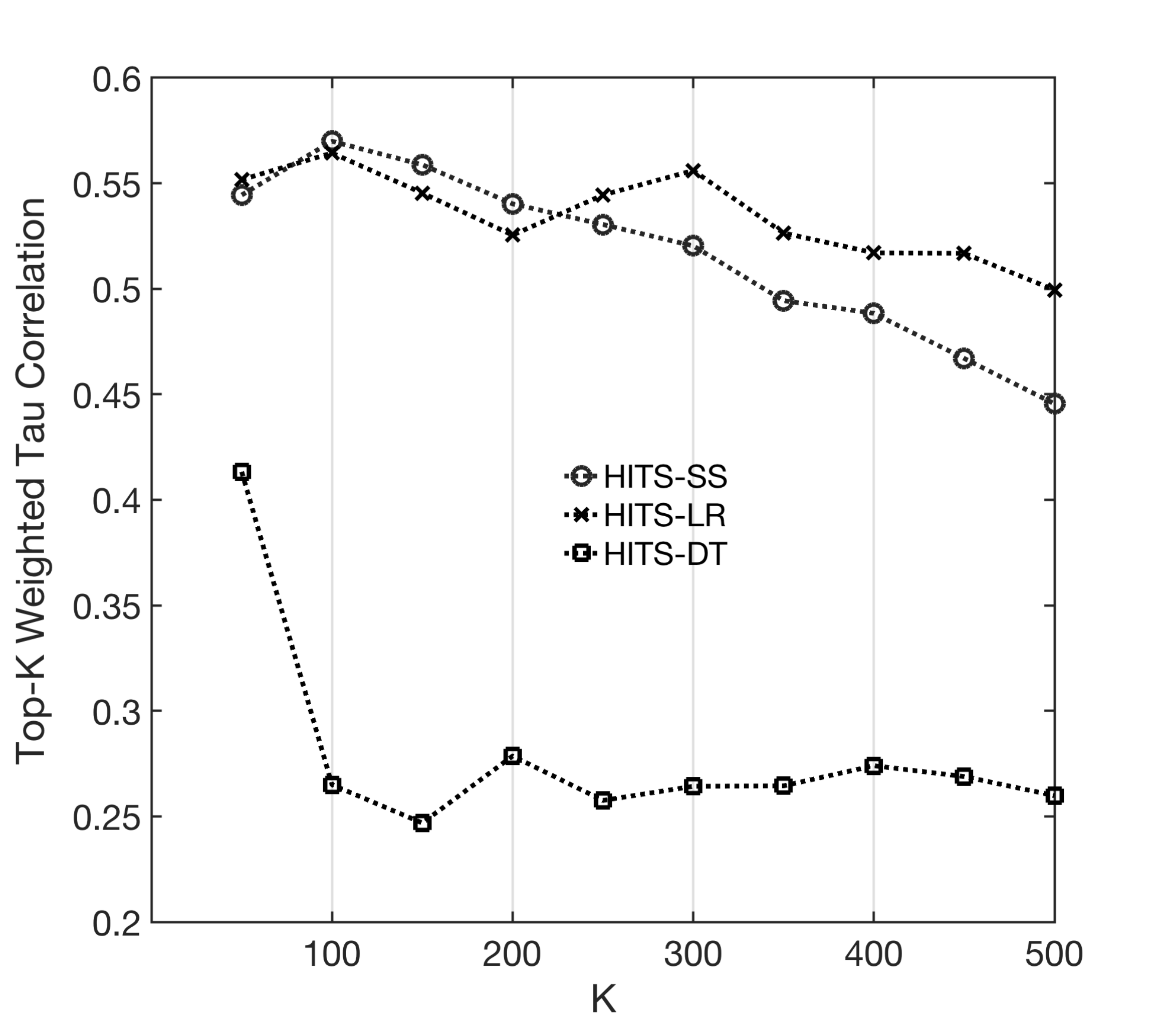}
    \caption{\emph{HITS} Weighted Kendall's Tau on \emph{Korea}}
    \end{subfigure}
    
    \begin{subfigure}{0.47\textwidth}
    \includegraphics[width=\textwidth]{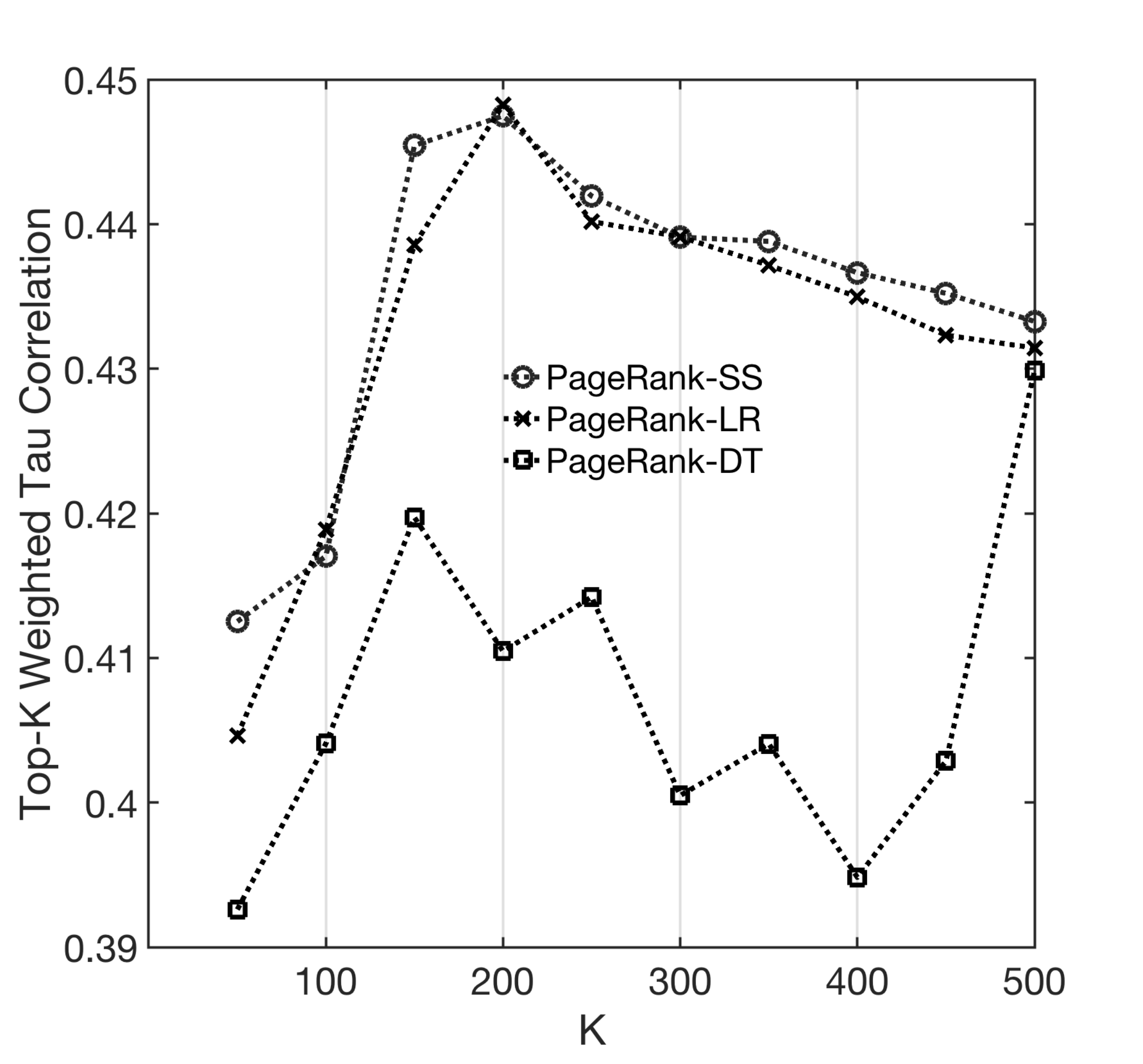}
    \caption{\emph{PageRank} Weighted Kendall's Tau on \emph{USA}}
    \end{subfigure}
    \begin{subfigure}{0.47\textwidth}
    \includegraphics[width=\textwidth]{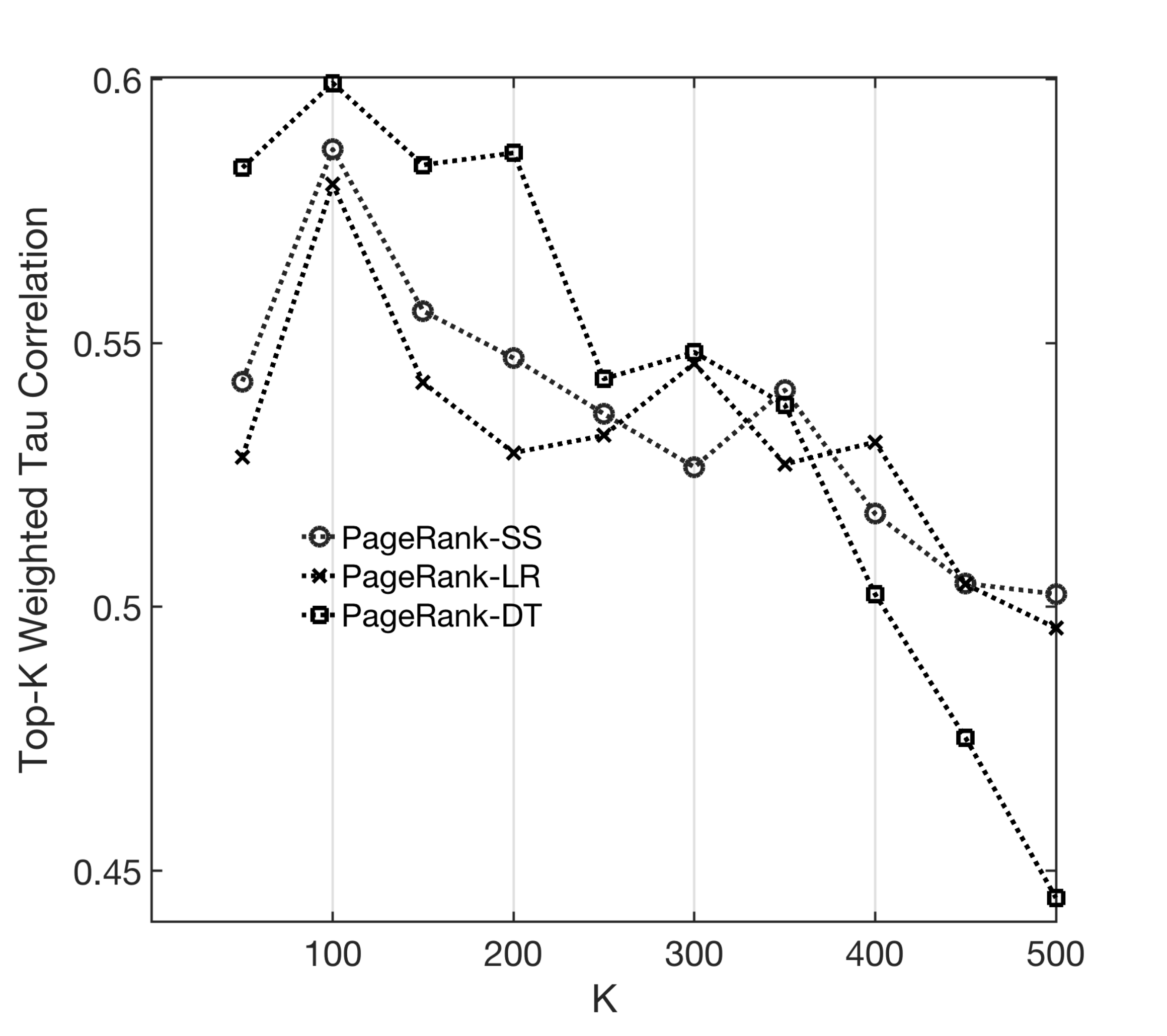}
    \caption{\emph{PageRank} Weighted Kendall's Tau on \emph{Korea}}
    \end{subfigure}
\caption{Comparison of Weighted Kendall's Tau correlation coefficients in testing under different macro-level churn ranking and micro-level churn prediction methods}
\label{fig:wkt}
\end{figure}

\begin{figure}
    \centering
    \begin{subfigure}{0.47\textwidth}
    \includegraphics[width=\textwidth]{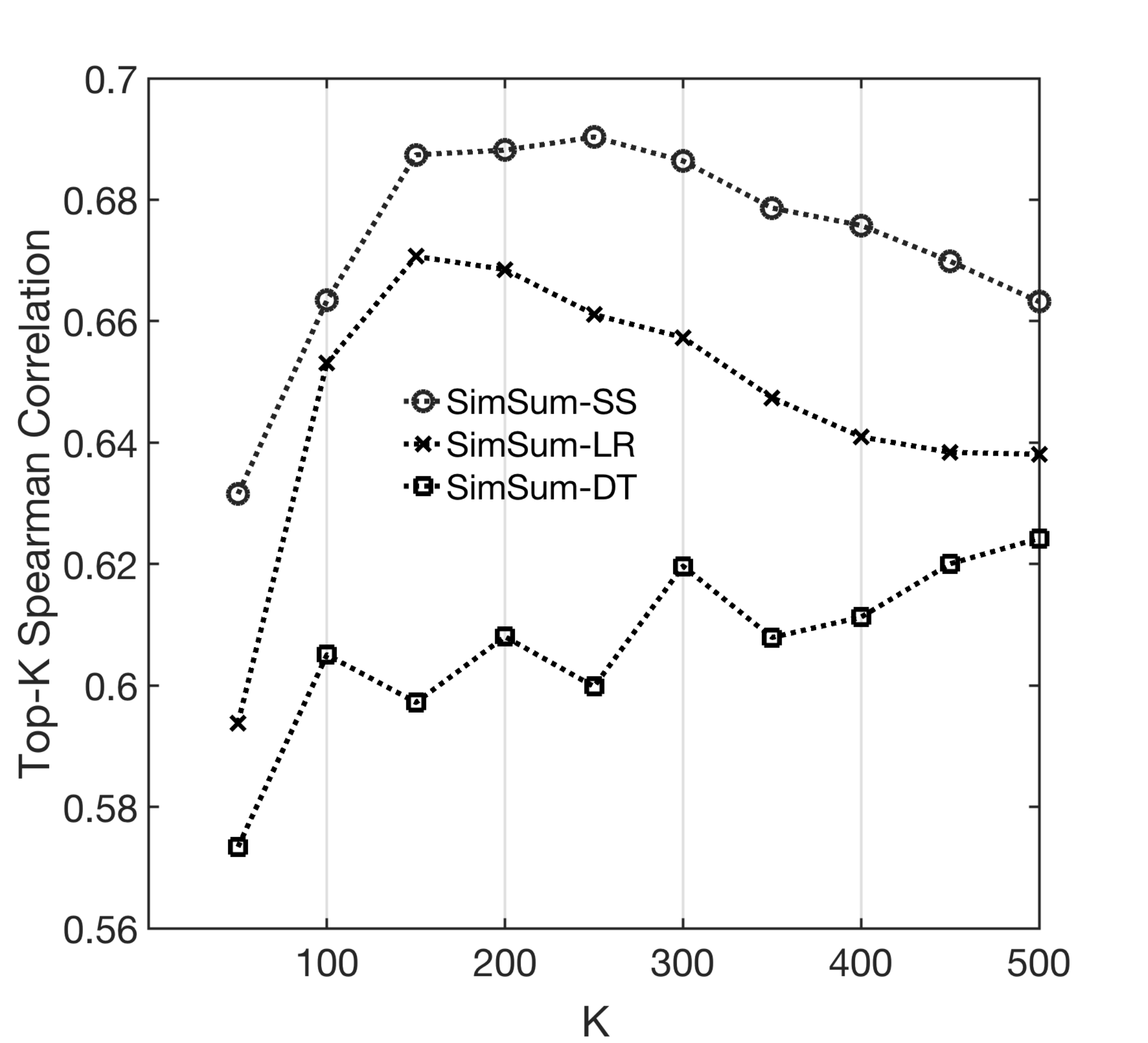} 
    \caption{\emph{SimSum} Spearman Coefficient on \emph{USA}}
    \end{subfigure}
    \begin{subfigure}{0.47\textwidth}
    \includegraphics[width=\textwidth]{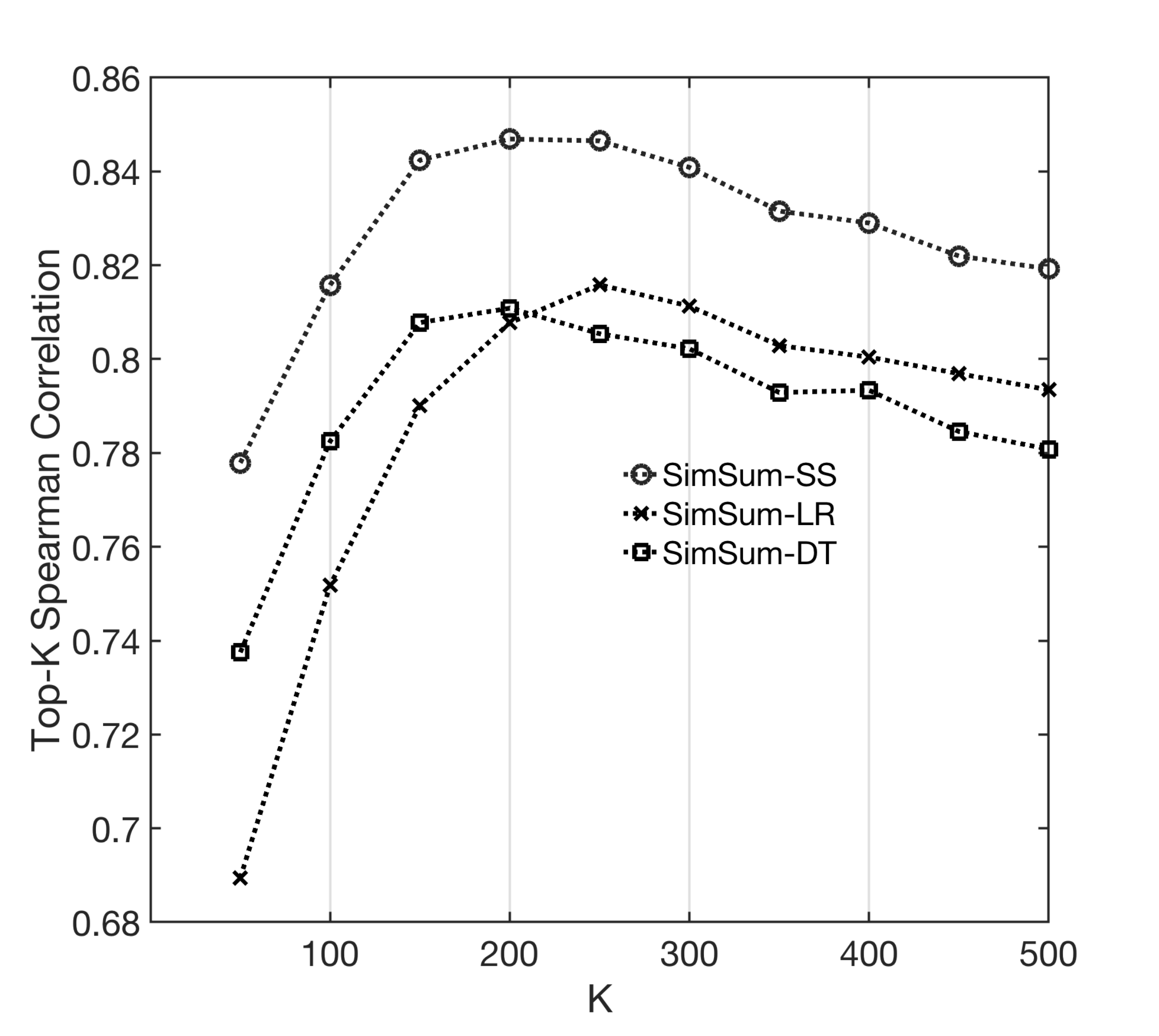}
    \caption{\emph{SimSum} Spearman Coefficient on \emph{Korea}}
    \end{subfigure}
    
    \begin{subfigure}{0.47\textwidth}
    \includegraphics[width=\textwidth]{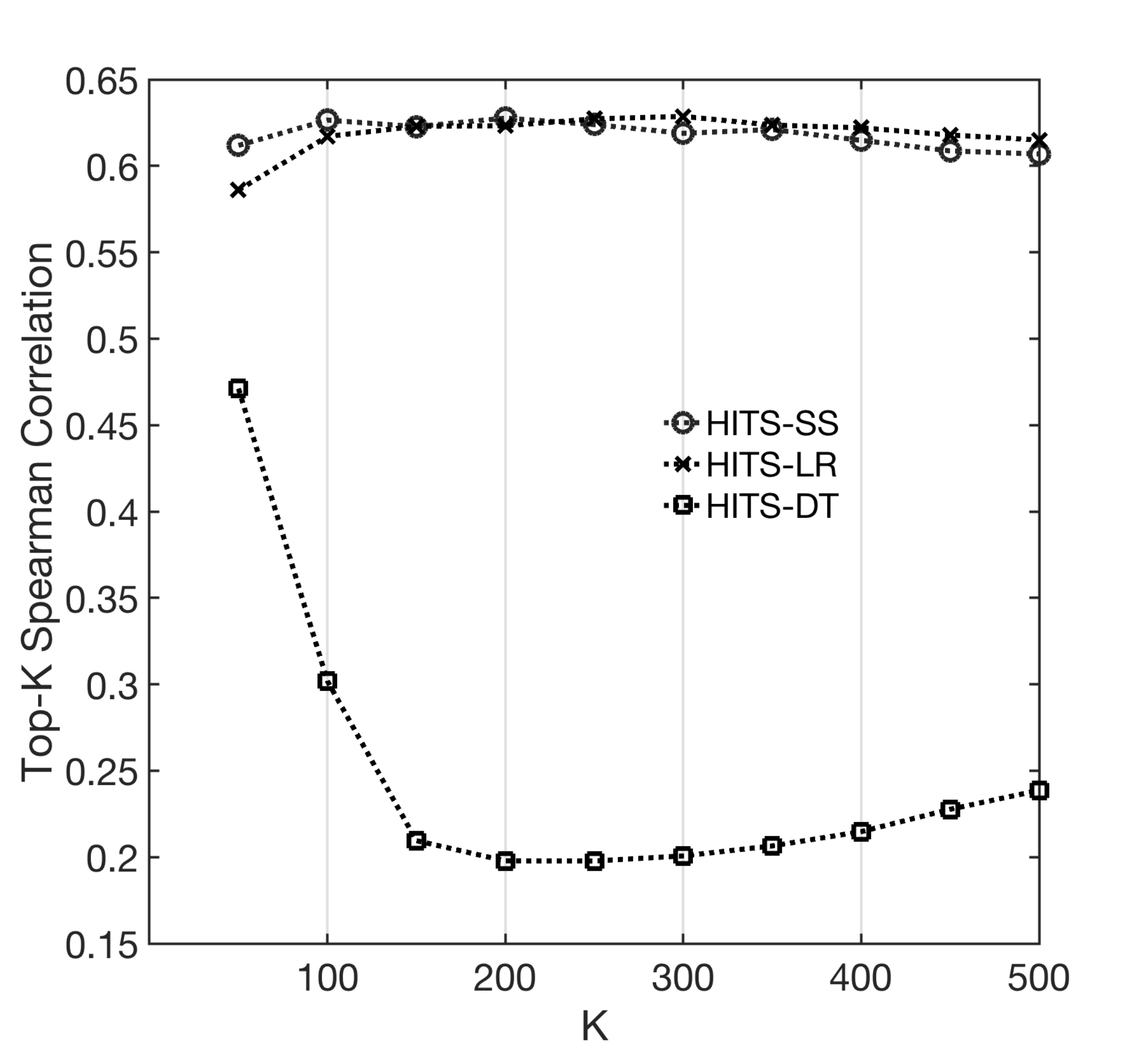}
    \caption{\emph{HITS} Spearman Coefficient on \emph{USA}}
    \end{subfigure}
    \begin{subfigure}{0.47\textwidth}
    \includegraphics[width=\textwidth]{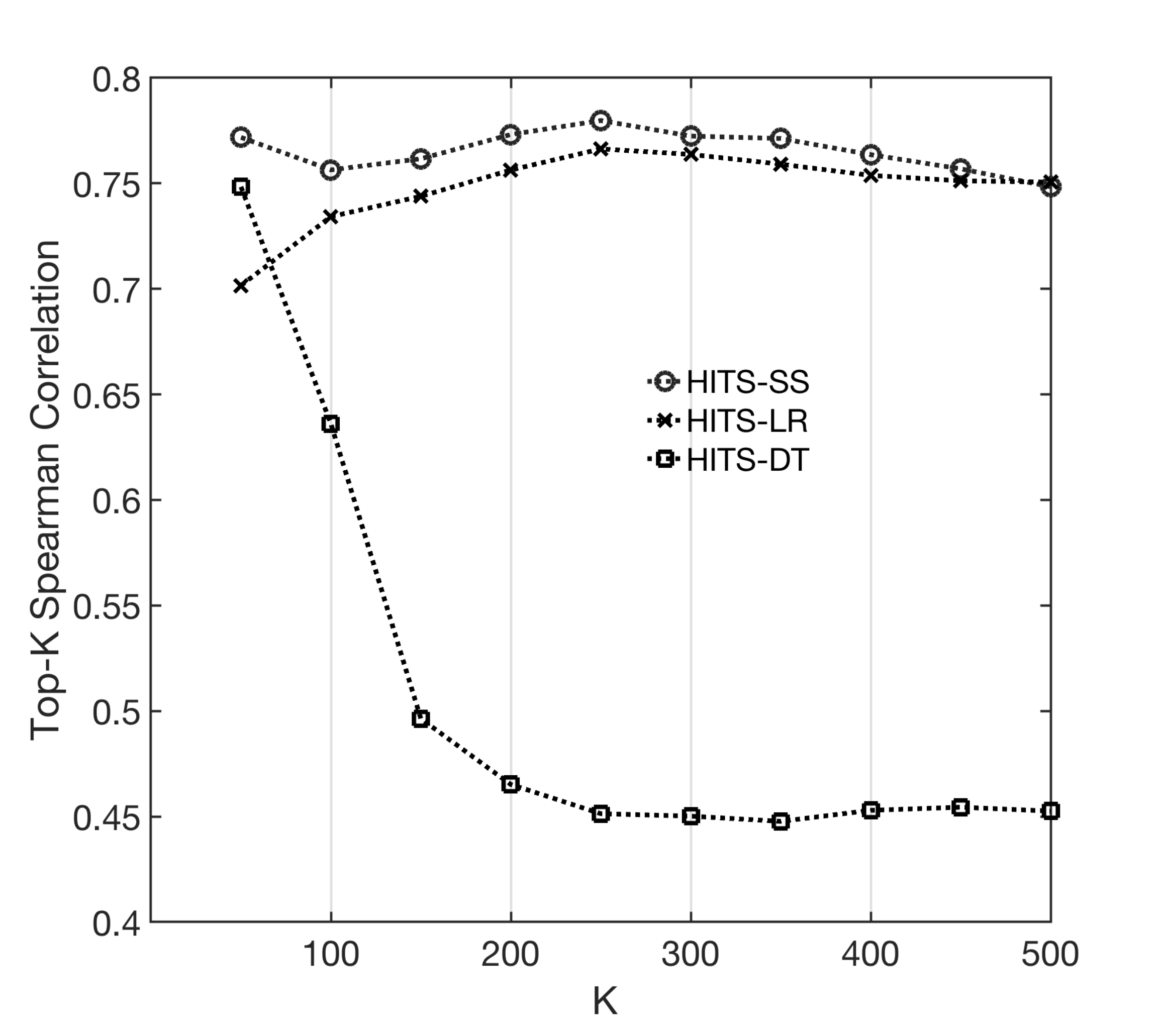}
    \caption{\emph{HITS} Spearman Coefficient on \emph{Korea}}
    \end{subfigure}
    
    \begin{subfigure}{0.47\textwidth}
    \includegraphics[width=\textwidth]{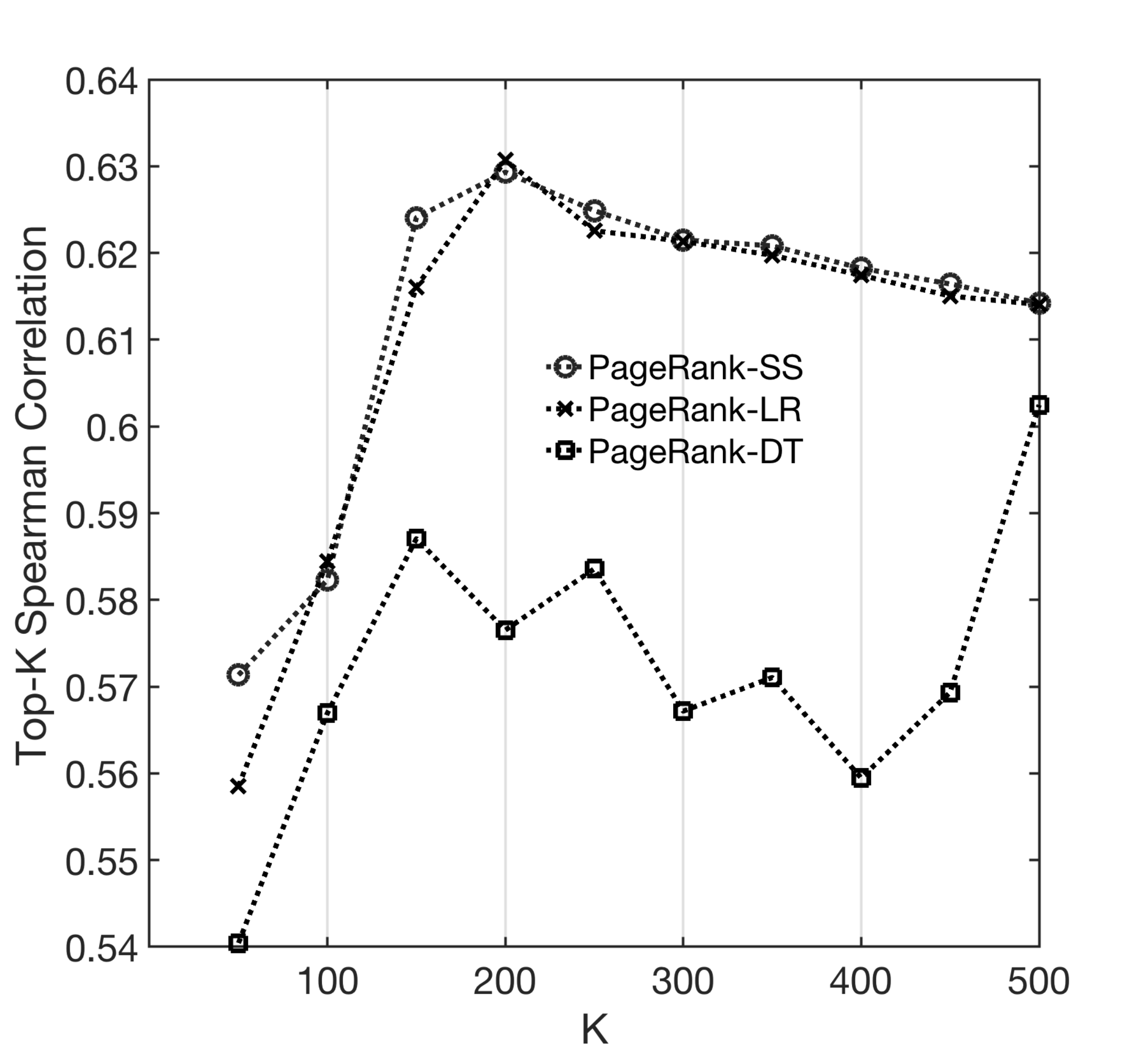}
    \caption{\emph{PageRank} Spearman Coefficient on \emph{USA}}
    \end{subfigure}
    \begin{subfigure}{0.47\textwidth}
    \includegraphics[width=\textwidth]{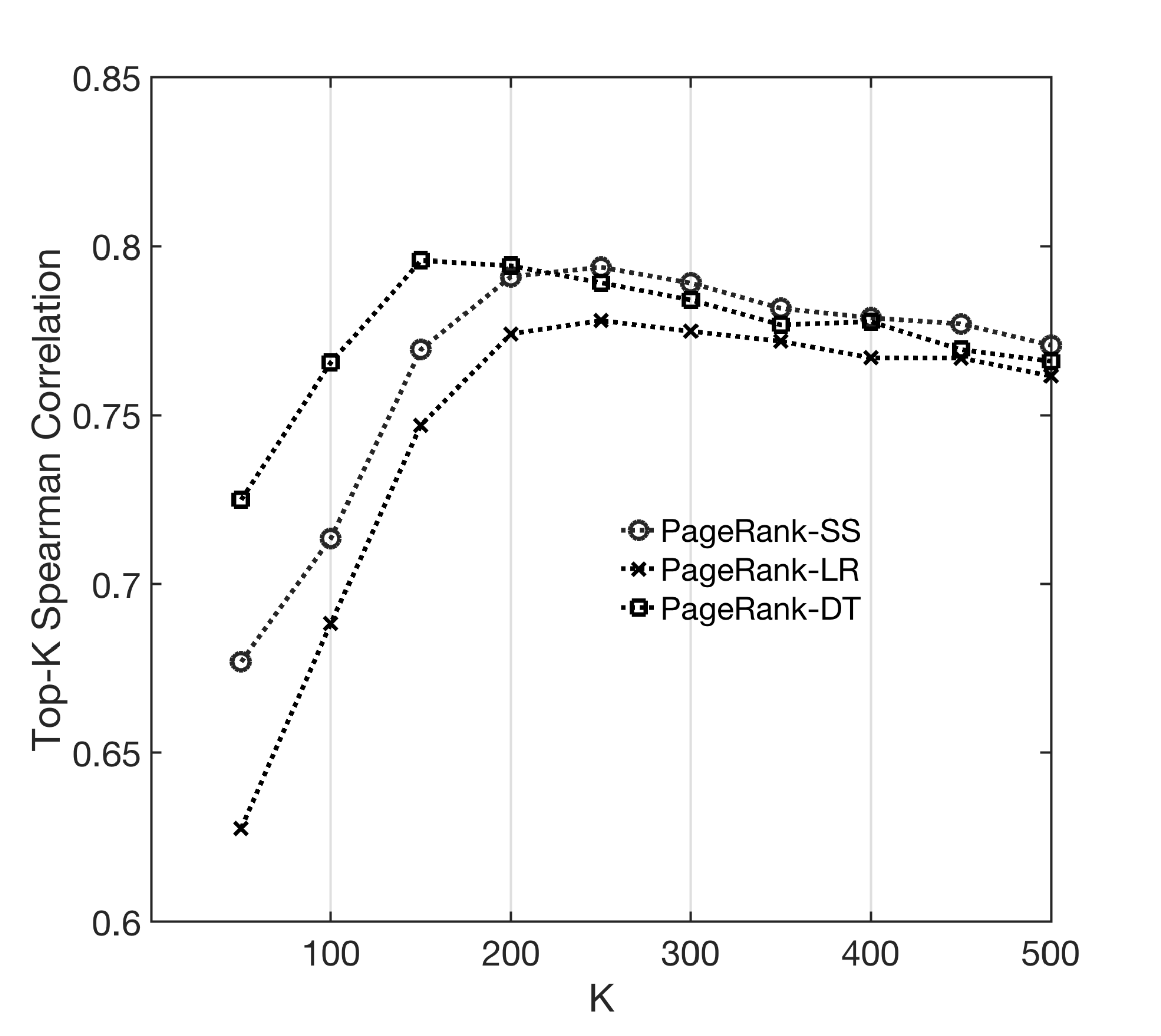}
    \caption{\emph{PageRank} Spearman Coefficient on \emph{Korea}}
    \end{subfigure}
\caption{Comparison of Spearman correlation coefficients in testing under different macro-level churn ranking and micro-level churn prediction methods}
\label{fig:sp}
\end{figure}
\begin{figure}
    \centering
    \begin{subfigure}{0.47\textwidth}
    \includegraphics[width=\textwidth]{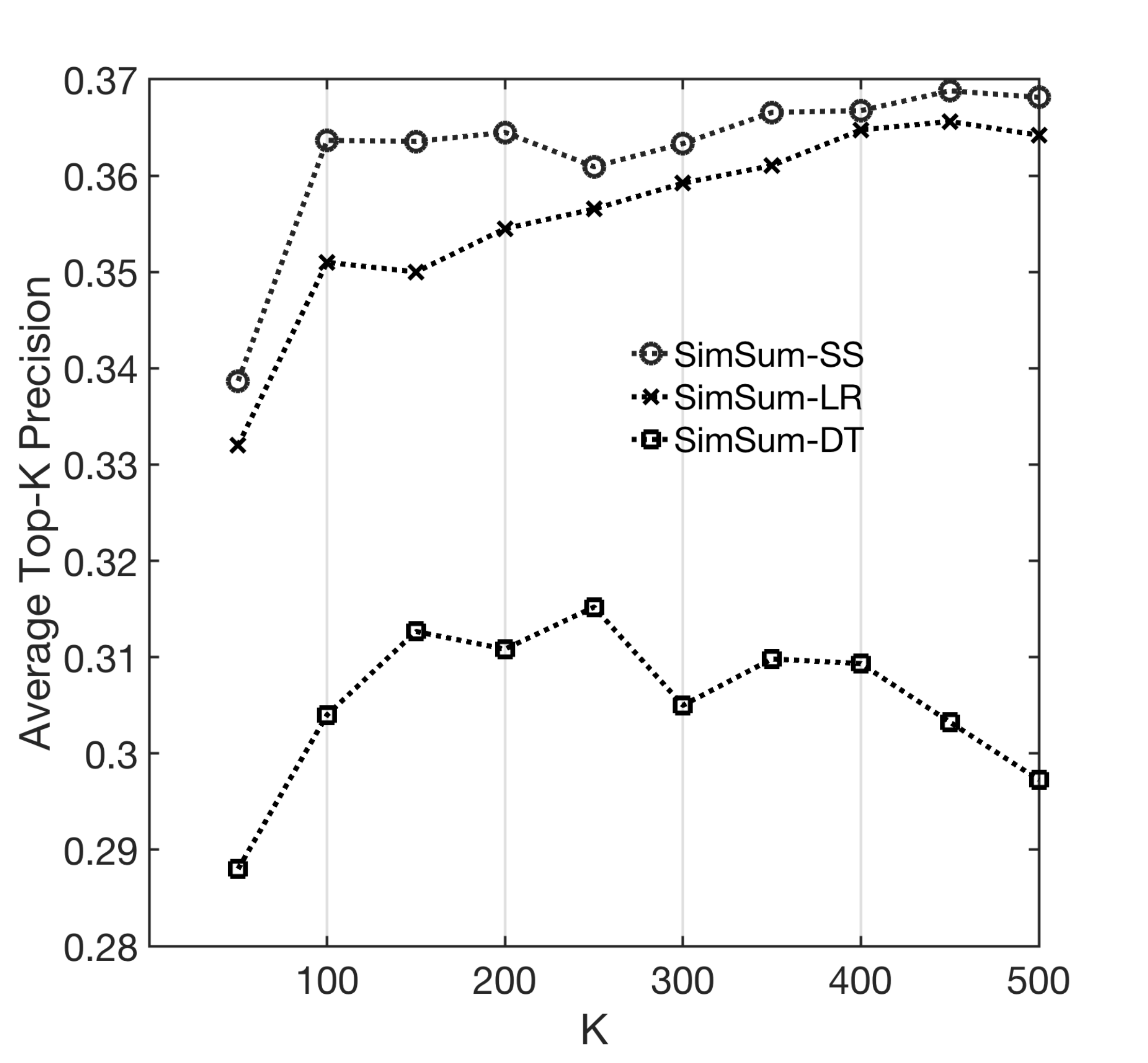} 
    \caption{\emph{SimSum} Average Precision at K on \emph{USA}}
    \end{subfigure}
    \begin{subfigure}{0.47\textwidth}
    \includegraphics[width=\textwidth]{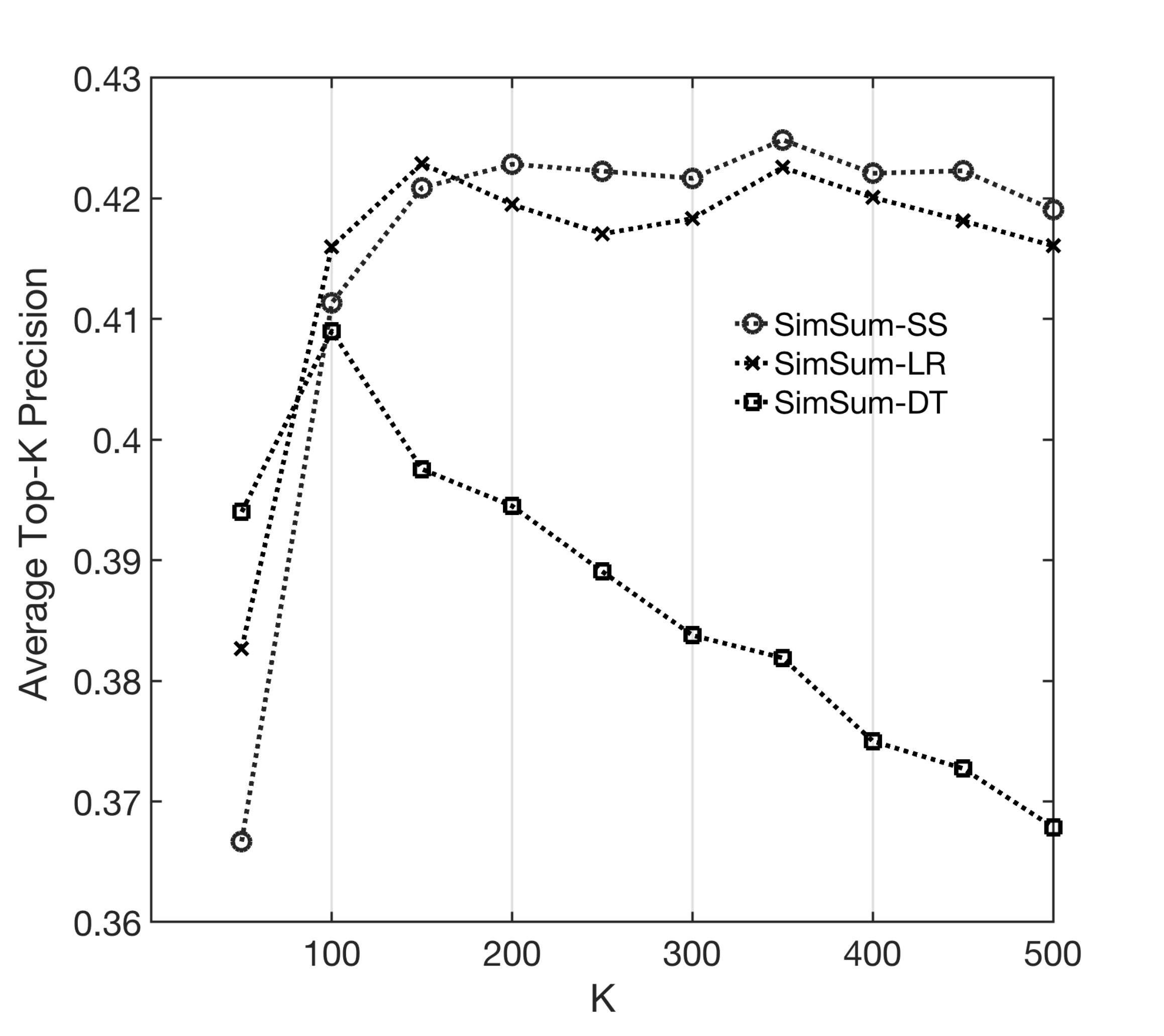}
    \caption{\emph{SimSum} Average Precision at K  on \emph{Korea}}
    \end{subfigure}
    
    \begin{subfigure}{0.47\textwidth}
    \includegraphics[width=\textwidth]{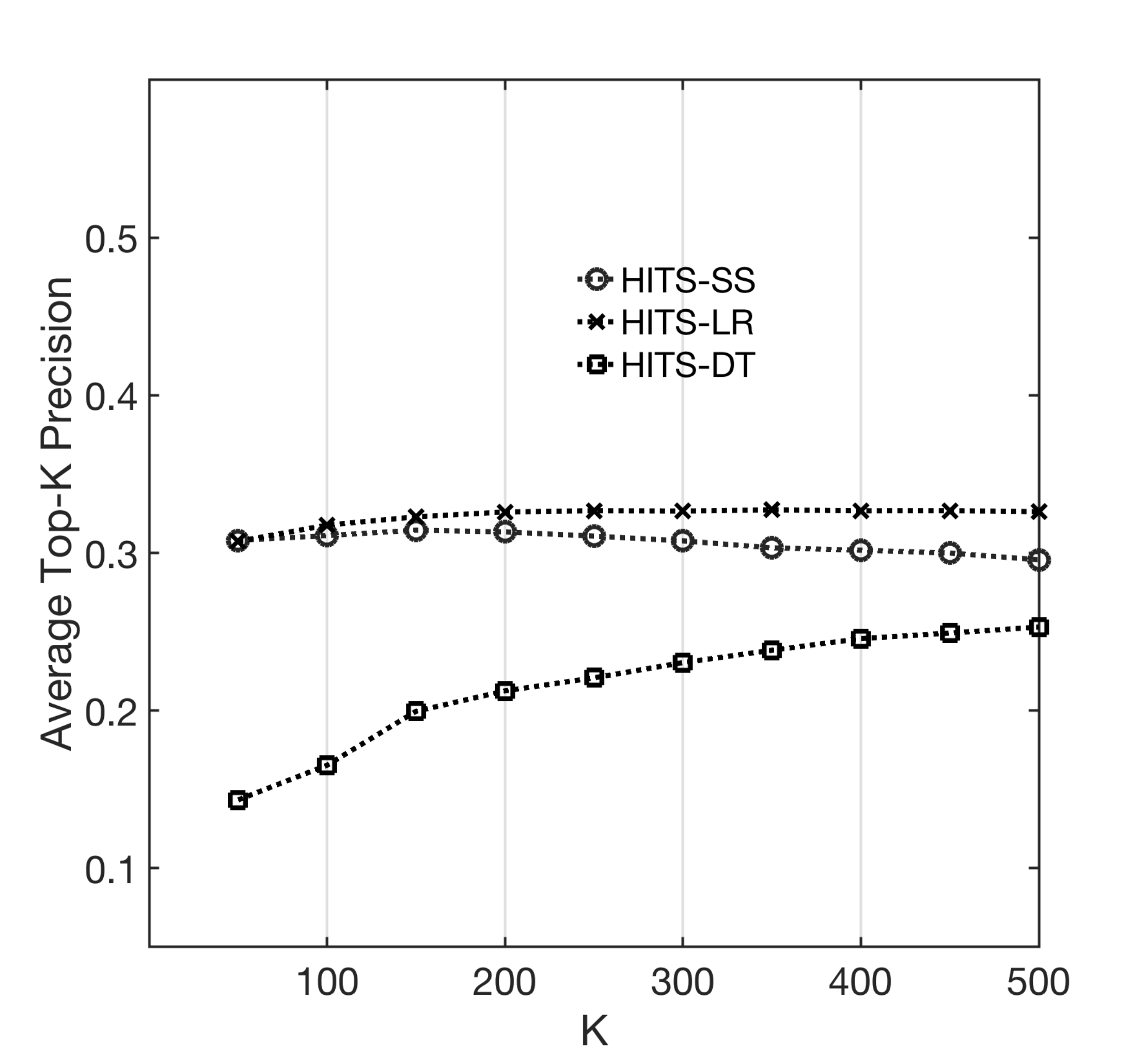}
    \caption{\emph{HITS} Average Precision at K  on \emph{USA}}
    \end{subfigure}
    \begin{subfigure}{0.47\textwidth}
    \includegraphics[width=\textwidth]{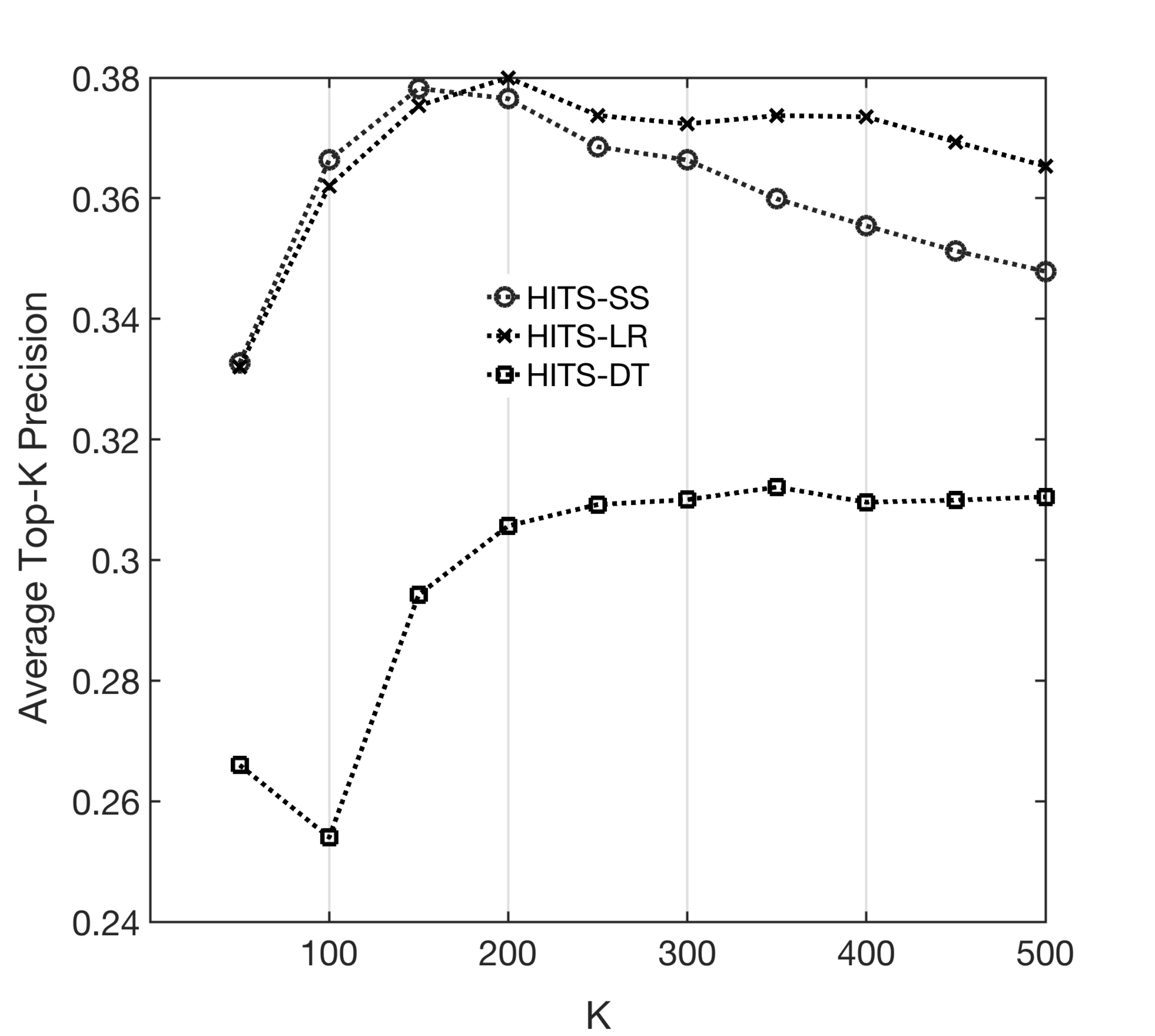}
    \caption{\emph{HITS} Average Precision at K on \emph{Korea}}
    \end{subfigure}
    
    \begin{subfigure}{0.47\textwidth}
    \includegraphics[width=\textwidth]{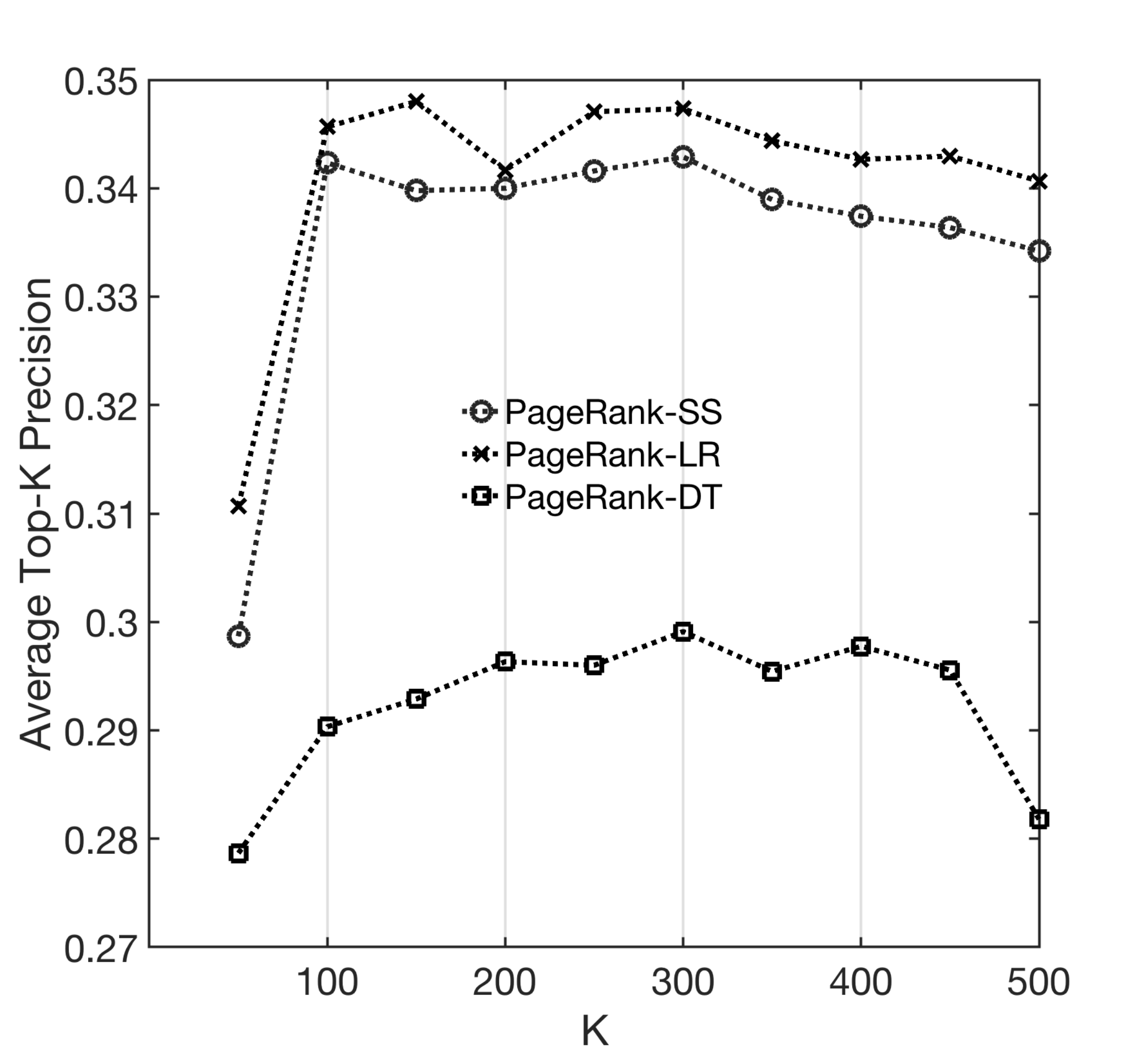}
    \caption{\emph{PageRank} Average Precision at K on \emph{USA}}
    \end{subfigure}
    \begin{subfigure}{0.47\textwidth}
    \includegraphics[width=\textwidth]{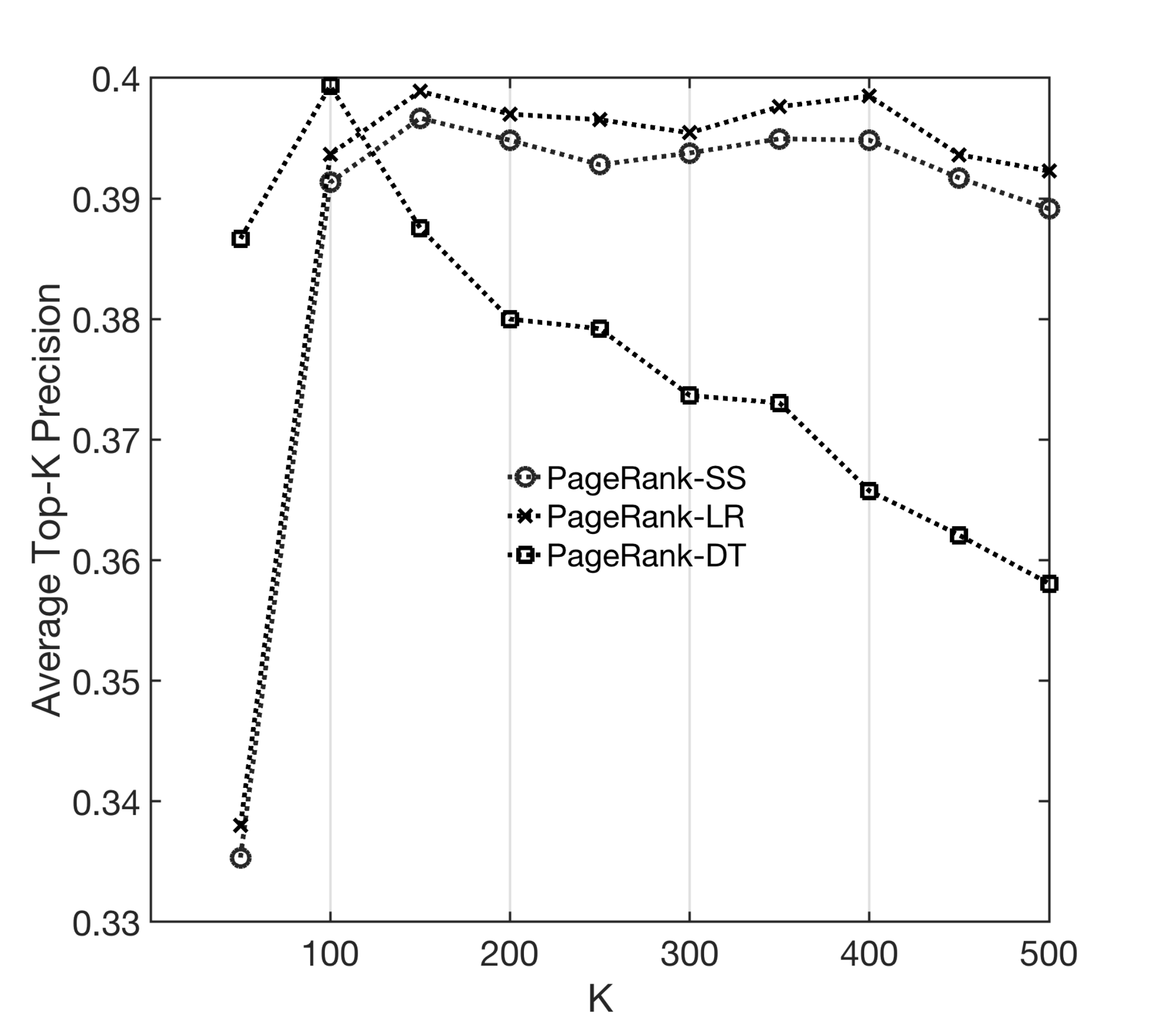}
    \caption{\emph{PageRank} Average Precision at K on \emph{Korea}}
    \end{subfigure}
\caption{Comparison of Average Precision at K in testing under different macro-level churn ranking and micro-level churn prediction methods}
\label{fig:apk}
\end{figure}

Since macro-level churn ranking is based on the results of micro-level churn prediction, we compare the performance of different combinations between the two parts. Each combination is referred to in the ``macro-micro'' format. For example, \emph{SimSum}-{\tt SS} represents the combination of using \emph{SimSum} as the macro-level churn ranking method and our semi-supervised method as the micro-level churn prediction model. Fig.~\ref{fig:kt}, Fig.~\ref{fig:wkt}, and Fig.~\ref{fig:sp} present the performance of different combinations in terms of Kendall's Tau, weighted Kendall's Tau and Spearman coefficients, respectively. Each performance metric is experimentally evaluated on the same test datasets described in Section~\ref{subsec:dataset}. Since micro-level churn prediction is performed daily, macro-level churn ranking is also calculated every day. The performance reported in the figures is the average over the testing period. As the ranked list of games can be as long as $20,000$, we only report the performance on the top $K$ games where the maximal $K$ value is set to 500. It can be observed that in most cases the macro-level churn ranking methods in combination with our semi-supervised method outperform those in combination with different baselines (i.e., {\tt LR}, {\tt DT}), especially when the value of $K$ is small. This is because the {\tt SS} achieves better performance in churn prediction than other baselines. This indicates that the quality of the churn prediction results is important for subsequent churn ranking, which takes the predicted churn probabilities as inputs. Moreover, we find that based on the Kendall's Tau, Weighted Kendall's Tau and Spearman coefficient, the performance of \emph{HITS} is more sensitive to the quality of churn prediction results. This may be because in the iteration, \emph{HITS} normalizes the scores over the sum of scores across all nodes, which can be large and lead to a skewed score distribution. In contrast, \emph{SimSum} and \emph{PageRank} normalize the scores based on only the scores of neighbors, which is more robust.


We also evaluate the performance of different churn ranking methods by two widely-adopted metrics in the recommendation domain: Average Precision at $K$ and Mean Average Precision (MAP). Precision at $K$ corresponds to the percentage of relevant results in the top $K$ games of the ranked list. A game at position $i$ in the ranked list based on our ranking scores, is considered relevant if it is in the top $i$ games of the ranked list based on the ground truth. The comparison is reported in Fig.~\ref{fig:apk}. Similarly, the scores reported are the average of all days in the testing period. In Fig.~\ref{fig:apk}, we observe similar patterns to those observed under the previous statistic metrics. In most cases, different macro-level churn ranking methods in combination with the proposed semi-supervised mode {\tt SS} achieve better performance than the combinations with other baseline. This again indicates the importance of the positive interaction between micro-level churn prediction and macro-level churn ranking. One shortcoming of this metric is that it fails to take into account the positions of relevant games. To overcome this limitation, we further adopt Mean Average Precision~\cite{zhu2004recall}, in which the relevance score is weighted by the position before being averaged. We report the comparison of the methods in terms of MAP in Table~\ref{tab:map}. Similar conclusions can be drawn from the results based on MAP. It is interesting to note that after putting larger weights on relevant games with higher ranking, the difference of \emph{HITS} between other two methods in the sensitivity to churn prediction quality is reduced.


\begin{table}
\centering
\caption{Mean Average Precision under different micro-level churn prediction and macro-level churn ranking methods on \emph{Korea} and \emph{USA}}
\begin{tabular}{|p{1.8cm}|p{1.5cm}|p{1.5cm}|p{1.5cm}|}
\hline
Area-Model & \emph{SimSum} & \emph{HITS} & \emph{PageRank}
\\ \hline
\emph{Korea}-{\tt SS} & \textbf{\underline{0.41}} &  \textbf{\underline{0.40}} & \textbf{\underline{0.40}}
\\ \hline
\emph{Korea}-{\tt LR} & 0.40 & 0.39 & 0.39
\\  \hline
\emph{Korea}-{\tt DT} & 0.40 & 0.40 & 0.39
\\  \hline
\emph{USA}-{\tt SS} &  \textbf{\underline{0.41}} &  \textbf{\underline{0.40}} & \textbf{\underline{0.38}} 
\\ \hline
\emph{USA}{-\tt LR} & 0.38 & 0.38 & 0.37
\\  \hline
\emph{USA}-{\tt DT} & 0.38 & 0.38 & 0.38
\\  \hline
\end{tabular}
\label{tab:map}
\end{table}

\section{Related Work}
\label{sec:related_work}
In this section, we review two categories of existing works that are relevant to this paper. The first category includes the existing works for (game) churn analysis. The early works~\cite{hadiji2014predicting, xie2015predicting, runge2014churn, tamassia2016predicting, xie2016predicting, drachen2016rapid} are based on more traditional machine learning models, such as logistic regression, random forests, SVM, naive Bayes, etc., and are experimentally evaluated on extremely small numbers of games (i.e., less than five). As shown in our experiments, their performance on large-scale real data with tens of thousands of mobile games and hundred of millions of user-app interactions is generally not satisfactory. In addition, we also observe scalability issues when they are applied to large-scale data. Some recent research has started to use more advanced models. \cite{perianez2016churn, bertens2017games, viljanen2017playtime} propose to use survival model for churn prediction, in which churn probability is modeled as a function of playtime. Kim et al.~\cite{kim2017churn} achieve good performance by using convolution neural networks (CNN) and long short-term memory networks (LSTM). There are also several recent deep-learning-based studies~\cite{wangperawong2016churn, umayaparvathi2017automated, martins2017predicting} for non-game churn prediction problems, which report better performance. This motivates us to employ deep neural network models. While being a generic solution, our model is able to accommodate the unique characteristics of mobile gaming. We provide a comparison between all existing works and ours in Table~\ref{table:references}. Much of the research relevant to macro-level churn ranking pays particular attention to popularity prediction, the opposite of churn ranking problem. However, they solve the problem from scratch and thus require efforts on feature engineering~\cite{malmi2014quality} or a complex model~\cite{zhu2015popularity}. In contrast, we propose to reuse the estimated churn probabilities of the micro-level churn prediction task to reduce the overhead. 

\begin{table}
\centering
\caption{Comparison between this paper and existing works in data size and key techniques}
{\small
\begin{tabular}{|p{0.7cm}|p{2.7cm}|p{4.3cm}|}
\hline
\textbf{Paper} & \textbf{Data size} & \textbf{Key techniques} \\ \hline
\cite{hadiji2014predicting} & 5 games \newline 50 thousand users & Decision tree, naive Bayes
\\ \hline
\cite{runge2014churn} & 2 games \newline 10 thousand users & Hidden Markov Model combined with a single layer neural network
\\ \hline
\cite{perianez2016churn, bertens2017games} & 1 game \newline 3 thousand users & Survival ensembles
\\ \hline
\cite{viljanen2017playtime} & 1 game \newline 1 thousand users & Survival model
\\ \hline
\cite{tamassia2016predicting} & 1 game \newline 10 thousand users & Hidden Markov Model
\\ \hline
\cite{xie2015predicting} & 2 games \newline 1 thousand users & SVM, decision tree, logistic regression
\\ \hline
\cite{drachen2016rapid} & 1 game \newline 130 thousand users & Heuristic decision tree
\\ \hline
\cite{xie2016predicting}   & 3 games \newline 60 thousand users & Logistic regression, decision tree, SVM
\\ \hline
\cite{kim2017churn}   & 3 games \newline 200 thousand users & Logistic regression, random forests, CNN, LSTM
\\ \hline
Ours & \textbf{40,000} games \newline 40 thousand users & Deep attributed edge embedding
\\ \hline
\end{tabular}
}
\label{table:references}
\end{table}

The second category contains recent works on graph embedding. Graph embedding automates the entire process of feature engineering by casting feature extraction as a representation learning problem. It frees models from human bottleneck introduced by handcrafted features and is able to utilize the full richness of data. However, most existing works such as Node2Vec~\cite{grover2016node2vec}, Deep Walk~\cite{perozzi2014deepwalk} and LINE~\cite{tang2015line} are \emph{node-centric} embedding. When it comes to game churn prediction, the entities to be embedded are edges that represent the relationship between players and games. Very limited work has studied edge embedding. Abu-El-Haija et al.~\cite{abu2017learning} propose to model edge embedding as a function of node embedding, in which the two ends of an edge are first embedded and then passed into a deep neural network with edge embedding as output. This method embeds edges in an indirect way and does not take into account any attribute information. In addition, these models~\cite{grover2016node2vec, perozzi2014deepwalk, tang2015line} all belong to the transductive framework, in which embedding cannot be generated if an object has never appeared in training. However, in our problem new users and new games appear continuously; new relationships between existing users and games may form anytime in the future. Therefore, for game churn prediction the capability of handling new edges is indispensable. 

Several very recent works have proposed the idea of inductive graph embedding~\cite{liang2017seano, hamilton2017inductive, yang2016revisiting, liu2018streaming}, inspired by which we propose our inductive edge embedding model for micro-level churn prediction. Our model improves these works in several major ways. First, our embedded features are learned in a \emph{semi-supervised} manner, where the supervised component is for churn prediction while the unsupervised component is for context recovery. Compared to unsupervised methods, embedding features learned in a semi-supervised way have been shown to achieve better performance~\cite{yang2016revisiting}. Second, our model captures graph dynamics by imposing temporal loss to the embeddings of the same edge in consecutive graph snapshots. Unlike all existing works, where embeddings represent structures or attribute information, embeddings in our model are designed to simultaneously capture contexts and graph dynamics. To our best knowledge, our model is the first to achieve such benefits.

\section{Conclusion}
\label{sec:conclusion}

Churn analysis of mobile games is a vital research and business problem that is backed up by a billion-dollar market. In this paper, we consider two important aspects of churn analysis of mobile games: micro-level churn prediction and macro-level churn ranking. We proposed a novel inductive semi-supervised embedding model for large-scale micro-level game churn prediction. It jointly learns a prediction function and an edge embedding function that can automatically map high-dimensional raw feature vectors to more informative latent feature vectors. We modeled the prediction function and the embedding function by deep neural networks, where the embedding component is designed to capture both contextual information and relationship dynamics. The contexts of an edge are sampled by a novel attributed random walk technique, which considers both topological adjacency and attribute similarities. To solve the macro-level churn ranking problem without much overhead, we proposed a simple method \emph{SimSum} based on the outputs of the micro-level churn prediction model and adapted 
the link analysis algorithms \emph{PageRank} and \emph{HITS} to further take into consideration the global information and mutual reinforcing effects among different games and users. We compared our approaches with several state-of-the-art baseline methods on large-scale real-world data collected from the Samsung Game Launcher platform. The extensive experimental results clearly demonstrate the effectiveness of our proposed solutions. 

Although the paper focuses on mobile game churn analysis, the proposed methods are not restricted to this specific problem and also work for more general problems. Since the inputs of our model only contain the attributes of objects and their relationships at different timestamps, the proposed model can be generalized to fulfill any churn analysis task where the underlying data can be modeled in a similar way, for instance, customer disengagement prediction in membership business (e.g., Apple Music, Costco, and insurance companies) and interest group unsubscription prediction in social networks (e.g., Facebook, Meetup, and Google+).

\begin{acknowledgements}
TBA
\end{acknowledgements}

\printbibliography

\section*{Author Biographies}
\leavevmode
\vspace{60pt}
\vbox{%
\begin{wrapfigure}{l}{80pt}
{\vspace*{-5pt}\includegraphics[width=0.2\textwidth]{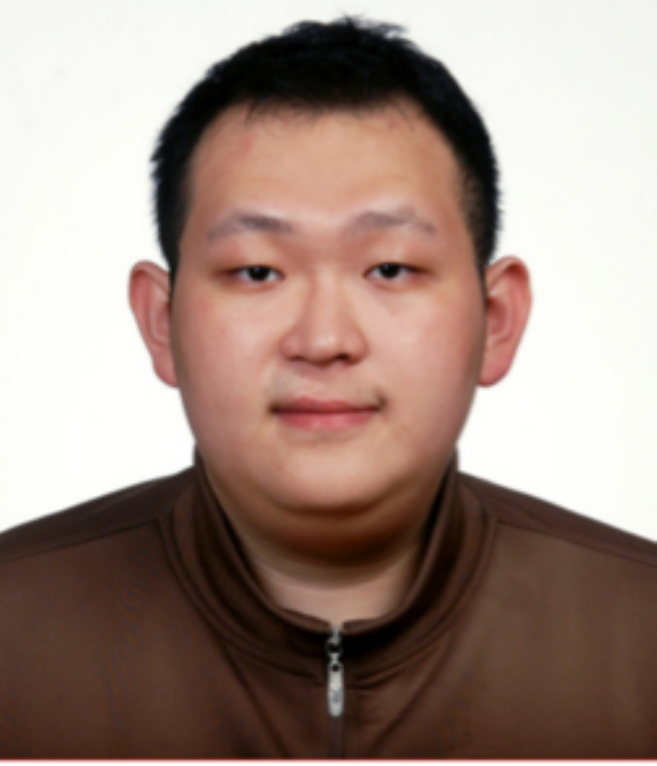}\vspace*{100pt}}%
\end{wrapfigure}
\noindent\small 
{\bf Xi Liu} is a Computer Engineering PhD candidate at Texas A\&M University. His research interests include graph representation learning, deep reinforcement learning and their application in recommender and data mining.}
\vspace{150pt}
\vbox{%
\begin{wrapfigure}{l}{80pt}
{\vspace*{-5pt}\includegraphics[width=0.2\textwidth]{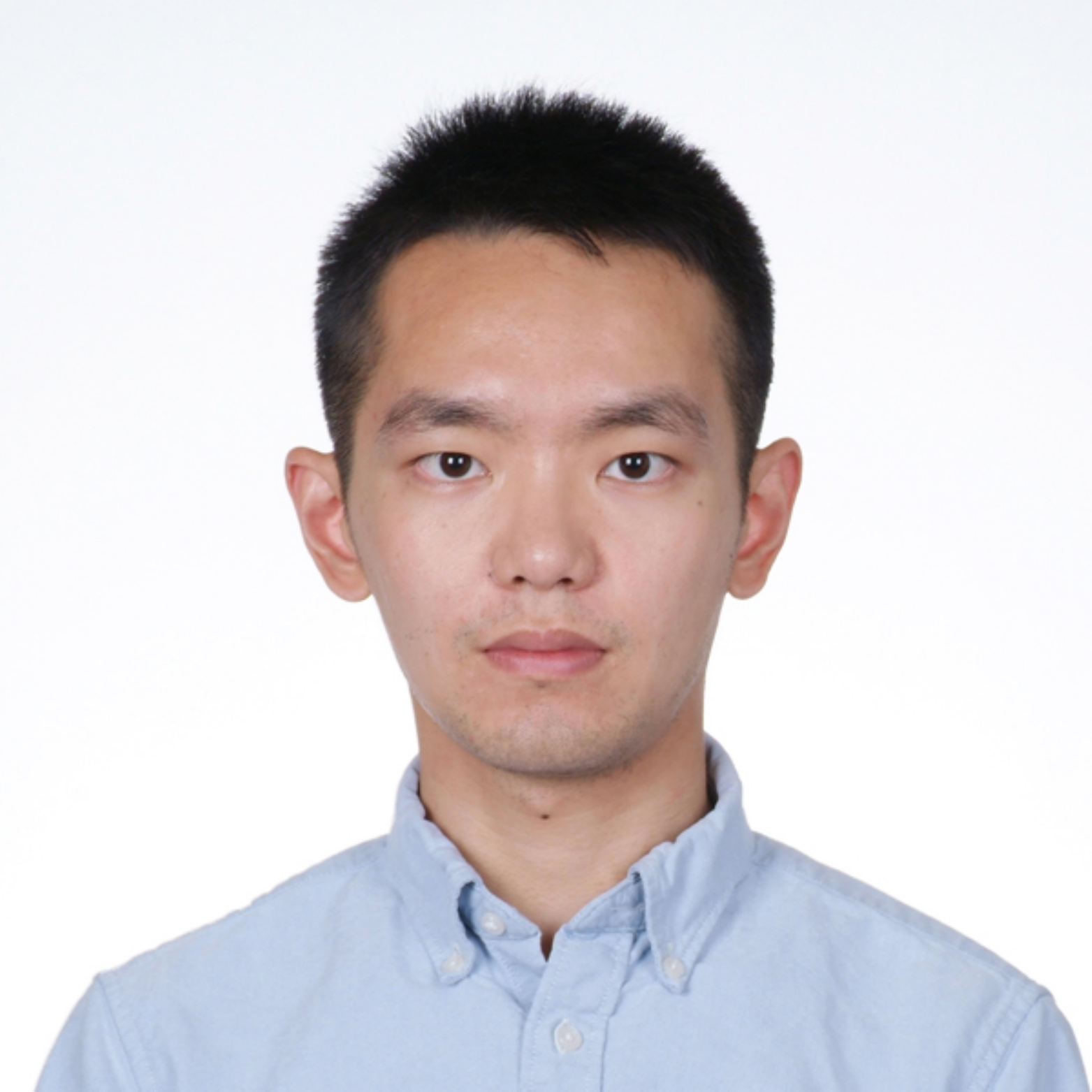}\vspace*{100pt}}%
\end{wrapfigure}
\noindent\small {\bf Muhe Xie} received his M.S. degree in data science from New York University. He is currently a software engineer at Samsung Research America. His research interests include machine learning and deep learning. He published paper in ICDM.}

\vspace{50pt}
\vbox{%
\begin{wrapfigure}{l}{80pt}
{\vspace*{-5pt}\includegraphics[width=0.2\textwidth]{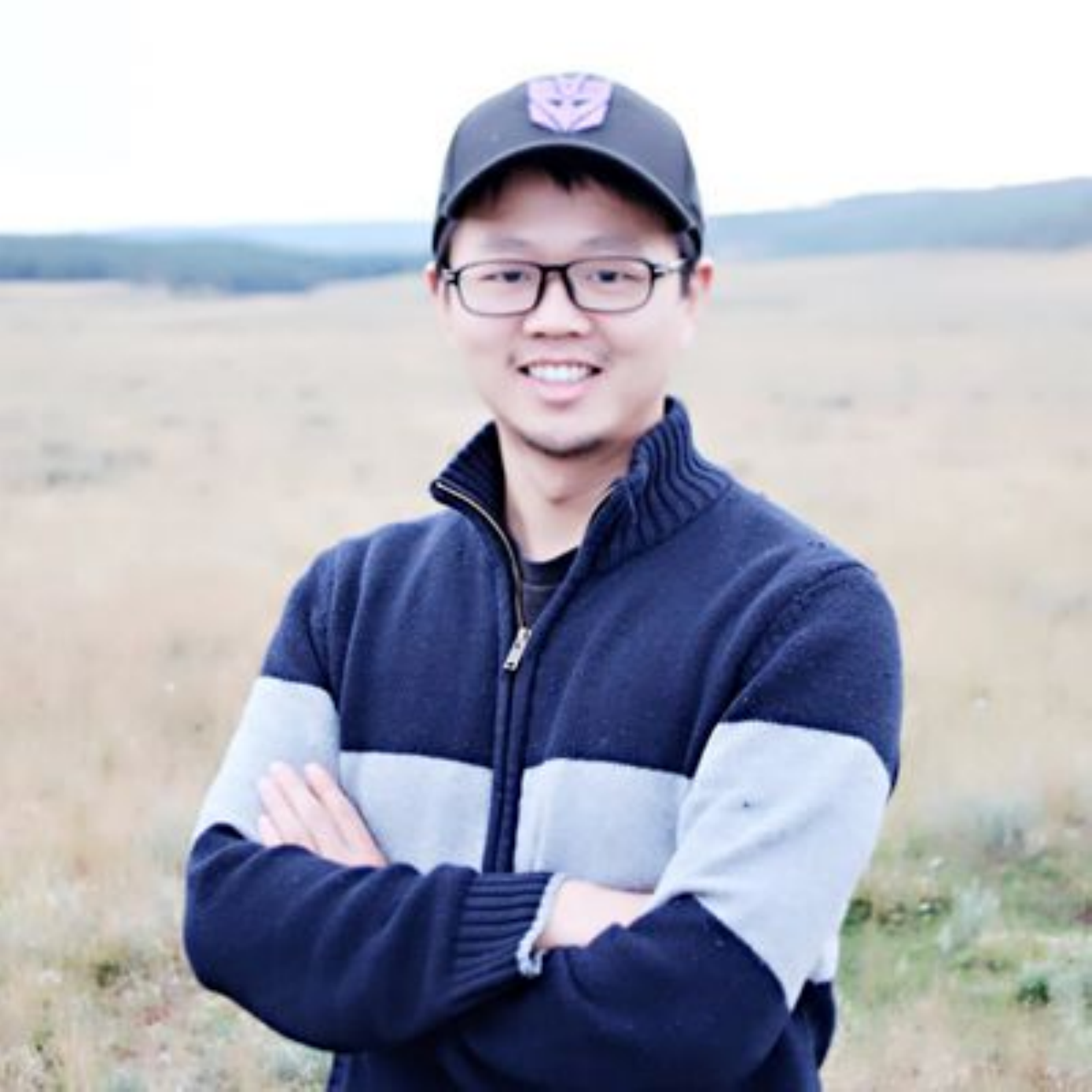}\vspace*{120pt}}%
\end{wrapfigure}
\noindent\small 
{\bf Xidao Wen} is a Ph.D candidate at PITT Computational Social Science Lab (Picso), from the School of Computing and Information, University of Pittsburgh. His research interests include data mining, machine learning and Information Visualization.}

\vspace{50pt}
\vbox{%
\begin{wrapfigure}{l}{80pt}
{\vspace*{-5pt}\includegraphics[width=0.2\textwidth]{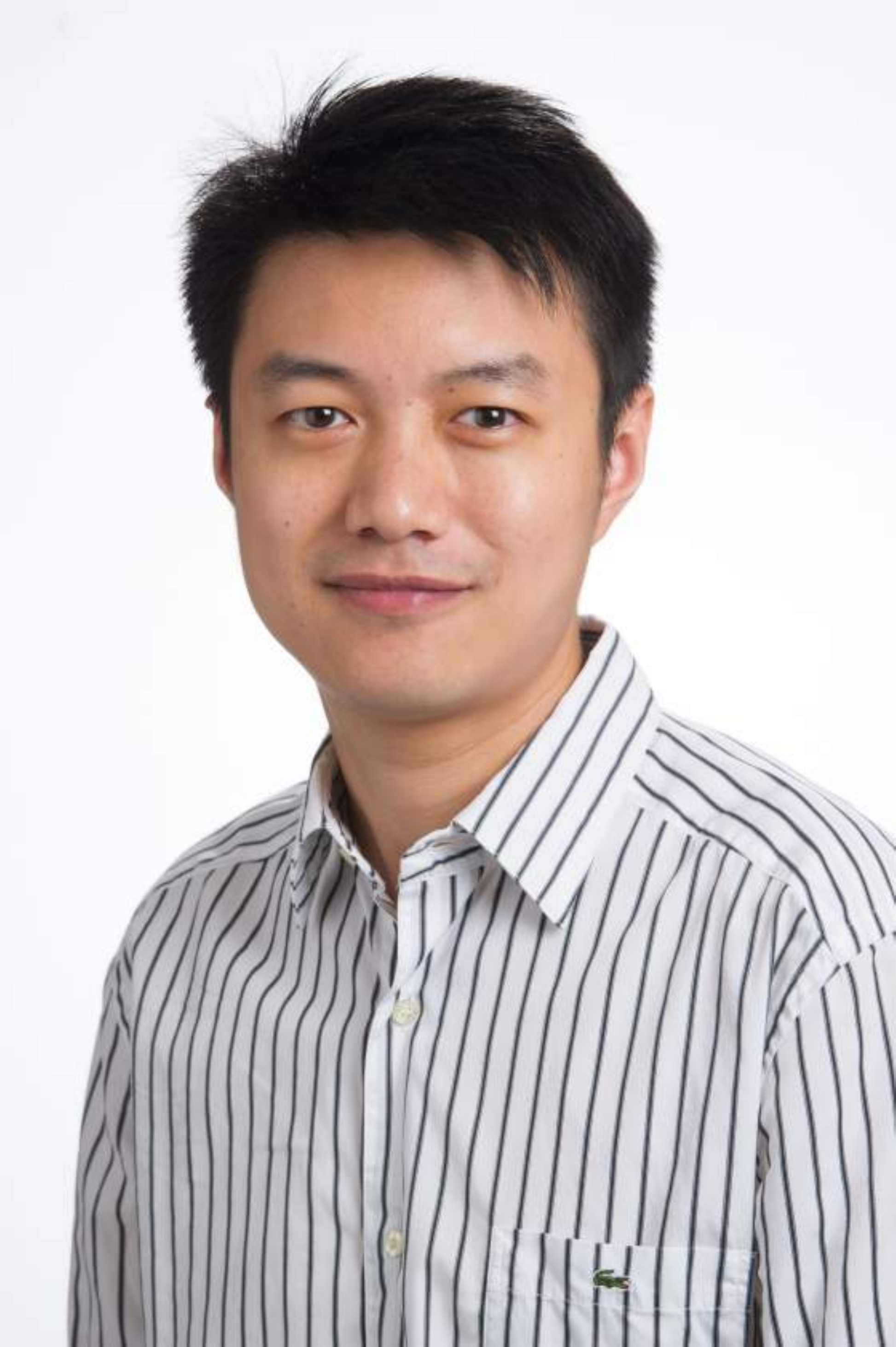}\vspace*{100pt}}%
\end{wrapfigure}
\noindent\small 
{\bf Rui Chen} received the Ph.D. degree in Computer Science from Concordia University. He is currently the Head of Data Science at Samsung Research America, leading the R\&D of production-scale machine/deep learning solutions for multiple Samsung services. Before that, he was a Research Assistant Professor in the Department of Computer Science at Hong Kong Baptist University and a Postdoctoral Fellow at the University of British Columbia. His research interests include machine learning, data privacy and databases. He has published more than 40 technical papers in top venues, including CSUR, VLDBJ, PVLDB, IEEE TKDE, IEEE TDSC, ACM KDD, ACM CCS and IEEE ICDE, and won CIKM 2015 Best Paper Runner Up, the Best Papers of ICDE 2016 and the Best Papers of ICDM 2018. He has served as program committee member for leading conferences, including ACM KDD, IEEE ICDM, ACM CIKM, PAKDD and DASFAA, and as reviewer for numerous flagship journals, including VLDBJ, PVLDB, IEEE TKDE, IEEE TDSC, IEEE TIFS, IEEE TOPS, ACM TOS and IEEE TSC.}

\vspace{50pt}
\vbox{%
\begin{wrapfigure}{l}{80pt}
{\vspace*{5pt}\includegraphics[width=0.2\textwidth]{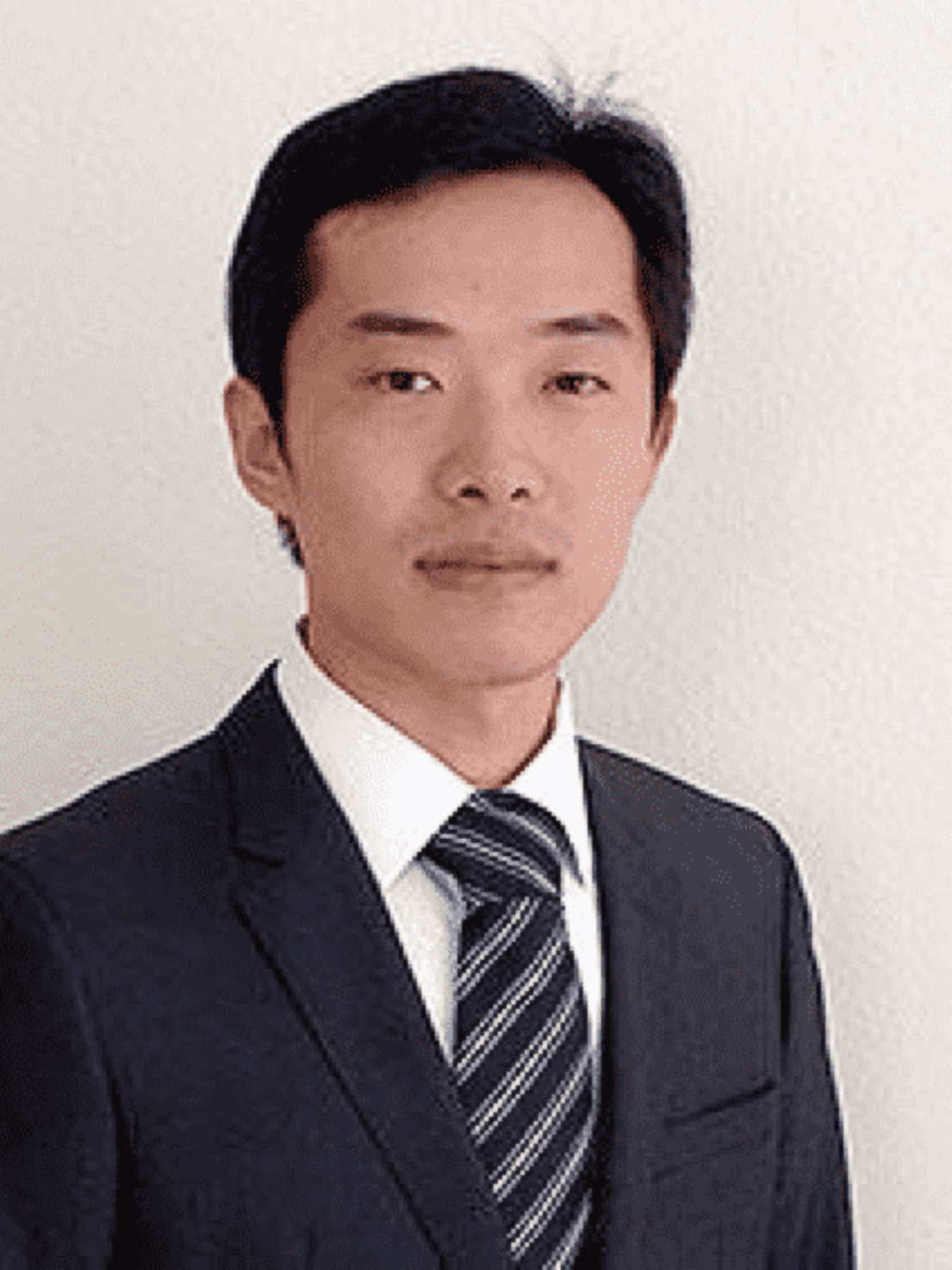}\vspace*{100pt}}%
\end{wrapfigure}
\noindent\small 
{\bf Yong Ge} received his Ph.D. in Information Technology from Rutgers, The State University of New Jersey in 2013, the M.S. degree in Signal and Information Processing from the University of Science and Technology of China (USTC) in 2008, and the B.E. degree in Information Engineering from Xi'an Jiao Tong University in 2005. He is currently an Assistant Professor at University of Arizona. His research interests include data mining and business analytics. He received the ICDM-2011 Best Research Paper Award, Excellence in Academic Research (one per school) at Rutgers Business School in 2013, and the Dissertation Fellowship at Rutgers University in 2012. He has published prolifically in refereed journals and conference proceedings, such as IEEE TKDE, ACM TOIS, ACM TKDD, ACM TIST, ACM SIGKDD, SIAM SDM, IEEE ICDM, and ACM RecSys. He has served as Program Committee members at premier conferences such as ACM SIGKDD conferences and IEEE ICDM conferences. Also he has served as a reviewer for numerous journals, including IEEE TKDE, ACM TKDD, ACM TIST, KAIS, Information Science, and TSMC-B.}

\vspace{50pt}
\vbox{%
\begin{wrapfigure}{l}{80pt}
{\vspace*{5pt}\includegraphics[width=0.2\textwidth]{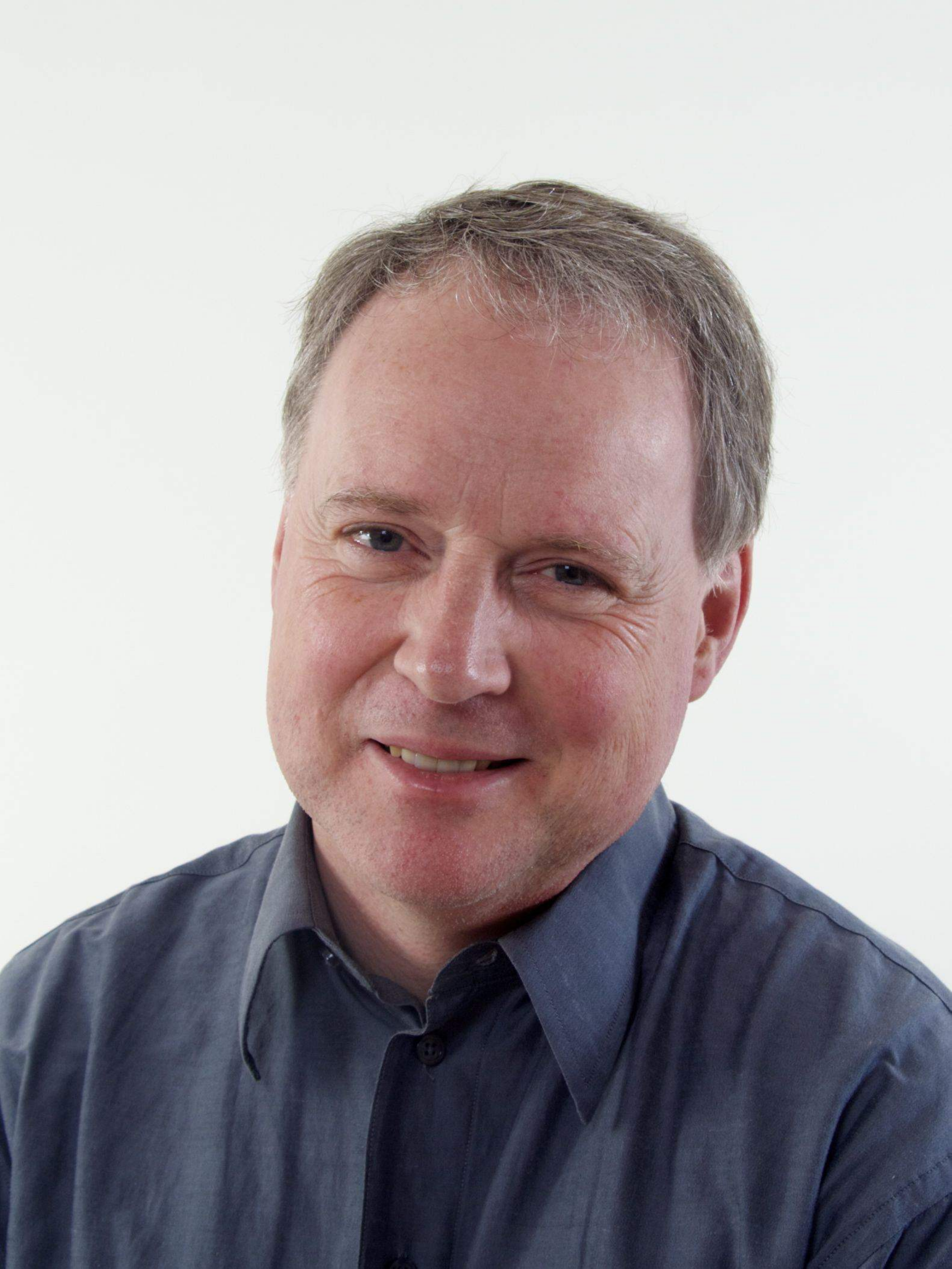}\vspace*{100pt}}%
\end{wrapfigure}
\noindent\small 
{\bf Nick Duffield} is a Professor in the Department of Electrical and Computer Engineering at Texas A\&M University, a Professor by Courtesy, in the Department of Computer Science and Engineering, and Director of the Texas A\&M Institute of Data Science. From 1995 until 2013, he worked at AT\&T Labs-Research, Florham Park, NJ, where he was a Distinguished Member of Technical Staff and an AT\&T Fellow. He previously held post-doctoral and faculty positions in Dublin, Ireland, and Heidelberg, Germany. He received a BA in Natural Sciences in 1982 and an MMath in 1983 from the University of Cambridge, UK, and a PhD in Mathematical Physics from the  University of London, U.K., in 1987. His research concerns data science and computer networking, with current projects involving algorithms for data streaming and machine learning, computer network measurement and resilience, and applications of data science to transportation, agriculture and hydrology. He is an IEEE Fellow, an IET Fellow, and has twice received the ACM SIGMETRICS Test-of-Time award. He is an elected member of the governing board of ACM SIGMETRICS. He was an Associate Editor for the IEEE/ACM Transactions on Networking from 2007-2011, and has been an Editor-at-Large since 2014. He is Chief Editor for Big Data Network at Frontiers.}

\vspace{15pt}
\vbox{%
\begin{wrapfigure}{l}{80pt}
{\vspace*{5pt}\includegraphics[width=0.2\textwidth]{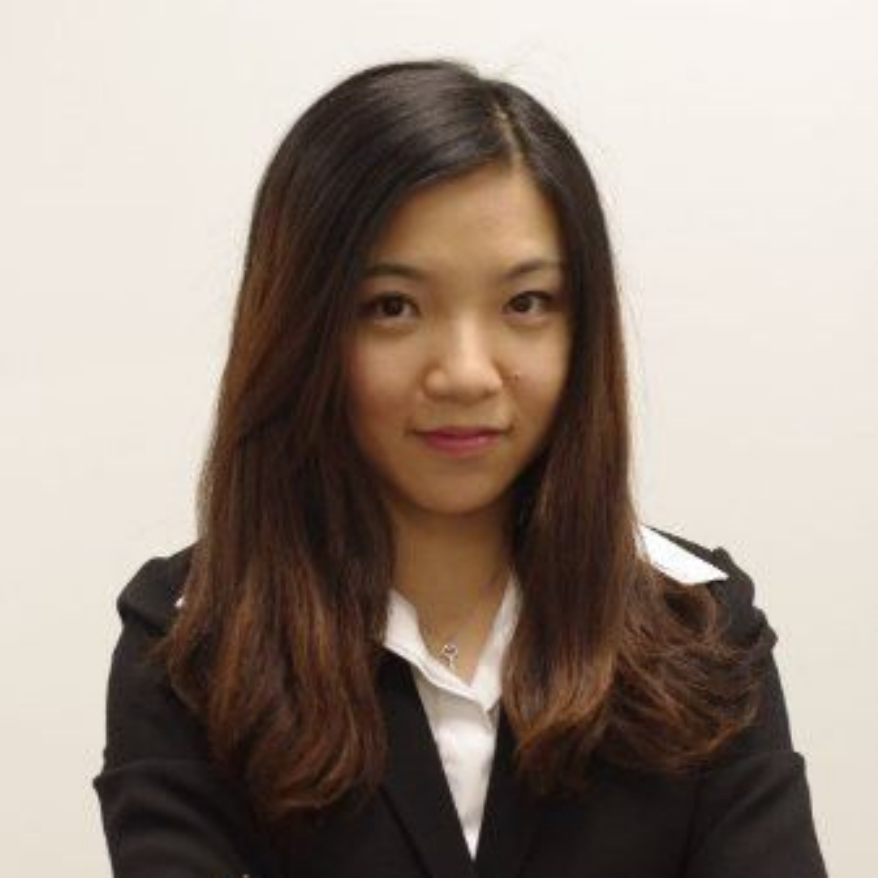}\vspace*{100pt}}%
\end{wrapfigure}
\noindent\small 
{\bf Na Wang} received her M.S. and Ph.D. degrees in Information Sciences and Technology from the Pennsylvania State University. She is currently a Staff Software Engineer at Samsung Research America. Her research interests include machine learning and human-computer interaction. She has published papers in top conference proceedings, including IEEE ICDM, ACM SIGCHI, ACM IUI, ACM MobileHCI, and ACM CSCW, and won Best Paper Award of CHI 2015.}

\correspond{Rui Chen, Samsung Research America, Mountain View, CA, USA. Email: rui.chen1@samsung.com}
\label{lastpage}
\end{document}